%% file: main.tex
\documentclass{article} 
\usepackage{iclr2027_conference,times}

\input{math_commands.tex}

\usepackage{hyperref}
\usepackage{url}

\usepackage{algorithm}
\usepackage{algorithmic}

\usepackage{threeparttable}
\usepackage{multirow}
\usepackage{booktabs}
\usepackage{makecell}
\usepackage[table]{xcolor}
\usepackage{pifont}
\usepackage{array}
\usepackage{xcolor}
\usepackage{amsmath}
\usepackage{amssymb}
\usepackage{amsthm}
\usepackage{wrapfig}
\usepackage{graphicx}

\usepackage{titletoc}

\newtheorem{theorem}{Theorem}

\title{PANO: Parallel Latent Space Neural Operator for Solving PDEs}

\author{
Zijiang Yang\textsuperscript{\rm 1}, Xiaomeng Wu\textsuperscript{\rm 1}, Chen Kuang\textsuperscript{\rm 1}, Dongmei Fu\textsuperscript{\rm 1}\textsuperscript{\rm 2}\thanks{Corresponding author.}\\
\textsuperscript{\rm 1}University of Science and Technology Beijing\\
\textsuperscript{\rm 2}Beijing Engineering Research Center of Industrial Spectrum Imaging\\
\texttt{zijiangyang@xs.ustb.edu.cn, fdm\_ustb@ustb.edu.cn}
}

\iclrfinalcopy 
\begin{document}

\maketitle

\begin{abstract}
Aggregating observations into a limited number of latent tokens and learning physical interactions in the latent space has emerged as an effective approach for Transformer-based neural operators to solve Partial Differential Equation (PDE) problems involving large-scale observations.
However, existing methods typically compress observations using a single aggregation pattern, making physical information not retained by that pattern invisible to the latent space and thereby constraining physical interactions within it.
To address this limitation, we propose \textbf{PA}rallel Latent Space \textbf{N}eural \textbf{O}perator (\textbf{PANO}), a novel neural operator for solving PDEs that formulates physical interactions among observations as an additive combination of interaction results from parallel latent spaces with shared parameters.
Specifically, PANO transforms observations into multiple distinct compressed representations through independent mappings, learns physical interactions in parallel based on these representations, and finally fuses the decoded results to predict the target physical field.
Within PANO, we further propose \textbf{C}oupled \textbf{A}ssignment \textbf{P}rojection (\textbf{CAP}) to construct bidirectionally consistent encoding and decoding mappings, and \textbf{Pa}irwise \textbf{L}atent Space \textbf{A}ttention Layer (\textbf{PaLA}) to jointly update latent representations of each latent space through adaptive cross-space interactions.
Extensive experiments on nine benchmarks spanning diverse physical systems, geometries, and observation scales demonstrate that PANO achieves state-of-the-art performance.
Notably, PANO reduces equivalent plastic strain prediction errors on Unilateral Stamping by 28.3\% (or more) and lift-coefficient prediction errors on AirCraft by 21.4\% (or more) compared with existing methods. Code is available at https://github.com/ZijiangY1116/PANO.
\end{abstract}

\section{Introduction}

\begin{figure}[!t]
    \centering
    \includegraphics[width=0.99\linewidth]{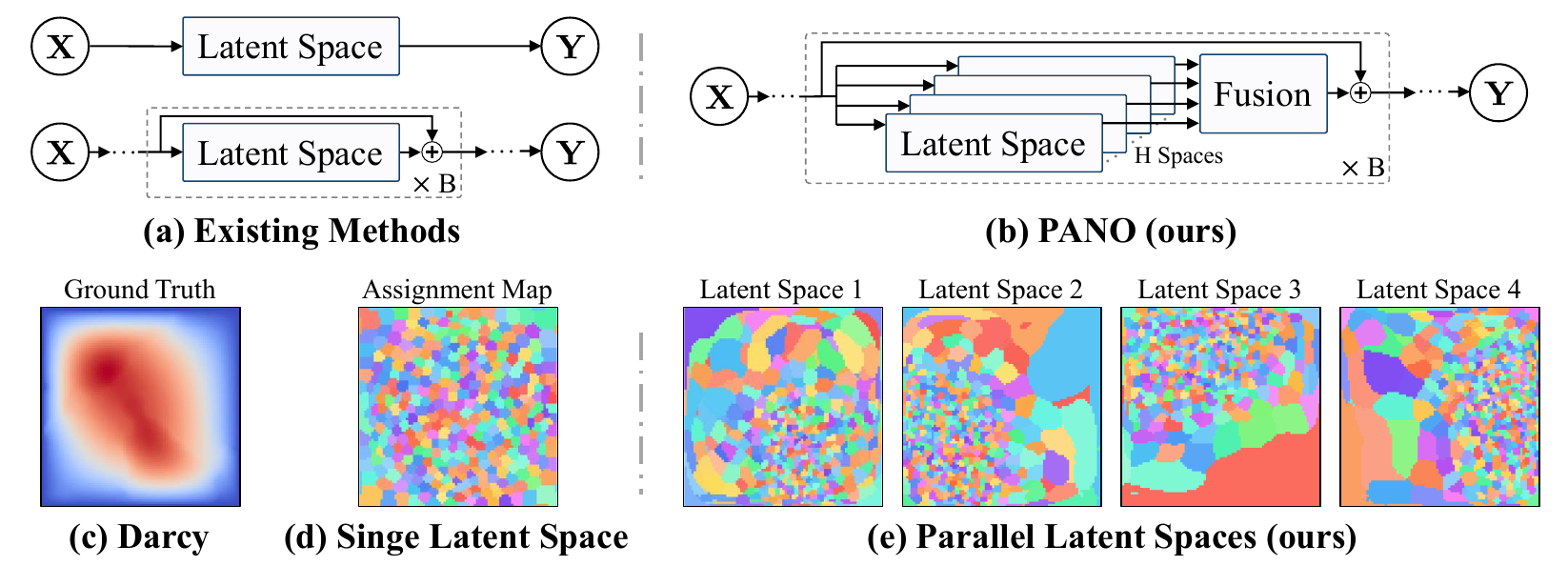}
    \caption{
      Motivation.
      (a) Existing methods encode input point tokens into a latent space through a single aggregation pattern.
      (b) PANO additively combines interaction results from parallel latent spaces.
      (c) Darcy ground truth.
      (d–e) Assignment maps of a single-latent-space model~\citep{lno} and PANO, with points colored by their highest-weight latent token.
      Compared with (d), the parallel latent spaces in PANO exhibit distinct local preferences, collectively enabling a more comprehensive latent representation of the input.
      $\mathbf{X}$ and $\mathbf{Y}$ denote the observation and the target physical field, respectively. $B$ denotes the number of blocks.
    }
    \label{fig:firstfig}
\end{figure}

Partial Differential Equations (PDEs) play a critical role in describing physical processes in natural and engineering systems~\citep{raissi2020hidden}.
Recently, Transformer-based neural operators have emerged as effective methods for solving PDEs on general geometries by embedding observations into point tokens and learning physical interactions among them~\citep{oformer,kovachki2023neural}.
For PDE problems involving large-scale observations, recent methods have achieved State-Of-The-Art (SOTA) performance by further aggregating point tokens into a limited number of latent tokens and learning physical interactions in latent spaces, thereby avoiding quadratic interaction costs in the observation space~\citep{upt,transolverv3}.
However, as shown in Figure~\ref{fig:firstfig}, these methods typically learn physical interactions among input tokens in a latent space constructed by a single aggregation mapping~\citep{lno,mono}, which provides only a compressed representation under a particular aggregation pattern, constraining latent-space physical interactions to the information retained by that pattern.

This limitation is particularly pronounced in models that concentrate multiple interaction layers in a single latent space, as subsequent physical interactions rely on the initial latent tokens obtained through encoding aggregation and cannot recover physical distinctions lost during encoding merely through deeper latent modules~\citep{lno}.
Another type of architecture progressively updates physical states by stacking encoding-interaction-decoding blocks with observation-space residual connections~\citep{transolverpp}.
This design provides a bypass for information insufficiently retained by the current aggregation and allows subsequent modules to reaggregate it, but does not ensure that tokens in the current latent space retain enough information to support effective physical interactions~\citep{linearno}.
Notably, Transolver proposes Physics-Attention and extends it to multi-head Physics-Attention, thereby offering a promising way to expand information retention during encoding~\citep{transolver}. However, the heads in multi-head physics attention are still required to jointly constitute a single representation, rather than serving as independent information pathways.
These issues motivate us to reconsider how information in point tokens can be more fully transferred to latent tokens to support physical interactions in latent spaces.

In this work, we propose \textbf{PA}rallel Latent Space \textbf{N}eural \textbf{O}perator (\textbf{PANO}), which formulates physical-state updates of point tokens as an additive combination of physical interaction results from parallel latent spaces.
PANO first encodes the same point tokens into multiple latent spaces through independent mappings, forming multiple distinct compressed representations.
Each latent space then learns physical interactions in parallel based on its corresponding aggregation results.
Finally, the latent tokens from each path are decoded back to the observation space and fused into a combined state update for the point tokens.
PANO reduces reliance on any single aggregation mapping, allowing information in point tokens to be more fully transferred to latent tokens through multiple aggregation patterns, thereby alleviating the representational bottleneck of existing methods (Figure~\ref{fig:firstfig} (e)).
Within PANO, we further propose \textbf{C}oupled \textbf{A}ssignment \textbf{P}rojection (\textbf{CAP}) for learnable latent-space construction with bidirectionally consistent encoding and decoding, and \textbf{Pa}irwise \textbf{L}atent Space \textbf{A}ttention Layer (\textbf{PaLA}) to jointly update latent tokens across latent spaces through adaptive cross-space interactions.
We evaluate PANO on nine benchmarks encompassing standard PDE tasks, as well as challenging engineering tasks involving airfoil flow extrapolation, multi-solid systems, and three-dimensional aerodynamics, showing that PANO outperforms existing SOTA methods.
Our main contributions are summarized as follows.
\begin{itemize}
    \item We propose PANO, a novel neural operator for solving PDEs, which constructs parallel latent spaces to jointly predict target physical fields from observations.

    \item Within PANO, we propose CAP for bidirectionally consistent encoding and decoding mappings, and PaLA for adaptive cross-space interactions.

    \item Extensive experiments on nine benchmarks spanning diverse physical systems, geometries, and observation scales demonstrate the advantages of PANO over existing SOTA methods.
\end{itemize}

\section{Related Work}

\noindent \textbf{Physics-Informed Neural Networks.}
Physics-informed neural networks (PINNs) use learnable models to provide continuous approximations of target physical fields~\citep{karniadakis2021physics,raissi2020hidden,niaki2021physics}.
Specifically, these models are optimized with physical residual losses derived from the governing equations and solution conditions~\citep{wandel2022spline,yang2023dmis,wu2026multi}.
Typically, PINNs rely on explicit and sufficiently complete definitions of physical systems, limiting their applicability to complex real-world problems~\citep{luo2025physics}.

\noindent \textbf{Neural Operators.}
Neural operators learn mappings from input functions to target physical fields using large amounts of observation data~\citep{lu2019deeponet,fno,kovachki2023neural,zhou2024unisolver,mpp,gno,gino,3d-geoca,gnot,Unisoma,wang2025learning}.
Among them, Transformer-based neural operators have achieved substantial progress by adaptively modeling nonlocal relationships and flexibly processing observations across geometries~\citep{ono,lno}.
However, the quadratic complexity of standard attention makes direct interactions in the observation space costly~\citep{vaswani2017attention}.
Although some methods~\citep{galerkin,oformer}
reduce interaction complexity through linear attention, scaling to large-scale PDE problems still presents substantial computational and memory challenges~\citep{transolver}.
Recent methods compress observations into latent tokens~\citep{lno,AROMA,upt,linearno,mono},
enabling PDE solving at the million-point scale~\citep{transolverpp,transolverv3}.

\section{Parallel Latent Space Neural Operator}

\noindent \textbf{Preliminaries.}
Given a PDE-solving problem defined on an observation domain $\mathcal D$ and observations $\mathbf{X}=\{\mathbf x_i\}_{i=1}^{N}$ collected at $N$ spatial locations, operator learning aims to learn a mapping from the input observations to the target physical field.
Depending on the physical system, the observations $\mathbf{X}\in\mathbb R^{N\times C_{\mathrm{in}}}$ comprise problem-specific conditions,
where $C_{\mathrm{in}}$ denotes the observation dimension.

\begin{figure}[!t]
    \centering
    \includegraphics[width=0.95\linewidth]{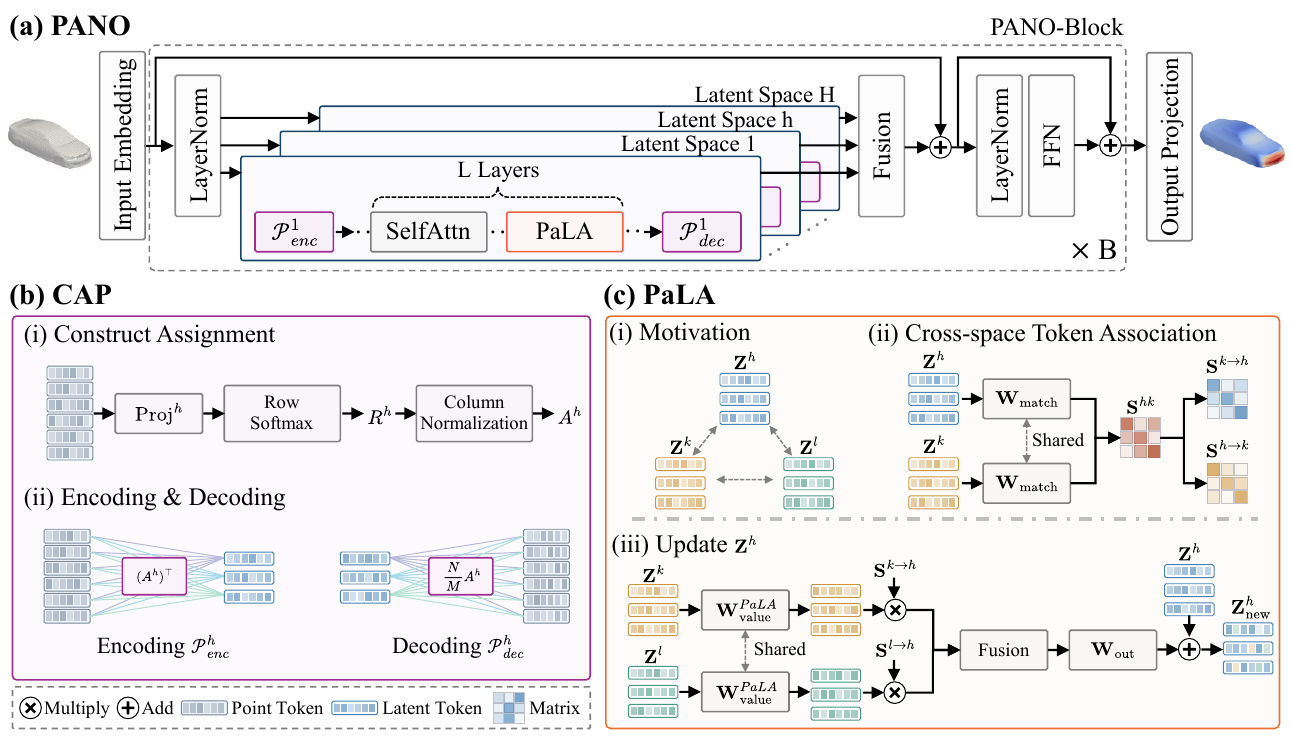}
    \caption{
    Illustration of PANO.
    (a) Observations are embedded into point tokens, updated through stacked PANO-Blocks, and projected to the target physical field.
    (b) CAP constructs shared assignments to obtain bidirectionally consistent encoding and decoding mappings.
    (c) PaLA learns cross-space token associations.
    SelfAttn denotes a standard self-attention layer.
    }
    \label{fig:mainfig}
\end{figure}

\noindent \textbf{PANO.}
The overall architecture is illustrated in Figure~\ref{fig:mainfig}.
PANO first embeds observations into point tokens through a pointwise projection, updates them through stacked Parallel Latent Space Blocks (PANO-Blocks), and finally predicts the target physical field through an output projection.

\subsection{Parallel Latent Space Block}

To reduce the dependence of physical interaction learning on a single aggregation representation, each PANO-Block encodes the same point tokens into $H$ parallel latent spaces through independent mappings, thereby supporting simultaneous physical interaction based on multiple aggregation patterns.
A PANO-Block without PaLA can be expressed as:
\begin{equation}
F_{\mathrm{latent}}
=
\frac{1}{H}(
\underbrace{
\mathcal{P}_{\mathrm{dec}}^{1}(
\mathcal{M}_{\mathrm{latent}}(
\mathcal{P}_{\mathrm{enc}}^{1}(\operatorname{LN}(F))
))
}_{\text{Latent space }1}
+\cdots+
\underbrace{
\mathcal{P}_{\mathrm{dec}}^{H}(
\mathcal{M}_{\mathrm{latent}}(
\mathcal{P}_{\mathrm{enc}}^{H}(\operatorname{LN}(F))
))
}_{\text{Latent space }H}
),
\label{eq:parallel_block}
\end{equation}
where $\operatorname{LN}$ denotes LayerNorm, $\mathcal{M}_{\mathrm{latent}}$ denotes the latent module, and $\mathcal{P}_{\mathrm{enc}}^h$ and $\mathcal{P}_{\mathrm{dec}}^h$ denote the $h$-th latent-space encoding mapping and decoding mapping, respectively.
$F,F_{\mathrm{latent}}\in\mathbb R^{N\times C}$ denote the input point tokens and the decoded outputs, respectively, with $C$ denoting the feature dimension.

Both $\mathcal{P}_{\mathrm{enc}}^h$ and $\mathcal{P}_{\mathrm{dec}}^h$ are constructed by CAP from the learnable assignments of the $h$-th latent space.
In the full PANO-Block, $\mathcal{M}_{\mathrm{latent}}$ comprises multiple self-attention layers and PaLA, which jointly update latent tokens through within-space and cross-space interactions, respectively.
Notably, the parameters of $\mathcal{M}_{\mathrm{latent}}$ are shared across latent spaces within each PANO-Block.
Moreover, following common practice~\citep{vaswani2017attention,transolver}, we combine the input point tokens with the decoded latent interaction results through a residual connection, then apply a pointwise FFN to further integrate features.

\subsection{Coupled Assignment Projection}

Mapping construction is crucial to the stability of latent spaces~\citep{mono}.
Therefore, we propose CAP, which constructs bidirectionally consistent encoding and decoding mappings from shared assignments to transfer information between point tokens and latent tokens.
Specifically, for the $h$-th latent space, we first compute row-normalized assignment weights through a learnable mapping, then normalize each column by the total weight received by the corresponding latent token to obtain the assignment matrix $A^h\in\mathbb R^{N\times M}$:
\begin{equation}
    R^h=\operatorname{Softmax}_{\mathrm{row}}
    \left(\operatorname{Proj}^h(\operatorname{LN}(F))\right),
    \qquad
    A^h=R^h\operatorname{Diag}\left((R^h)^\top\mathbf{1}_N\right)^{-1},
    \label{eq:cap_assignment}
\end{equation}
where $\operatorname{Proj}^h$ denotes the $h$-th learnable assignment projector, $R^h\in\mathbb R^{N\times M}$ denotes the assignment matrix after row-wise Softmax, $M$ denotes the number of latent tokens per space, $\mathbf{1}_N$ is the all-ones vector of length $N$, and $\operatorname{Diag}$ converts a vector into a diagonal matrix.
$R^h$ describes the relative assignment weights of each point token across latent tokens.
$A^h$ normalizes the total weight received by each latent token to one.
Then, we construct the mappings:
\begin{equation}
    \mathcal{P}_{\mathrm{enc}}^h(\cdot)
    =(A^h)^\top(\cdot),
    \qquad
    \mathcal{P}_{\mathrm{dec}}^h(\cdot)
    =\frac{N}{M}A^h(\cdot).
    \label{eq:cap_projection}
\end{equation}
Encoding aggregates point tokens into latent tokens, while decoding transfers the updated information back to the observation space along the same assignment relations, with $N/M$ adjusting the total-weight scale for the difference in token counts.

\noindent \textbf{Latent-Space Stability Analysis.}
The construction of $\mathcal{P}_{\mathrm{enc}}^h(\cdot)$ and $\mathcal{P}_{\mathrm{dec}}^h(\cdot)$ ensures that each latent token receives and returns information according to the same relative weight distribution.
For convenience, we use $\mathbf{1}_M$ to denote the all-ones vector of length $M$.
Since $(A^h)^\top\mathbf{1}_N=\mathbf{1}_M$, the two directions satisfy the following total-weight constraints:
\begin{equation}
    \mathcal{P}_{\mathrm{enc}}^h(\mathbf{1}_N)=\mathbf{1}_M,
    \qquad
    \left(\frac{N}{M}A^h\right)^\top\mathbf{1}_N
    =\frac{N}{M}\mathbf{1}_M,
    \label{eq:cap_weight_constraints}
\end{equation}
The first constraint avoids feature-scale differences caused by unequal total weights received by latent tokens.
The second ensures that each latent token distributes the same total weight $N/M$ to the point tokens during decoding.
These constraints support latent-space stability in terms of feature scale and weight allocation while preserving flexibility in the assignment patterns of latent tokens.

Moreover, we show that the decoding fusion admits an additive kernel representation for $H\geq1$ parallel latent spaces using CAP and a shared-parameter single-head self-attention layer.
We further demonstrate that parallel latent spaces can expand representational capacity while reducing the observation-space information jointly omitted during encoding.
The proof of Theorem~\ref{thm:cap_additive_kernel} and proofs of these two properties are provided in Appendix~\ref{sec:proof_of_theorem1} and Appendix~\ref{sec:properties_theorem1}, respectively.
\begin{theorem}[\textbf{The Additive Kernel Representation of Parallel Latent Spaces with CAP}]
\label{thm:cap_additive_kernel}
Let $(\mathcal D,\mu)$ be a physical domain equipped with a probability measure $\mu$, and let $\mathcal F:\mathcal D\to\mathbb R^C$ be a bounded measurable point-token field.
Assuming finite and measurable assignment logits in each space, the averaged decoded outputs $\mathcal T_H$ admit an input-dependent additive kernel representation: $\mathcal T_H(\mathcal F)(x)=\int_{\mathcal D}\overline{\kappa}_{\mathcal F}(x,\xi)\mathcal F(\xi)\mathbf{W}_{\mathrm{value}}^{self}\,d\mu(\xi), \overline{\kappa}_{\mathcal F}(x,\xi)=\frac{1}{H}\sum_{h=1}^{H}\kappa_{\mathcal F}^{h}(x,\xi),$
where $x\in\mathcal D$ is the output evaluation location, $\xi\in\mathcal D$ is the integration variable, $\mathbf{W}_{\mathrm{value}}^{self}\in\mathbb R^{C\times C}$ is the shared value projection matrix of self-attention, and $\kappa_{\mathcal F}^{h}$ is the kernel induced by the $h$-th latent space.
\end{theorem}

\subsection{Pairwise Latent Space Attention Layer}

Decoding-based fusion integrates outputs from latent spaces only in the observation space, without directly establishing interactions across latent spaces.
To complement this fusion, we propose PaLA, which learns cross-space token associations to exchange information between latent spaces.

Given latent token sequences $\mathbf{Z}^h,\mathbf{Z}^k\in\mathbb R^{M\times C}$ from two distinct latent spaces $h$ and $k$, we compute cross-space attention scores through a shared matching projection:
$\mathbf{S}^{hk}=\frac{(\mathbf{Z}^h\mathbf{W}_{\mathrm{match}})(\mathbf{Z}^k\mathbf{W}_{\mathrm{match}})^\top}{\sqrt{C}},$
where $\mathbf{W}_{\mathrm{match}}\in\mathbb R^{C\times C}$ denotes the learnable shared matching matrix, and $\mathbf{S}^{hk}\in\mathbb R^{M\times M}$ denotes the token matching scores between the two latent token sequences.
We then normalize the scores row-wise in each direction to obtain the attention weights $\mathbf{S}^{h\rightarrow k}=\operatorname{Softmax}_{\mathrm{row}}((\mathbf{S}^{hk})^\top)$ and $\mathbf{S}^{k\rightarrow h}=\operatorname{Softmax}_{\mathrm{row}}(\mathbf{S}^{hk})$, where the arrows indicate the direction of information transfer.
This construction lets latent spaces $h$ and $k$ share token matching relations while independently determining how to aggregate information from each other.

For each latent token sequence $\mathbf{Z}^h$, PaLA averages information from the other latent spaces and applies an output projection and a residual connection to update it:
\begin{equation}
    \mathbf{Z}_{\mathrm{new}}^h
    =\mathbf{Z}^h+
    (
    \frac{1}{H-1}\sum_{k\ne h}
    \mathbf{S}^{k\rightarrow h}
    (\mathbf{Z}^k\mathbf{W}_{\mathrm{value}}^{PaLA})
    )\mathbf{W}_{\mathrm{out}},
    \label{eq:pala_update}
\end{equation}
where $\mathbf{W}_{\mathrm{value}}^{PaLA},\mathbf{W}_{\mathrm{out}}\in\mathbb R^{C\times C}$ denote the value and output projections shared across latent spaces, respectively, and $\mathbf{Z}_{\mathrm{new}}^h$ denotes the updated latent tokens.
When $H=1$, no cross-space interaction is available, and PaLA reduces to the identity mapping.
We then apply a token-wise FFN with LayerNorm and a residual connection to integrate the updated latent token features.

We further show that decoding fusion of cross-space interactions admits a cross-space kernel representation for $H>1$ parallel latent spaces using CAP and PaLA with shared parameters.
\begin{theorem}[\textbf{The Cross-Space Kernel Representation of Parallel Latent Spaces with CAP and PaLA}]
\label{thm:pala_cross_kernel}
Under the same assumptions on the domain, input field, and assignments as in Theorem~\ref{thm:cap_additive_kernel}, for $H>1$ parallel latent spaces, the averaged decoded cross-space updates $\mathcal T_{\mathrm{cross}}$ admit an input-dependent cross-space kernel representation: $\mathcal T_{\mathrm{cross}}(\mathcal F)(x)=\int_{\mathcal D}\kappa_{\mathcal F}^{\mathrm{cross}}(x,\xi)\mathcal F(\xi)\mathbf{W}_{\mathrm{value}}^{PaLA}\mathbf W_{\mathrm{out}}\,d\mu(\xi),$
where $\mathbf{W}_{\mathrm{value}}^{PaLA}$ and $\mathbf W_{\mathrm{out}}$ are the shared value and output projection matrices of PaLA, $\kappa_{\mathcal F}^{\mathrm{cross}}$ is the kernel induced by cross-space attention of PaLA, and the representation holds for $\mu$-almost every $x\in\mathcal D$.
\end{theorem}

Building on Theorem~\ref{thm:pala_cross_kernel}, we demonstrate that PaLA enables information exchange between latent spaces. We further demonstrate that PaLA can enrich the representation of any latent space by transferring information omitted by its own encoding but retained by other spaces through cross-space interactions.
The proof of Theorem~\ref{thm:pala_cross_kernel} and analyses of these properties are provided in Appendix~\ref{sec:proof_of_theorem2} and Appendix~\ref{sec:properties_theorem2}, respectively.

\subsection{PANO Variants}

PANO can be scaled by increasing the number of parallel latent spaces $H$, the number of latent tokens per space $M$, the feature dimension $C$, and the depth.
To evaluate different scaling configurations, we introduce PANO-Parallel (PANO-P) and PANO-Deep (PANO-D).
Both variants contain two PANO-Blocks, with the following configurations:
\begin{itemize}
    \item \textbf{PANO-Parallel (PANO-P)} uses $H=4$, $M=512$, and $C=96$.
    Each block contains
    $2\times\mathrm{SelfAttn}\rightarrow\mathrm{PaLA}\rightarrow2\times\mathrm{SelfAttn}$,
    emphasizing physical interaction learning through more parallel aggregation patterns.

    \item \textbf{PANO-Deep (PANO-D)} uses $H=2$, $M=1024$, and $C=192$.
    Each block contains
    $3\times\mathrm{SelfAttn}\rightarrow\mathrm{PaLA}\rightarrow3\times\mathrm{SelfAttn}$,
    allocating greater capacity to processing within each latent space.
\end{itemize}
Both PANO-P and PANO-D use a total of $HM=2048$ latent tokens per block to learn physical interactions among the input point tokens.

\section{Experiments}

\subsection{Experiment Settings}

\begin{figure}[!t]
    \centering
    \includegraphics[width=0.99\linewidth]{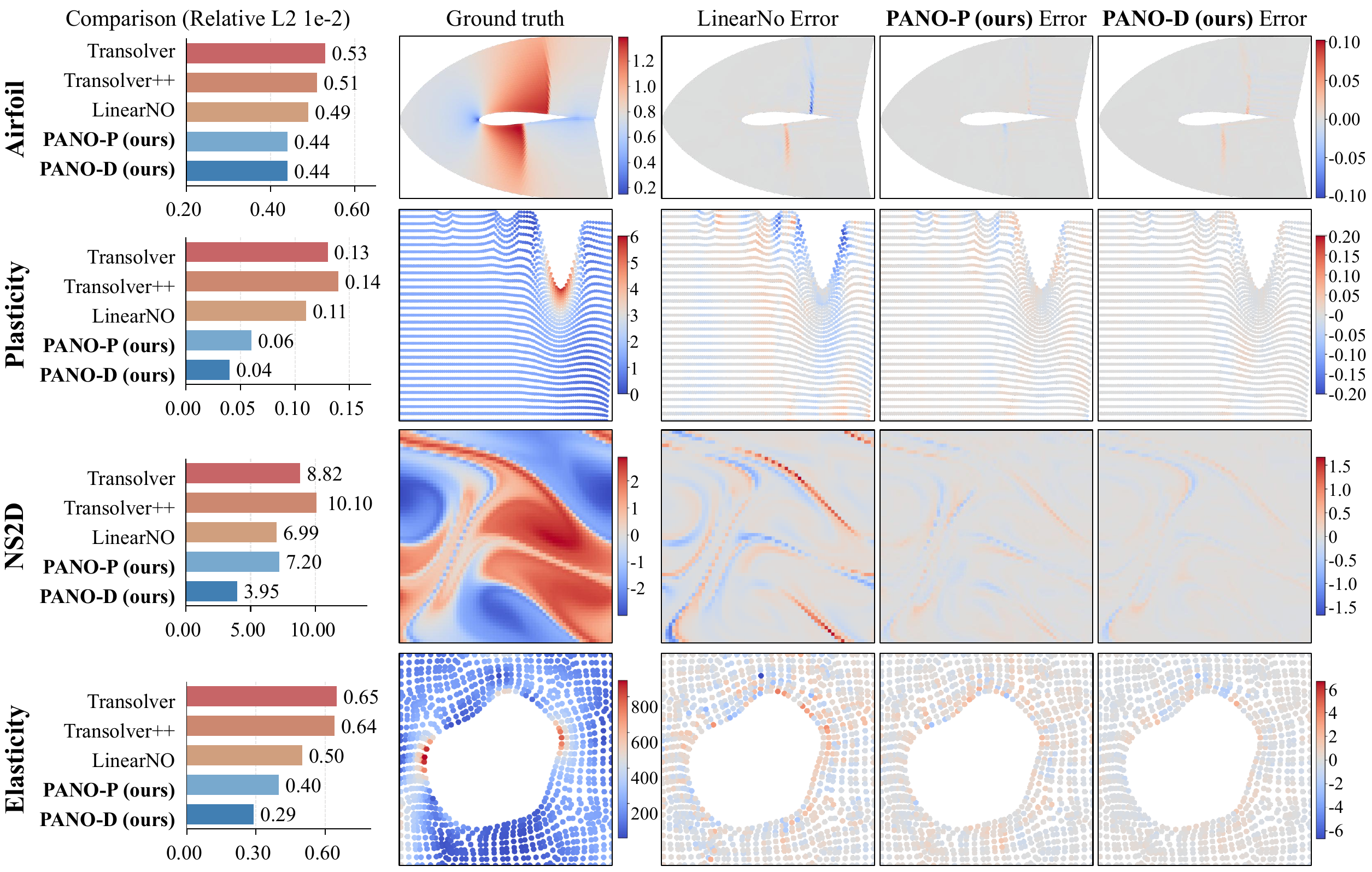}
    \caption{
    Comparison on standard benchmarks in relative L2 error ($\downarrow$).
    PANO-D achieves the best performance.
    Moreover, compared with LinearNO, PANO exhibits lower prediction errors in regions with sharp variations in the physical fields.
    }
    \label{fig:standard_benchmark_compare_preview}
\end{figure}

\noindent \textbf{Benchmarks.}
We first evaluate PANO on five standard benchmarks~\citep{fno,geo-fno}, including Airfoil, Plasticity, Navier-Stokes (NS2D), Elasticity, and Darcy.

\begin{wraptable}{r}{0.5\textwidth}
   \caption{
    Relative L2 errors ($\downarrow$) on Darcy at three resolutions.
    Best and second-best results are \textbf{bold} and \underline{underlined}, respectively.
  }
  \centering
  \small
  \setlength{\tabcolsep}{3pt}
  \renewcommand{\arraystretch}{1.05}
  \begin{tabular*}{\linewidth}{@{\extracolsep{\fill}}l|ccc@{}}
    \toprule
    Method & Darcy85 & Darcy141 & Darcy211 \\
    \midrule
    LNO
    & 0.0081 & 0.0074 & 0.0073 \\
    Transolver
    & \underline{0.0055} & 0.0064 & 0.0063 \\
    LinearNO
    & \textbf{0.0050} & 0.0053 & 0.0053 \\
    \midrule
    \textbf{PANO-P (ours)}
    & \underline{0.0055} & \underline{0.0042}
    & \underline{0.0041} \\
    \textbf{PANO-D (ours)}
    & \textbf{0.0050} & \textbf{0.0039}
    & \textbf{0.0039} \\
    \bottomrule
  \end{tabular*}
  \label{tab:multi_resolution_eval}
\end{wraptable}

To evaluate the performance of PANO on real-world advanced benchmarks, we further conduct experiments on four benchmarks, including AirfRANS~\citep{airfrans}, Unilateral Stamping~\citep{Unisoma}, ShapeNetCar~\citep{shapenetcar}, and AirCraft~\citep{transolverpp}.
These benchmarks cover two-dimensional airfoil flow prediction and out-of-distribution generalization, stamping in multi-solid systems, three-dimensional car aerodynamics, and three-dimensional AirCraft flow prediction with approximately 330K observations, respectively.

\noindent \textbf{Baselines.}
We compare PANO with ten representative methods, including GNOT~\citep{gnot}, GINO~\citep{gino}, LNO~\citep{lno}, Transolver~\citep{transolver}, Transolver++~\citep{transolverpp}, LinearNO~\citep{linearno}, OFormer~\citep{oformer}, ONO~\citep{ono}, MeshGraphNet~\citep{meshgraphnet}, and Unisoma~\citep{Unisoma}.
Complete results across all benchmarks are provided in the appendix~\ref{sec:detailed_results}.

\noindent \textbf{Implementations.}
Detailed implementation descriptions are provided in the Appendix~\ref{sec:implementation_details}.

\subsection{Standard Benchmarks}

\begin{table*}[t]
  \caption{
    Comparison on AirfRANS under the standard setting and
    OOD Reynolds-number and angle-of-attack settings.
    Vol. and Surf. denote prediction errors for surrounding and
    surface physical fields, respectively.
    $C_L$ reports the lift-coefficient prediction error.
    $\rho_L$ denotes the Spearman rank correlation of $C_L$.
    The best and second-best results are \textbf{bold} and \underline{underlined},
    respectively.
  }
  \centering
  \small
  \setlength{\tabcolsep}{3pt}
  \renewcommand{\arraystretch}{1.05}
  \begin{tabular}{l|cccccccc}
    \toprule
    \multirow{2}{*}{Method}
    & \multicolumn{4}{c}{Standard Setting}
    & \multicolumn{2}{c}{OOD Reynolds}
    & \multicolumn{2}{c}{OOD Angles} \\
    \cmidrule(lr){2-5}
    \cmidrule(lr){6-7}
    \cmidrule(lr){8-9}
    & Vol.$\downarrow$ & Surf.$\downarrow$
    & $C_L\downarrow$ & $\rho_L\uparrow$
    & $C_L\downarrow$ & $\rho_L\uparrow$
    & $C_L\downarrow$ & $\rho_L\uparrow$ \\
    \midrule
    GNOT~\citep{gnot}
    & 0.0049 & 0.0152 & 0.1992 & 0.9942
    & 0.3268 & 0.9865 & 0.3497 & 0.9868 \\
    GINO~\citep{gino}
    & 0.0297 & 0.0482 & 0.1821 & 0.9958
    & 0.4180 & 0.9645 & 0.2583 & 0.9923 \\
    LNO~\citep{lno}
    & 0.0214 & 0.0268 & 0.1480 & 0.9744
    & 0.7440 & 0.9399 & 0.3732 & 0.9905 \\
    Transolver~\citep{transolver}
    & 0.0023 & 0.0085 & 0.1230 & 0.9978
    & 0.1622 & 0.9904 & 0.2438 & 0.9948 \\
    LinearNO~\citep{linearno}
    & 0.0011 & 0.0077 & 0.0491 & \underline{0.9992}
    & 0.2400 & 0.9951
    & 0.0987 & 0.9963 \\
    \midrule
    \textbf{PANO-P (ours)}
    & \textbf{0.0005} & \textbf{0.0010}
    & \underline{0.0236} & \textbf{0.9997}
    & \underline{0.1314} & \underline{0.9970}
    & \underline{0.0671} & \underline{0.9977} \\
    \textbf{PANO-D (ours)}
    & \underline{0.0009} & \underline{0.0031}
    & \textbf{0.0225} & \textbf{0.9997}
    & \textbf{0.0834} & \textbf{0.9985}
    & \textbf{0.0658} & \textbf{0.9980} \\
    \bottomrule
  \end{tabular}
  \label{tab:airfrans_bench}
\end{table*}

\begin{table*}[t]
  \caption{
    Comparison on Unilateral Stamping~\citep{Unisoma}
    in relative L2 error ($\downarrow$).
    The best and second-best results are \textbf{bold} and \underline{underlined},
    respectively.
  }
  \centering
  \small
  \setlength{\tabcolsep}{6pt}
  \renewcommand{\arraystretch}{1.05}
  \begin{tabular}{l|cccccc}
    \toprule
    \multirow{2}{*}{Method}
    & \multicolumn{2}{c}{Geometry}
    & \multicolumn{2}{c}{Stress}
    & \multicolumn{2}{c}{PEEQ} \\
    \cmidrule(lr){2-3}
    \cmidrule(lr){4-5}
    \cmidrule(lr){6-7}
    & Metal & Rubber & Metal & Rubber & Metal & Rubber \\
    \midrule
    OFormer~\citep{oformer}
    & 0.0119 & 0.0098 & 0.1103 & 0.0981 & 0.3269 & 0.2675 \\
    ONO~\citep{ono}
    & 0.0433 & 0.0437 & 0.1377 & 0.1311 & 0.5140 & 0.4748 \\
    MeshGraphNet~\citep{meshgraphnet}
    & 0.0391 & 0.0383 & 0.1252 & 0.1417 & 0.5805 & 0.5694 \\
    Transolver~\citep{transolver}
    & 0.0079 & 0.0075 & 0.0839 & 0.0862 & 0.2938 & 0.2693 \\
    Unisoma~\citep{Unisoma}
    & 0.0071 & 0.0063 & 0.0807 & 0.0773 & 0.2591 & 0.2114 \\
    \midrule
    \textbf{PANO-P (ours)}
    & \textbf{0.0052} & \textbf{0.0044}
    & \textbf{0.0743} & \textbf{0.0667}
    & \textbf{0.1827} & \textbf{0.1515} \\
    \textbf{PANO-D (ours)}
    & \underline{0.0067} & \underline{0.0058}
    & \underline{0.0756} & \underline{0.0710}
    & \underline{0.2248} & \underline{0.1822} \\
    \bottomrule
  \end{tabular}
  \label{tab:unilateral_stamping}
\end{table*}

\begin{wraptable}{r}{0.4\textwidth}
  \caption{
    Comparison on ShapeNetCar.
    Vol. and Surf. report prediction errors for the surrounding
    velocity field and surface pressure field, respectively.
  }
  \centering
  \small
  \setlength{\tabcolsep}{5pt}
  \renewcommand{\arraystretch}{1.05}
  \begin{tabular*}{\linewidth}{@{\extracolsep{\fill}}l|cc@{}}
    \toprule
    Method & Vol.$\downarrow$ & Surf.$\downarrow$ \\
    \midrule
    LNO          & 0.0269 & 0.0870 \\
    Transolver   & 0.0221 & 0.0785 \\
    Transolver++ & 0.0226 & 0.0800 \\
    LinearNO     & \underline{0.0194} & \textbf{0.0754} \\
    \midrule
    \textbf{PANO-P (ours)} & 0.0214 & 0.0817 \\
    \textbf{PANO-D (ours)} & \textbf{0.0187} & \underline{0.0784} \\
    \bottomrule
  \end{tabular*}
  \label{tab:shapenetcar_bench}
\end{wraptable}

Figure~\ref{fig:standard_benchmark_compare_preview} shows the results on standard benchmarks. PANO outperforms existing SOTA methods.
Specifically, compared with the best existing method on each task, PANO-D reduces relative L2 errors by 10.2\%, 63.6\%, 43.5\%, and 42.0\% on Airfoil, Plasticity, NS2D, and Elasticity, respectively.
PANO-P also outperforms existing methods on Airfoil, Plasticity, and Elasticity, while achieving performance comparable to LinearNO on NS2D.

\noindent \textbf{Evaluation at Multiple Observation Densities.}
Table~\ref{tab:multi_resolution_eval} compares methods on Darcy flow.
PANO-D and PANO-P achieve the best performance and the second-best performance at all resolutions, respectively.
Compared with the best existing method, PANO-P reduces errors by 20.8\% and 22.6\% on Darcy141 and Darcy211, respectively, while PANO-D achieves a 26.4\% reduction on both.
These results indicate that PANO can utilize denser observations more effectively.

\subsection{Advanced Benchmarks}
\noindent \textbf{Evaluation on AirfRANS.}
Table~\ref{tab:airfrans_bench} reports the results on AirfRANS.
PANO-P and PANO-D achieve the best or second-best results across all settings.

Under the standard setting, PANO-P reduces prediction errors for surrounding and surface physical fields by 54.5\% and 87.0\% (or more), respectively.
Compared with LinearNO, PANO-P and PANO-D reduce the prediction error of $C_L$ by approximately 51.9\% and 54.2\%, respectively. 
Moreover, both achieve the highest $\rho_L$ of 0.9997.
Under both OOD settings, PANO-D and PANO-P achieve the best and second-best lift-coefficient prediction results, respectively, highlighting the strong generalization ability of PANO.

\begin{wraptable}{r}{0.5\textwidth}
    \caption{
        Comparison of parameter counts, computational costs, and forward time.
    }
    \centering
    \small
    \setlength{\tabcolsep}{3pt}
    \renewcommand{\arraystretch}{1.05}
    \begin{tabular*}{\linewidth}{@{\extracolsep{\fill}}l|ccc@{}}
        \toprule
        Model & Params. (M) & GFLOPs & Forward (ms) \\
        \midrule
        Transolver   & 3.86 & 402.03 & 111.20 \\
        Transolver++ & 3.34 & \underline{356.36} & 127.19 \\
        LinearNO     & \underline{3.32} & 362.70 & 154.76 \\
        \midrule
        PANO-P       & \textbf{1.99} & \textbf{171.78} & \textbf{61.20} \\
        PANO-D       & 8.31 & 396.21 & \underline{75.81} \\
        \bottomrule
    \end{tabular*}
    \label{tab:param_flops_delay}
\end{wraptable}

\noindent \textbf{Evaluation on Unilateral Stamping.}
We evaluate predictions of deformed geometry, stress, and equivalent plastic strain (PEEQ) on Unilateral Stamping.
As shown in Table~\ref{tab:unilateral_stamping}, both PANO variants outperform existing methods across all six metrics.
Compared with Unisoma, which is specifically designed for multi-solid systems, PANO-P reduces geometry prediction errors by 26.8\% and 30.2\%, stress errors by 7.9\% and 13.7\%, and PEEQ errors by 29.5\% and 28.3\% for metal and rubber, respectively.
Notably, PANO-P outperforms PANO-D across all metrics.
This result suggests that providing more parallel aggregation patterns can be more beneficial than increasing model depth for representing multi-solid systems.

\noindent \textbf{Evaluation on ShapeNetCar.}
Table~\ref{tab:shapenetcar_bench} reports the results on ShapeNetCar.
PANO-P and PANO-D achieve comparable performance to existing methods.
PANO-D reduces the prediction error of the surrounding velocity field by 3.6\% compared with LinearNO. For surface pressure prediction, PANO-D performs only slightly worse than LinearNO and outperforms the other SOTA methods.

\begin{figure}[!t]
    \centering
    \includegraphics[width=0.95\linewidth]{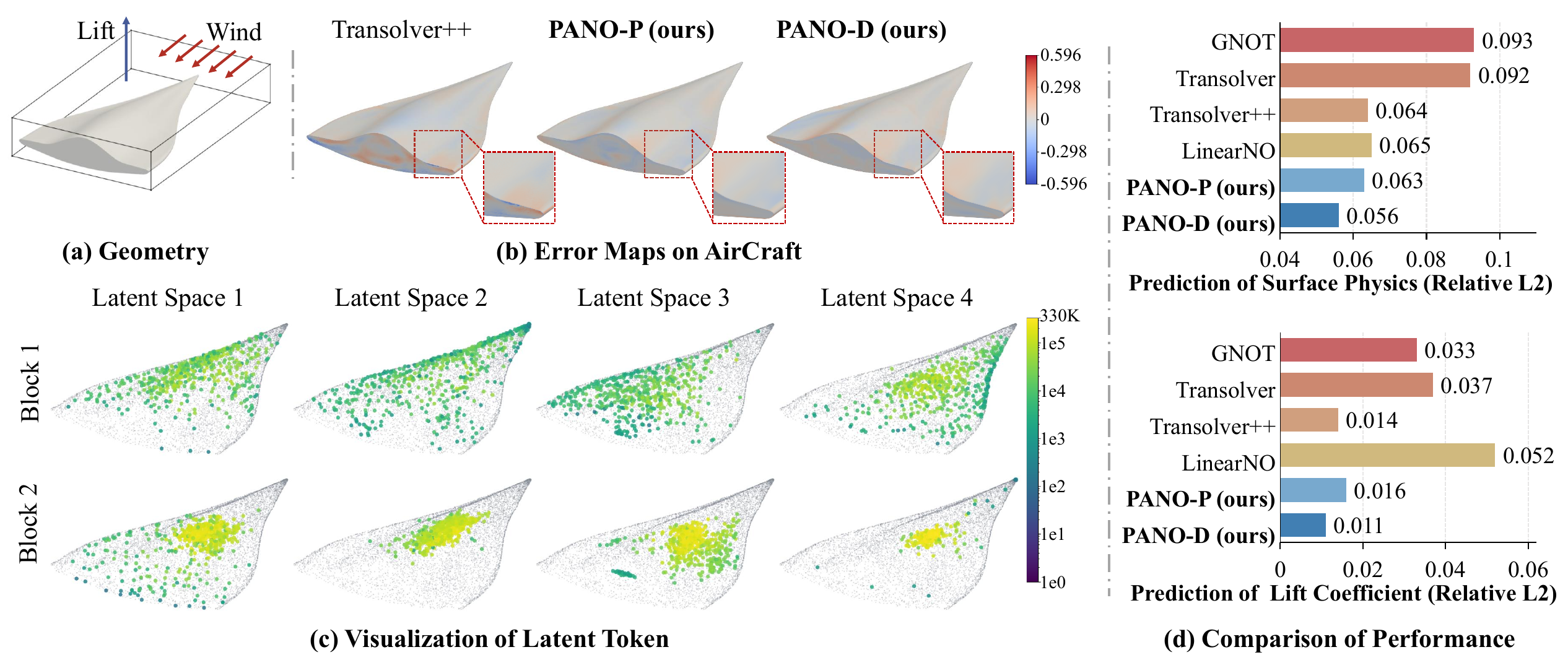}
    \caption{
    Prediction performance and latent-space representations on AirCraft.
    (a) AirCraft geometry.
    (b) Prediction error maps.
    (c) Visualization of latent tokens learned by PANO-P.
    We plot each latent token at its weighted center in the physical domain and color it according to the effective number of points it covers.
    (d) Relative L2 errors ($\downarrow$) for surface physical fields and lift coefficients.
    PANO outperforms existing methods.
    Moreover, PANO-P jointly represents physical states over the AirCraft through multiple latent spaces with distinct regional preferences at each stage.
    }
    \label{fig:AirCraft_vis}
\end{figure}

\noindent \textbf{Evaluation on AirCraft.}
We further evaluate PANO on AirCraft to assess the predictive performance with large-scale observations.
As shown in Figure~\ref{fig:AirCraft_vis}, PANO achieves SOTA performance.
Compared with Transolver++, PANO-D reduces prediction errors for surface physical fields and the lift coefficient by 12.5\% and 21.4\%, respectively.
These results demonstrate that PANO can scale to large-scale three-dimensional aerodynamic problems.
Figure~\ref{fig:AirCraft_vis} (b) further shows that PANO achieves lower prediction errors near AirCraft edges than Transolver++.
Figure~\ref{fig:AirCraft_vis} (c) visualizes the distributions of latent tokens from each latent space of PANO-P in the observation space.
Each latent space of PANO-P exhibits distinct regional preferences, indicating that PANO-P jointly represents physical states through multiple spatial aggregation patterns.

\subsection{Model Analysis}

We conduct comprehensive ablation studies of PANO-Block, CAP, and PaLA.
Unless otherwise specified, all variants use the depth and the feature dimension settings of PANO-P.

\noindent \textbf{Efficiency.}
We compare parameter counts and computational costs using a synthetic input with 100K observations.
For Transolver, Transolver++, and LinearNO, we use eight layers, eight attention heads, 256 feature channels, and 32 slices.
As shown in Table~\ref{tab:param_flops_delay}, PANO-P has the fewest parameters and lowest FLOPs among the compared models.
PANO-D has more parameters but comparable FLOPs to existing methods.
Notably, PANO-P and PANO-D substantially reduce the number of sequential latent-space construction operations, thereby achieving significantly higher inference efficiency.

\begin{figure}[!t]
    \centering
    \includegraphics[width=0.95\linewidth]{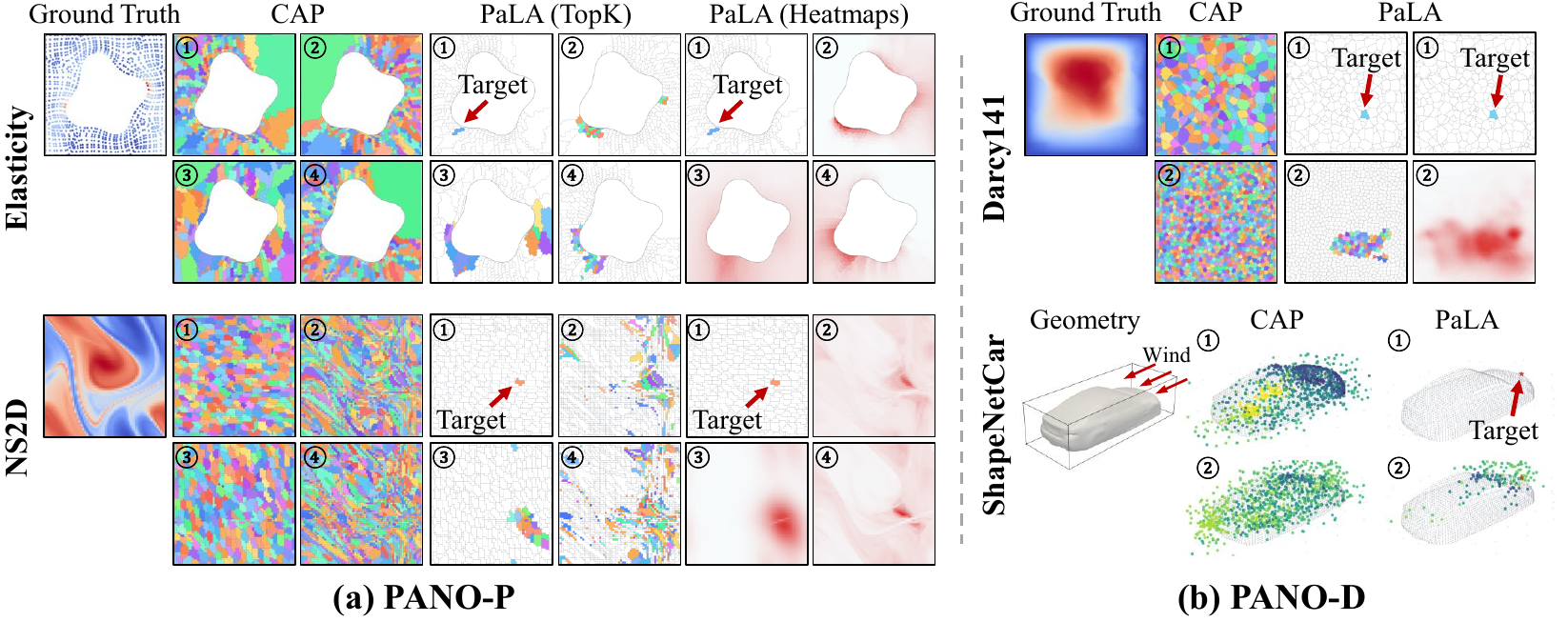}
    \caption{
    Visualization of CAP and PaLA with (a) PANO-P and (b) PANO-D.
    CAP is visualized through assignment maps for 2D tasks and latent-token spatial centers for 3D tasks.
    For PaLA, we select a target token in the first latent space and visualize its attention responses in the other spaces.
    Specifically, we highlight the Top$K$ tokens with the highest attention weights for the target token in each space.
    For 2D benchmarks, we additionally visualize the full response heatmaps of PaLA for the target token.
    Circled numbers (\textcircled{\scriptsize 1}, \textcircled{\scriptsize 2}, \textcircled{\scriptsize 3}, \textcircled{\scriptsize 4}) indicate the corresponding latent space.
    }
    \label{fig:cap_pala_vis}
\end{figure}

\begin{table*}[t]
  \caption{
    Ablation studies on PANO in relative L2 error ($\downarrow$) on Unilateral Stamping.
    (Left) Ablation of the number of parallel latent spaces $H$
    and the number of latent tokens per space $M$.
    (Right) Ablation of CAP and PaLA. The best and second-best results are \textbf{bold} and \underline{underlined},
    respectively.
  }
  \centering
    \begin{minipage}[t]{0.63\textwidth}
      \vspace{0pt}
      \centering
      \small
      \setlength{\tabcolsep}{3pt}
      \renewcommand{\arraystretch}{1.05}
      \resizebox{\linewidth}{!}{
      \begin{tabular}{ccccc|cccccc}
        \toprule
        \multirow{2}{*}{$HM$}
        & \multirow{2}{*}{$H$}
        & \multirow{2}{*}{$M$}
        & \multirow{2}{*}{$C$}
        & \multicolumn{1}{c|}{Params.}
        & \multicolumn{2}{c}{Geometry}
        & \multicolumn{2}{c}{Stress}
        & \multicolumn{2}{c}{PEEQ} \\
        \cmidrule(lr){6-7}
        \cmidrule(lr){8-9}
        \cmidrule(lr){10-11}
        & & & & (M) & Metal & Rubber & Metal & Rubber & Metal & Rubber \\
        \midrule
        512 & 1 & 512 & 96 & 1.26
        & 0.0066 & 0.0057
        & 0.0769 & 0.0713
        & 0.2256 & 0.1876 \\
    
        512 & 2 & 256 & 96 & 1.54
        & 0.0065 & 0.0056
        & 0.0772 & 0.0709
        & 0.2192 & 0.1814 \\
    
        512 & 4 & 128 & 96 & 1.69
        & 0.0064 & 0.0055
        & 0.0761 & 0.0710
        & 0.2224 & 0.1824 \\
        \midrule
        2048 & 1 & 2048 & 96 & 1.56
        & 0.0062 & \underline{0.0054}
        & 0.0765 & 0.0725
        & 0.2132 & 0.1805 \\
    
        2048 & 1 & 2048 & 192 & 5.42
        & 0.0064 & \underline{0.0054}
        & \underline{0.0750} & \underline{0.0691}
        & 0.2157 & \underline{0.1770} \\
    
        2048 & 2 & 1024 & 96 & 1.84
        & \underline{0.0060} & \underline{0.0054}
        & 0.0760 & 0.0700
        & \underline{0.2115} & 0.1778 \\
    
        2048 & 4 & 512 & 96 & 1.99
        & \textbf{0.0052} & \textbf{0.0044}
        & \textbf{0.0743} & \textbf{0.0667}
        & \textbf{0.1827} & \textbf{0.1515} \\
        \bottomrule
      \end{tabular}
      }
    \end{minipage}
  \hfill
    \begin{minipage}[t]{0.35\textwidth}
    \vspace{0pt}
    \centering
    \small
    \setlength{\tabcolsep}{3pt}
    \renewcommand{\arraystretch}{1.05}
    \resizebox{\linewidth}{!}{
    \begin{tabular}{cc|ccc}
      \toprule
      \multirow{2}{*}{CAP}
      & \multirow{2}{*}{PaLA}
      & \multicolumn{3}{c}{Stamping (Metal)} \\
      \cmidrule(lr){3-5}
      & & Geometry & Stress & PEEQ \\
      \midrule
      \ding{55} & \ding{55}
      & 0.0061 & 0.0759 & 0.2172 \\

      \ding{51} & \ding{55}
      & 0.0055 & \underline{0.0745} & 0.2029 \\

      \ding{55} & \ding{51}
      & \underline{0.0053}
      & \underline{0.0745}
      & \underline{0.1941} \\
      \midrule
      \ding{51} & \ding{51}
      & \textbf{0.0052}
      & \textbf{0.0743}
      & \textbf{0.1827} \\
      \bottomrule
    \end{tabular}
    }
  \end{minipage}
  \label{tab:ablation_hm_pano}
\end{table*}

\noindent \textbf{Parallel Latent Space Block.}
Table~\ref{tab:ablation_hm_pano} (left) reports the ablation of the number of parallel latent spaces $H$ and the number of latent tokens per space $M$.
Under both fixed total-token settings, $HM=512$ and $HM=2048$, introducing parallel latent spaces improves predictive performance.
At $HM=512$, compared with the model with a single latent space, the variant with four parallel latent spaces reduces PEEQ errors by 1.4\% and 2.8\% for metal and rubber, respectively.

Increasing the total number of latent tokens provides greater model capacity and generally improves performance. At $HM=2048$, the variant with four parallel latent spaces further reduces PEEQ errors by 14.3\% and 16.1\% for metal and rubber, respectively.
Moreover, under the same total-token setting of $HM=2048$, the model with four parallel latent spaces also outperforms the single-latent-space variant with a larger feature dimension.

Table~\ref{tab:latent_space_perturbation_stamping} reports the latent-space perturbation results. The results show that each latent space retains distinct and indispensable information that contributes to the final prediction.

\begin{table*}[t]
    \centering

    \begin{minipage}[t]{0.50\textwidth}
        \caption{
        Perturbation results on Unilateral Stamping in relative L2 error ($\downarrow$).
        Each numbered row removes the output of the corresponding latent space from Block 1 during output aggregation.
        '-' denotes the full model without perturbation.
        The best and second-best results are \textbf{bold} and \underline{underlined}, respectively.
        }
        \vspace{0pt}
        \centering
        \small
        \setlength{\tabcolsep}{3pt}
        \renewcommand{\arraystretch}{1.05}
        \resizebox{\linewidth}{!}{
        \begin{tabular}{c|cccccc}
            \toprule
            \multirow{2}{*}{Space ID}
            & \multicolumn{2}{c}{Geometry}
            & \multicolumn{2}{c}{Stress}
            & \multicolumn{2}{c}{PEEQ} \\
            \cmidrule(lr){2-3}
            \cmidrule(lr){4-5}
            \cmidrule(lr){6-7}
            & Metal & Rubber
            & Metal & Rubber
            & Metal & Rubber \\
            \midrule
            1
            & 0.0145 & 0.0136
            & 0.1111 & 0.1189
            & 0.4055 & 0.4764 \\

            2
            & \underline{0.0087} & \underline{0.0077}
            & \underline{0.0911} & \underline{0.0974}
            & \underline{0.2890} & \underline{0.2556} \\

            3
            & 0.0118 & 0.0103
            & 0.1076 & 0.1294
            & 0.3256 & 0.3431 \\

            4
            & 0.0166 & 0.0144
            & 0.1049 & 0.1638
            & 0.5391 & 0.5427 \\
            \midrule
            --
            & \textbf{0.0052} & \textbf{0.0044}
            & \textbf{0.0743} & \textbf{0.0667}
            & \textbf{0.1827} & \textbf{0.1515} \\
            \bottomrule
        \end{tabular}
        }
        \label{tab:latent_space_perturbation_stamping}
    \end{minipage}
    \hfill
    \begin{minipage}[t]{0.45\textwidth}
        \caption{
            Decoding-weight statistics on Airfoil.
            Scale is the maximum token decoding weight relative to the balanced value $N/M$.
            Top10 is the fraction contributed by the ten tokens with the largest weights.
            CV is the Coefficient of Variation across token decoding weights.
        }
        \vspace{0pt}
        \centering
        \small
        \setlength{\tabcolsep}{3pt}
        \renewcommand{\arraystretch}{1.05}
        \resizebox{\linewidth}{!}{
        \begin{tabular}{cc|ccc}
            \toprule
            \multirow{2}{*}{Method}
            & \multirow{2}{*}{Block ID}
            & \multicolumn{3}{c}{Decoding Weight} \\
            \cmidrule(lr){3-5}
            & & Scale & Top10 & CV \\
            \midrule
            Softmax & 1
            & 95.88$\times$
            & 62.71\%
            & 6.40 \\

            Softmax & 2
            & 26.35$\times$
            & 20.84\%
            & 1.91 \\

            \textbf{CAP (ours)} & 1
            & 1.00$\times$
            & 1.95\%
            & $\boldsymbol{\approx 0}$ \\

            \textbf{CAP (ours)} & 2
            & 1.00$\times$
            & 1.95\%
            & $\boldsymbol{\approx 0}$ \\
            \bottomrule
        \end{tabular}
        }
        \label{tab:cap_decoding_weight}
    \end{minipage}
\end{table*}

\noindent \textbf{CAP.}
Table~\ref{tab:ablation_hm_pano} (right) compares CAP with standard Softmax-based projection.
With PaLA disabled, introducing CAP reduces Geometry, Stress, and PEEQ errors by 9.8\%, 1.8\%, and 6.6\% for metal, respectively.
With PaLA enabled, CAP further reduces the three errors by 1.9\%, 0.3\%, and 5.9\%, respectively.
Table~\ref{tab:cap_decoding_weight} further reports decoding-weight statistics in terms of Scale, the Top10 total-weight fraction, and the coefficient of variation.
CAP substantially suppresses the extreme imbalance in token weights observed under standard Softmax-based projection.
These results show that CAP constructs more stable and effective parallel latent spaces.

\noindent \textbf{PaLA.}
Table~\ref{tab:ablation_hm_pano} (right) also reports the ablation of PaLA.
Compared with PANO using standard Softmax-based projection without PaLA, introducing PaLA reduces Geometry, Stress, and PEEQ errors by 13.1\%, 1.8\%, and 10.6\%, respectively.
Moreover, PaLA is complementary to CAP. With CAP enabled, PaLA further reduces the PEEQ error by 10.0\%.
These results demonstrate the importance of cross-space interactions.

\noindent \textbf{Visualizations.}
We visualize CAP assignments and PaLA cross-space attention responses across multiple benchmarks in Figure~\ref{fig:cap_pala_vis}.
CAP learns distinct spatial aggregation patterns across these benchmarks, allowing parallel latent spaces to represent the same input with different regional preferences and spatial granularities.
For PaLA, given a target token in latent space 1, high-weight responses in the other latent spaces exhibit structured spatial distributions that vary with their aggregation patterns.
These responses cover both regions near the target and more distant locations, showing that PaLA learns spatially structured cross-space token associations without relying on fixed token correspondences.

\section{Conclusion}
In this work, we proposed PANO, a novel neural operator for solving PDEs on general geometries.
PANO formulates physical-state updates of point tokens as an additive combination of interaction results from parallel latent spaces.
Within PANO, we propose CAP to construct bidirectionally consistent encoding and decoding mappings, and PaLA to jointly update latent representations through adaptive cross-space interactions.
Extensive experiments on nine benchmarks demonstrate the performance, OOD generalization, and scalability of PANO.
This work shows that the representational capacity of latent-space neural operators depends not only on latent modules but also on how input tokens are aggregated to latent tokens.
By constructing distinct representations in parallel latent spaces with shared parameters, PANO provides a parameter-efficient way to enrich the information available for latent interactions and alleviate the bottleneck of a single compressed representation.

\bibliography{iclr2027_conference}
\bibliographystyle{iclr2027_conference}

\clearpage
\appendix

\startcontents[appendix]

\section*{Appendix Contents}
\printcontents[appendix]{}{1}{\setcounter{tocdepth}{2}}

\clearpage

\section{Notation Summary}

Table~\ref{tab:notation_core} summarizes the notation used in the core formulation of PANO.
Table~\ref{tab:notation_theory} summarizes the notation introduced in the theoretical analysis in the main text.
Notation used only in the theoretical analysis in the appendix is defined upon first use.

\begin{table}[t]
    \centering
    \small
    \caption{Summary of notation used in the core formulation of PANO.}
    \setlength{\tabcolsep}{5pt}
    \renewcommand{\arraystretch}{1.12}
    \begin{tabular}{@{}p{0.30\textwidth}p{0.65\textwidth}@{}}
        \toprule
        \textbf{Notation} & \textbf{Description} \\
        \midrule

        \multicolumn{2}{@{}l}{\textbf{Domain, dimensions, and observations}} \\
        $\mathcal D$
        & Physical domain of the PDE problem. \\
        $N$
        & Number of observation points. \\
        $C_{\mathrm{in}}$
        & Input observation dimension at each point. \\
        $C$
        & Feature dimension of point tokens and latent tokens. \\
        $H$
        & Number of parallel latent spaces in each PANO-Block. \\
        $M$
        & Number of latent tokens per latent space. \\
        $i\in\{1,\ldots,N\}$
        & Observation-point index. \\
        $h,k\in\{1,\ldots,H\}$
        & Latent-space indices; $h\ne k$ for cross-space interactions. \\
        $\mathbf x_i\in\mathbb R^{C_{\mathrm{in}}}$
        & Input observation at the $i$-th spatial location. \\
        $\mathbf X\in\mathbb R^{N\times C_{\mathrm{in}}}$
        & Input observation matrix. \\

        \midrule
        \multicolumn{2}{@{}l}{\textbf{Parallel Latent Space Block}} \\
        $F\in\mathbb R^{N\times C}$
        & Input point tokens of a PANO-Block. \\
        $F_{\mathrm{latent}}\in\mathbb R^{N\times C}$
        & Decoded outputs averaged across parallel latent spaces. \\
        $\mathcal P_{\mathrm{enc}}^h$
        & CAP encoding mapping from point tokens to latent tokens in space $h$. \\
        $\mathcal P_{\mathrm{dec}}^h$
        & CAP decoding mapping from latent tokens in space $h$ to point tokens. \\
        $\mathcal M_{\mathrm{latent}}^h$
        & Latent module for space $h$. \\
        $\operatorname{LN}$
        & Layer normalization. \\
        $\mathrm{SelfAttn}$
        & Self-attention layer for interactions within a latent space. \\
        $\mathrm{FFN}$
        & Feed-forward network applied independently to each token. \\

        \midrule
        \multicolumn{2}{@{}l}{\textbf{Coupled Assignment Projection}} \\
        $\operatorname{Proj}^h$
        & Learnable assignment projector for space $h$. \\
        $R^h\in\mathbb R^{N\times M}$
        & Assignment weights after row-wise Softmax. \\
        $A^h\in\mathbb R^{N\times M}$
        & Assignment matrix shared by encoding and decoding in space $h$. \\
        $\mathbf 1_N,\mathbf 1_M$
        & All-ones vectors of lengths $N$ and $M$, respectively. \\
        $\operatorname{Softmax}_{\mathrm{row}}$
        & Softmax applied independently to each matrix row. \\
        $\operatorname{Diag}$
        & Operator that converts a vector into a diagonal matrix. \\

        \midrule
        \multicolumn{2}{@{}l}{\textbf{Pairwise Latent Space Attention Layer}} \\
        $\mathbf Z^h\in\mathbb R^{M\times C}$
        & Latent token sequence in space $h$ before the PaLA update. \\
        $\mathbf Z_{\mathrm{new}}^h\in\mathbb R^{M\times C}$
        & Latent token sequence after the PaLA attention update. \\
        $\mathbf W_{\mathrm{match}}\in\mathbb R^{C\times C}$
        & Matching projection matrix shared across latent spaces. \\
        $\mathbf W_{\mathrm{value}}^{PaLA}\in\mathbb R^{C\times C}$
        & PaLA value projection matrix shared across latent spaces. \\
        $\mathbf W_{\mathrm{value}}^{self}\in\mathbb R^{C\times C}$
        & Self-attention value projection matrix shared across latent spaces. \\
        $\mathbf W_{\mathrm{out}}\in\mathbb R^{C\times C}$
        & PaLA output projection matrix shared across latent spaces. \\
        $\mathbf S^{hk}\in\mathbb R^{M\times M}$
        & Matching scores between tokens in spaces $h$ and $k$. \\
        $\mathbf S^{k\rightarrow h}\in\mathbb R^{M\times M}$
        & Row-normalized attention weights for transferring information from source space $k$ to target space $h$. \\
        $\mathbf S^{h\rightarrow k}\in\mathbb R^{M\times M}$
        & Row-normalized attention weights for the reverse transfer, from space $h$ to space $k$. \\

        \bottomrule
    \end{tabular}
    \label{tab:notation_core}
\end{table}

\begin{table}[t]
    \centering
    \small
    \caption{Summary of notation introduced in the theoretical analysis in the main text.}
    \setlength{\tabcolsep}{5pt}
    \renewcommand{\arraystretch}{1.12}
    \begin{tabular}{@{}p{0.30\textwidth}p{0.65\textwidth}@{}}
        \toprule
        \textbf{Notation} & \textbf{Description} \\
        \midrule

        $\mu$
        & Probability measure on the physical domain $\mathcal D$. \\
        $\mathcal F:\mathcal D\to\mathbb R^C$
        & Bounded measurable point-token field. \\
        $x\in\mathcal D$
        & Spatial location at which the output field is evaluated. \\
        $\xi\in\mathcal D$
        & Integration variable over the physical domain. \\

        \midrule
        $\mathcal T_H$
        & Operator that averages the decoded outputs of $H$ parallel latent spaces using CAP and shared-parameter self-attention. \\
        $\kappa_{\mathcal F}^{h}(x,\xi)$
        & Input-dependent integral kernel induced by CAP encoding, self-attention, and CAP decoding in space $h$. \\
        $\overline{\kappa}_{\mathcal F}(x,\xi)$
        & Additive kernel obtained by averaging the kernels of parallel latent spaces. \\

        \midrule
        $\mathcal T_{\mathrm{cross}}$
        & Operator that averages the CAP-decoded cross-space updates produced by PaLA without shortcut. \\
        $\kappa_{\mathcal F}^{\mathrm{cross}}(x,\xi)$
        & Input-dependent cross-space kernel induced by CAP encoding, PaLA attention, and CAP decoding. \\

        \bottomrule
    \end{tabular}
    \label{tab:notation_theory}
\end{table}

\section{Proof of Theorem~\ref{thm:cap_additive_kernel}}
\label{sec:proof_of_theorem1}
\input{sec/theorem_1}

\section{Properties of Theorem~\ref{thm:cap_additive_kernel}}
\label{sec:properties_theorem1}
\input{sec/properties_theorem1}

\section{Proof of Theorem~\ref{thm:pala_cross_kernel}}
\label{sec:proof_of_theorem2}
\input{sec/theorem_2}

\section{Properties of Theorem~\ref{thm:pala_cross_kernel}}
\label{sec:properties_theorem2}
\input{sec/properties_theorem2}

\section{Detailed Results}
\label{sec:detailed_results}
\input{sec/detailed_results}

\section{More Comparison}
\label{sec:more_comparision}

We further compare PANO with MoNo~\citep{mono}, a recent latent-space neural operator that focuses on constructing stable multiscale latent representations.
MoNo progressively compresses physical states into a hierarchy of increasingly compact latent spaces, and identifies assignment imbalance and token collapse as key issues in this hierarchical construction.
To address these issues, MoNo introduces CoTAP, an optimal-transport-based projection that enforces balanced assignments between adjacent spaces, thereby stabilizing the progressive latent-space hierarchy.

MoNo is fundamentally different from PANO.
In particular, PANO focuses on the representational bottleneck introduced by compressing observations through a single latent space.
Since MoNo compresses the representation from the previous level through only a single latent space at each stage, it remains subject to the representational bottleneck induced by single-latent-space compression.
In particular, each deeper latent space in MoNo is constructed solely from the compressed representation retained by the preceding level.
Therefore, once some observation-space information is discarded during an earlier encoding stage, subsequent deeper latent spaces can only further process the information that has already been preserved, without directly recovering the omitted information.

Based on Theorems~\ref{thm:cap_additive_kernel} and~\ref{thm:pala_cross_kernel}, this limitation can be further formalized.
For the single latent-space encoding at the $l$-th level, let $\mathbf{A}^{(l)}$ denote its assignment matrix.
The ignored part $\mathbf{F}^{(l-1)}_{\mathrm{ignore}}$ satisfying:
\begin{equation}
    \left(\mathbf{A}^{(l)}\right)^{\top}\mathbf{F}^{(l-1)}_{\mathrm{ignore}} = 0
\end{equation}
is invisible to the latent representation constructed at that level.
Since subsequent deeper levels are constructed only from the already compressed representation, information omitted by the current encoding cannot re-enter the later latent representations.
Therefore, MoNo remains subject to the single-latent-space representational bottleneck characterized in this work.

The core projection method of MoNo, CoTAP, is also different from CAP in this work.
Specifically, CoTAP formulates the projection as a balanced optimal-transport problem and imposes marginal constraints on both sides of the transport plan, thereby enforcing balanced assignment mass for both the source and target representations.
In contrast, CAP does not impose such full bidirectional balancing constraints.
CAP normalizes the assignment along the latent-token dimension so that each latent token receives the same total assignment weight during encoding, and reuses the same assignment relations for encoding and decoding.
This construction is sufficient to maintain latent-space stability in terms of feature scale and weight allocation.

Therefore, CAP is naturally compatible with the parallel architecture of PANO.
Efficient construction of multiple parallel latent spaces requires the projection mechanism itself to remain lightweight.
CoTAP introduces Sinkhorn iterations to enforce balanced transport.
Replicating this procedure across multiple parallel branches incurs additional iterative overhead.
As shown in Table~\ref{tab:mono_comparison}, we further provide a comparison with MoNo.
PANO achieves better performance than MoNo across these benchmarks.

\begin{table*}[t]
    \caption{
        Comparison with MoNo on standard benchmarks and AirCraft.
        For the standard benchmarks, we compare the relative L2 error ($\downarrow$).
        For AirCraft, we report the complete evaluation metrics, including the relative L2 error of surface physical fields, denoted as Surf. ($\downarrow$), the lift-coefficient prediction error $C_L$ ($\downarrow$), and the coefficient of determination for lift prediction $R_L^2$ ($\uparrow$).
    }
    \centering
    \small
    \setlength{\tabcolsep}{4.2pt}
    \renewcommand{\arraystretch}{1.05}
    \resizebox{\textwidth}{!}{
    \begin{tabular}{lcccc|ccc|ccc}
        \toprule
        \multirow{2}{*}{Method}
        & \multicolumn{4}{c|}{Standard Benchmarks}
        & \multicolumn{3}{c|}{Darcy}
        & \multicolumn{3}{c}{AirCraft} \\
        \cmidrule(lr){2-5}
        \cmidrule(lr){6-8}
        \cmidrule(lr){9-11}
        &
        Airfoil$\downarrow$
        & Plasticity$\downarrow$
        & NS2D$\downarrow$
        & Elasticity$\downarrow$
        & 85$\downarrow$
        & 141$\downarrow$
        & 211$\downarrow$
        & Surf.$\downarrow$
        & $C_L\downarrow$
        & $R_L^2\uparrow$ \\
        \midrule

        MoNo~\citep{mono}
        & 0.0048
        & 0.0006
        & 0.0522
        & 0.0033
        & 0.0059
        & 0.0049
        & 0.0046
        & 0.077
        & 0.016
        & 0.999 \\

        \midrule

        \textbf{PANO-P (ours)}
        & \textbf{0.0044}
        & 0.0006
        & 0.0720
        & 0.0040
        & 0.0055
        & 0.0042
        & 0.0041
        & 0.063
        & 0.016
        & 1.000 (0.9995) \\

        \textbf{PANO-D (ours)}
        & \textbf{0.0044}
        & \textbf{0.0004}
        & \textbf{0.0395}
        & \textbf{0.0029}
        & \textbf{0.0050}
        & \textbf{0.0039}
        & \textbf{0.0039}
        & \textbf{0.056}
        & \textbf{0.011}
        & \textbf{1.000 (0.9996)} \\

        \bottomrule
    \end{tabular}}
    \label{tab:mono_comparison}
\end{table*}

\section{Benchmarks}
\label{sec:benchmark_intro}
\input{sec/dataset_intro}

\section{Implementation Details}
\label{sec:implementation_details}
\input{sec/implementation_details}

\section{Showcases}
\label{sec:showcases}
\input{sec/showcase}

\end{document}

%% file: math_commands.tex
\usepackage{amsmath,amsfonts,bm}

\def\eqref#1{equation~\ref{#1}}

\def\1{\bm{1}}

\DeclareMathAlphabet{\mathsfit}{\encodingdefault}{\sfdefault}{m}{sl}
\SetMathAlphabet{\mathsfit}{bold}{\encodingdefault}{\sfdefault}{bx}{n}



%% file: sec/theorem_1.tex
To focus on the kernel representation, the following analysis considers the single-head attention core, omitting LayerNorm, residual connections, and the FFN.

\subsection{The Integral Kernel Representation for a Single Latent Space}

We first consider the special case $H=1$, consisting of CAP encoding, a single self-attention layer, and CAP decoding.
Following the proof method of prior work~\citep{transolver, linearno, mono}, we show that this composition can be represented as an integral operator.

\begin{theorem}[\textbf{The Integral Kernel Representation of CAP with a Single Self-Attention Layer}]
\label{thm:cap_integral_modified}
Let $(\mathcal D,\mu)$ be the original physical domain equipped with a probability measure $\mu$.
For any bounded measurable point-token field $\mathcal F:\mathcal D\to\mathbb R^C$, assuming finite and measurable CAP assignment logits, the composition $\mathcal T$ of CAP encoding, a single-head latent self-attention layer, and CAP decoding can be represented as an input-dependent integral-kernel operator on the original physical domain:
\begin{equation}
    \mathcal T(\mathcal F)(x)
    =
    \int_{\mathcal D}
    \kappa_{\mathcal F}(x,\xi)\mathcal F(\xi)\mathbf W_{\mathrm{value}}^{self}
    \,d\mu(\xi),
    \qquad \mu\text{-a.e. }x\in\mathcal D,
    \label{eq:cap_integral_modified}
\end{equation}
where $x\in\mathcal D$ denotes the spatial location at which the output is evaluated, $\xi\in\mathcal D$ is the integration variable, $\mathbf W_{\mathrm{value}}^{self}$ is the value projection matrix, and $\kappa_{\mathcal F}$ is the input-dependent kernel jointly induced by CAP and latent self-attention.
\end{theorem}

\begin{proof}
Let $a_{\mathcal F}(x,j)$ denote the continuous CAP encoding density with respect to $\mu$ induced by the input field $\mathcal F$, where $x\in\mathcal D$ is a spatial location, $j\in\{1,\ldots,M\}$ indexes the latent tokens, and $M$ is the number of latent tokens.
Specifically, let $r_{\mathcal F}(x,j)$ denote the assignment weight obtained by row-wise Softmax. The continuous CAP density is defined as:
\begin{equation}
    a_{\mathcal F}(x,j)
    =
    \frac{r_{\mathcal F}(x,j)}
    {\int_{\mathcal D}r_{\mathcal F}(\xi,j)\,d\mu(\xi)}.
\end{equation}

CAP encoding aggregates the input field into a latent token sequence $\mathbf Z_{\mathrm{encode}}$ through weighted integration, with its $j$-th token given by:
\begin{equation}
    \mathbf Z_{\mathrm{encode}}(j)
    =
    \int_{\mathcal D}
    a_{\mathcal F}(\xi,j)\mathcal F(\xi)\,d\mu(\xi).
    \label{eq:cap_continuous_encoding}
\end{equation}
Let $\mathbf Z_{\mathrm{decode}}$ denote the latent token sequence after latent interactions and before decoding. CAP maps this sequence back to the original physical domain through the same assignment relations:
\begin{equation}
    \mathcal T(\mathcal F)(x)
    =
    \frac{1}{M}\sum_{j=1}^{M}
    a_{\mathcal F}(x,j)\mathbf Z_{\mathrm{decode}}(j).
    \label{eq:cap_continuous_decoding}
\end{equation}
Here, $\mathbf Z_{\mathrm{encode}}(j),\mathbf Z_{\mathrm{decode}}(j)\in\mathbb R^C$ are the representations of the $j$-th latent token before and after latent interactions, respectively, and $1/M$ is the CAP decoding normalization factor under the probability measure.

Let $\mathbf B_{\mathcal F}\in\mathbb R^{M\times M}$ be the self-attention weight matrix computed from $\mathbf Z_{\mathrm{encode}}$, where $(\mathbf B_{\mathcal F})_{jk}$ denotes the attention weight assigned by the $j$-th latent token to the $k$-th latent token.
A single self-attention layer updates the encoded sequence as $\mathbf Z_{\mathrm{decode}}=\mathbf B_{\mathcal F}\mathbf Z_{\mathrm{encode}}\mathbf W_{\mathrm{value}}^{self}$.
Substituting the self-attention update and the CAP encoding expression successively into Equation~\eqref{eq:cap_continuous_decoding} gives:
\begin{equation}
\begin{aligned}
    \mathcal T(\mathcal F)(x)
    &=
    \frac{1}{M}\sum_{j=1}^{M}
    a_{\mathcal F}(x,j)\mathbf Z_{\mathrm{decode}}(j)
    \\
    &=
    \frac{1}{M}\sum_{j=1}^{M}
    a_{\mathcal F}(x,j)
    \sum_{k=1}^{M}
    (\mathbf B_{\mathcal F})_{jk}
    \mathbf Z_{\mathrm{encode}}(k)\mathbf W_{\mathrm{value}}^{self}
    \\
    &=
    \frac{1}{M}\sum_{j,k=1}^{M}
    a_{\mathcal F}(x,j)(\mathbf B_{\mathcal F})_{jk}
    \int_{\mathcal D}
    a_{\mathcal F}(\xi,k)\mathcal F(\xi)\mathbf W_{\mathrm{value}}^{self}
    \,d\mu(\xi).
\end{aligned}
\label{eq:cap_continuous_expansion}
\end{equation}
Finite and measurable assignment logits ensure that the CAP column-normalization denominators are positive, so $a_{\mathcal F}(x,j)$ is finite.
In addition,
by the CAP normalization, $a_{\mathcal F}(\xi,j)$ is nonnegative and integrates to one over the physical domain for each latent token:
\begin{equation}
    a_{\mathcal F}(x,j)\geq 0,
    \qquad
    \int_{\mathcal D}a_{\mathcal F}(\xi,j)\,d\mu(\xi)=1,
    \qquad j=1,\ldots,M.
    \label{eq:cap_density_normalization_modified}
\end{equation}
Since $\mathcal F$ is bounded and the attention weights are nonnegative with $\sum_k(\mathbf B_{\mathcal F})_{jk}=1$, we have:
\begin{equation}
\begin{aligned}
    &\frac{1}{M}\sum_{j,k=1}^{M}
    a_{\mathcal F}(x,j)(\mathbf B_{\mathcal F})_{jk}
    \int_{\mathcal D}
    a_{\mathcal F}(\xi,k)
    \|\mathcal F(\xi)\mathbf W_{\mathrm{value}}^{self}\|\,d\mu(\xi)
    \\
    &\qquad\leq
    \frac{1}{M}\sum_{j=1}^{M}a_{\mathcal F}(x,j)
    \|\mathcal F\|_{L^\infty(\mu)}\|\mathbf W_{\mathrm{value}}^{self}\|
    <\infty,
\end{aligned}
\label{eq:cap_integrability_modified}
\end{equation}
where $\|\mathbf W_{\mathrm{value}}^{self}\|$ denotes the operator norm induced by the vector norm.

Therefore, we can exchange the finite sums with the integral to obtain:
\begin{equation}
    \mathcal T(\mathcal F)(x)
    =
    \int_{\mathcal D}
    \underbrace{
    \left[
    \frac{1}{M}\sum_{j,k=1}^{M}
    a_{\mathcal F}(x,j)(\mathbf B_{\mathcal F})_{jk}
    a_{\mathcal F}(\xi,k)
    \right]
    }_{\displaystyle \kappa_{\mathcal F}(x,\xi)}
    \mathcal F(\xi)\mathbf W_{\mathrm{value}}^{self}\,d\mu(\xi).
    \label{eq:cap_continuous_kernel}
\end{equation}
This establishes the desired input-dependent integral-kernel representation.
\end{proof}

\subsection{The Additive Kernel Representation for Parallel Latent Spaces}

Building on Theorem~\ref{thm:cap_integral_modified}, we consider $H\geq1$ parallel latent spaces with independent CAP mappings and shared attention projection parameters.
The proof of Theorem~\ref{thm:cap_additive_kernel} is as follows.

\begin{proof}
Let $\mathcal T^h$ denote the composition of CAP encoding, self-attention, and CAP decoding in the $h$-th latent space.
By Theorem~\ref{thm:cap_integral_modified}, averaging the outputs of the latent spaces gives:
\begin{equation}
\begin{aligned}
    \mathcal T_H(\mathcal F)(x)
    &=
    \frac{1}{H}\sum_{h=1}^{H}\mathcal T^h(\mathcal F)(x)
    \\
    &=
    \frac{1}{H}\sum_{h=1}^{H}
    \int_{\mathcal D}
    \kappa_{\mathcal F}^{h}(x,\xi)
    \mathcal F(\xi)\mathbf W_{\mathrm{value}}^{self}\,d\mu(\xi).
\end{aligned}
\label{eq:cap_parallel_expansion}
\end{equation}

The single-space proof establishes the absolute integrability of each integrand. Since $H$ is finite, these representations hold on a common set of full measure and satisfy:
\begin{equation}
    \frac{1}{H}\sum_{h=1}^{H}
    \int_{\mathcal D}
    \left\|
    \kappa_{\mathcal F}^{h}(x,\xi)
    \mathcal F(\xi)\mathbf W_{\mathrm{value}}^{self}
    \right\|\,d\mu(\xi)
    <\infty,
    \qquad \mu\text{-a.e. }x\in\mathcal D.
    \label{eq:cap_parallel_integrability}
\end{equation}
Therefore, we can exchange the finite sum with the integral and combine the individual kernels into an additive kernel:
\begin{equation}
    \mathcal T_H(\mathcal F)(x)
    =
    \int_{\mathcal D}
    \underbrace{
    \left[
    \frac{1}{H}\sum_{h=1}^{H}\kappa_{\mathcal F}^{h}(x,\xi)
    \right]
    }_{\displaystyle \overline{\kappa}_{\mathcal F}(x,\xi)}
    \mathcal F(\xi)\mathbf W_{\mathrm{value}}^{self}\,d\mu(\xi).
    \label{eq:cap_parallel_kernel}
\end{equation}
When $H=1$, $\overline{\kappa}_{\mathcal F}=\kappa_{\mathcal F}^{1}$, so the single-space case is naturally included in this representation.
\end{proof}

%% file: sec/properties_theorem1.tex
We further use the discrete kernel representation of Theorem~\ref{thm:cap_additive_kernel} to analyze two properties of parallel latent spaces: 1) a larger kernel rank bound and 2) a smaller common encoding null space.

\subsection{Discrete Kernel Representation of Parallel Latent Spaces}

We provide the discrete representation of Theorem~\ref{thm:cap_additive_kernel} to connect it with the forward computation of parallel CAP encoding, latent self-attention, and decoding fusion in PANO.
Let the $i$-th row of the discrete input matrix $F$ be $\mathcal F(x_i)$ at observation location $x_i$, and use the CAP assignment matrix $A^h$ defined in the main text.
CAP encodes the input as $\mathbf Z_{\mathrm{encode}}^h=(A^h)^\top F$, after which self-attention produces the sequence for decoding:
\begin{equation}
    \mathbf Z_{\mathrm{decode}}^h
    =
    \mathbf B_F^h\mathbf Z_{\mathrm{encode}}^h\mathbf W_{\mathrm{value}}^{self},
    \qquad
    \mathbf B_F^h
    =
    \operatorname{Softmax}_{\mathrm{row}}
    \left(
    \frac{
    (\mathbf Z_{\mathrm{encode}}^h\mathbf W_q)
    (\mathbf Z_{\mathrm{encode}}^h\mathbf W_k)^\top
    }{\sqrt{d_a}}
    \right),
    \label{eq:cap_parallel_discrete_attention}
\end{equation}
where $\mathbf B_F^h\in\mathbb R^{M\times M}$ is the attention weight matrix of the $h$-th latent space, $\mathbf W_q,\mathbf W_k\in\mathbb R^{C\times d_a}$ and $\mathbf W_{\mathrm{value}}^{self}\in\mathbb R^{C\times C}$ are the shared query, key, and value projection matrices, and $d_a$ is the query and key feature dimension.

Applying CAP decoding to the updated sequences and averaging across spaces gives:
\begin{equation}
\begin{aligned}
    F_{\mathrm{latent}}
    &=
    \frac{1}{H}\sum_{h=1}^{H}
    \mathcal P_{\mathrm{dec}}^h
    \left(\mathbf Z_{\mathrm{decode}}^h\right)
    \\
    &=
    \frac{1}{H}\sum_{h=1}^{H}
    \frac{N}{M}A^h
    \mathbf B_F^h\mathbf Z_{\mathrm{encode}}^h\mathbf W_{\mathrm{value}}^{self}
    \\
    &=
    \frac{1}{H}\sum_{h=1}^{H}
    \frac{N}{M}A^h\mathbf B_F^h(A^h)^\top
    F\mathbf W_{\mathrm{value}}^{self}
    \\
    &=
    \underbrace{
    \left[
    \frac{N}{HM}\sum_{h=1}^{H}
    A^h\mathbf B_F^h(A^h)^\top
    \right]
    }_{\displaystyle \overline{\mathbf K}_F}
    F\mathbf W_{\mathrm{value}}^{self}.
\end{aligned}
\label{eq:cap_parallel_discrete_expansion}
\end{equation}
Thus, the discrete additive kernel matrix acting on the observation space is:
\begin{equation}
    \overline{\mathbf K}_F
    =
    \frac{1}{H}\sum_{h=1}^{H}\mathbf K_F^h,
    \qquad
    \mathbf K_F^h
    =
    \frac{N}{M}A^h\mathbf B_F^h(A^h)^\top,
    \qquad
    \overline{\mathbf K}_F,\mathbf K_F^h\in\mathbb R^{N\times N}.
    \label{eq:cap_parallel_discrete_kernel}
\end{equation}
Equation~(\ref{eq:cap_parallel_discrete_kernel}) is an exact algebraic identity for finite $N,M,H$, with input dependence arising from the assignment matrices and attention weights in each space.

\subsection{Expanded Kernel Rank Bound}
\label{sec:kernel_rank_bound}

For a given input $F$, let $A_{\oplus}=[A^1,\ldots,A^H]\in\mathbb R^{N\times HM}$ be the horizontal concatenation of the assignment matrices.
The discrete additive kernel of Equation~(\ref{eq:cap_parallel_discrete_expansion}) can be written as:
\begin{equation}
    \overline{\mathbf K}_F
    =
    \frac{N}{HM}
    A_{\oplus}
    \operatorname{BlockDiag}
    \left(\mathbf B_F^1,\ldots,\mathbf B_F^H\right)
    A_{\oplus}^{\top},
    \label{eq:cap_kernel_block_factorization}
\end{equation}
where $\operatorname{BlockDiag}$ arranges the attention matrices into a block-diagonal matrix.
Since the rank of a matrix product cannot exceed the rank of any factor, we have:
\begin{equation}
    \operatorname{rank}(\mathbf K_F^h)
    \leq \operatorname{rank}(A^h)
    \leq \min(N,M),
    \qquad
    \operatorname{rank}(\overline{\mathbf K}_F)
    \leq \operatorname{rank}(A_{\oplus})
    \leq \min(N,HM).
    \label{eq:cap_kernel_rank_bounds}
\end{equation}
Thus, parallel construction expands the allowable rank bound of the spatial kernel matrix from $\min(N,M)$ to $\min(N,HM)$.
This analysis shows that parallel latent spaces can expand the representational capacity of the model while sharing parameters.

\subsection{Reduced Common Encoding Null Space}
\label{sec:common_encoding_nullspace}

\noindent \textbf{Fixed-Assignment Perturbation Analysis.}
To illustrate how information can be omitted during encoding, we consider a perturbation $\Delta F\in\mathbb R^{N\times C}$ to the input point tokens while holding the assignment matrices fixed.
For the $h$-th latent space, the difference between the encoded representations before and after the perturbation is:
\begin{equation}
    (A^h)^\top(F+\Delta F)-(A^h)^\top F
    =
    (A^h)^\top\Delta F.
    \label{eq:cap_encoding_perturbation}
\end{equation}
When a nonzero perturbation satisfies $(A^h)^\top\Delta F=0$, the input changes while its encoded representation remains unchanged, making the information carried by this perturbation indistinguishable to that latent space.
We characterize these invisible perturbations through the null spaces of the encoding mappings to analyze the information jointly omitted by multiple latent spaces.

Since the condition applies column-wise to $\Delta F$, we analyze a spatial perturbation direction $u\in\mathbb R^N$ in a single feature channel.
Specifically, for a fixed assignment matrix, the spatial information directions omitted by the $h$-th encoding form the null space $\mathcal U^h=\ker((A^h)^\top)$.
We further consider such a direction that is invisible to all latent spaces:
\begin{equation}
    A_{\oplus}^{\top}u=0
    \quad\Longleftrightarrow\quad
    (A^h)^\top u=0
    \quad\text{for all }h=1,\ldots,H.
    \label{eq:cap_joint_invisibility}
\end{equation}
Thus, the directions jointly omitted by all parallel latent spaces form the intersection of their null spaces:
\begin{equation}
    \mathcal U_{\mathrm{common}}
    =
    \bigcap_{h=1}^{H}\mathcal U^h
    =
    \ker\left(A_{\oplus}^{\top}\right).
    \label{eq:cap_common_nullspace}
\end{equation}
Therefore, the common null space is contained in the encoding null space of every latent space:
\begin{equation}
    \mathcal U_{\mathrm{common}}
    \subseteq \mathcal U^h,
    \qquad h=1,\ldots,H.
    \label{eq:cap_nullspace_inclusion}
\end{equation}
This shows that parallel latent spaces constructed through distinct mappings can reduce jointly omitted information, allowing information missed by one latent space to be captured by others.

\noindent \textbf{Actual Input-Dependent Encoding.}
We further consider the information omitted during actual encoding.
During an actual forward pass, the assignment matrices are also constructed from the input features $F$.
Based on this realized assignment matrix, we define the orthogonal projector onto its column space as:
\begin{equation}
    P^h
    =
    A^h(A^h)^\dagger,
    \label{eq:actual_encoding_projector}
\end{equation}
where $\dagger$ denotes the Moore--Penrose pseudoinverse.
The actual input features can then be decomposed into the components retained and omitted by this latent-space encoding:
\begin{equation}
    F_{\mathrm{keep}}^h=P^hF,
    \qquad
    F_{\mathrm{ignore}}^h=(I_N-P^h)F,
    \qquad
    F=F_{\mathrm{keep}}^h+F_{\mathrm{ignore}}^h.
    \label{eq:actual_feature_decomposition}
\end{equation}

By construction, the omitted component satisfies:
\begin{equation}
    (A^h)^\top F_{\mathrm{ignore}}^h=0,
    \qquad
    (A^h)^\top F
    =
    (A^h)^\top F_{\mathrm{keep}}^h.
    \label{eq:actual_ignored_component}
\end{equation}
Therefore, although the assignment matrix is adaptively constructed from $F$, the component $F_{\mathrm{ignore}}^h$ is invisible to aggregation under that assignment.
For parallel latent spaces, we use the concatenated assignment matrix $A_{\oplus}$ to construct the joint projector and obtain the input component jointly omitted by all latent spaces:
\begin{equation}
    P_{\oplus}
    =
    A_{\oplus}A_{\oplus}^{\dagger},
    \qquad
    F_{\mathrm{ignore}}^{\oplus}
    =
    (I_N-P_{\oplus})F.
    \label{eq:actual_joint_projector}
\end{equation}
Since the column space of each $A^h$ is contained in that of $A_{\oplus}$, the jointly omitted component is invisible to the encoding of every latent space:
\begin{equation}
    (A^h)^\top F_{\mathrm{ignore}}^{\oplus}=0,
    \qquad
    h=1,\ldots,H.
    \label{eq:actual_joint_invisibility}
\end{equation}

\noindent \textbf{Quantitative Evaluation.}
We measure the information omitted during actual forward passes of PANO-P on Airfoil and quantify it with $\left\|F_{\mathrm{ignore}}^h\right\|_F^2/\left\|F\right\|_F^2$.
As shown in Table~\ref{tab:actual_encoding_ignored_fraction}, each latent space has a distinct and nonzero omitted component, whereas the information jointly omitted by all latent spaces is substantially smaller than that omitted by any individual space.

\begin{table}[t]
    \caption{
        Fractions of actual input features omitted by each latent space and jointly by all latent spaces of PANO-P on Airfoil.
    }
    \centering
    \small
    \setlength{\tabcolsep}{5pt}
    \renewcommand{\arraystretch}{1.05}
    \begin{tabular}{lccccc}
        \toprule
        Block
        & Space 1
        & Space 2
        & Space 3
        & Space 4
        & Jointly \\
        \midrule
        Block 1
        & 7.88\%
        & 0.09\%
        & 0.55\%
        & 10.22\%
        & 0.000014\% \\
        Block 2
        & 0.11\%
        & 0.66\%
        & 0.18\%
        & 0.69\%
        & 0.000083\% \\
        \bottomrule
    \end{tabular}
    \label{tab:actual_encoding_ignored_fraction}
\end{table}

%% file: sec/theorem_2.tex
We further consider the composition of CAP encoding, a single PaLA layer, and CAP decoding for $H>1$ to characterize the cross-space kernels introduced by interactions between latent spaces.
As in the analysis of Theorem~\ref{thm:cap_additive_kernel}, we focus on the single-head attention core, omitting LayerNorm, residual connections, and the FFN.
The proof of Theorem~\ref{thm:pala_cross_kernel} is as follows.

\begin{proof}
Following the notation of Theorem~\ref{thm:cap_additive_kernel}, let $a_{\mathcal F}^h(x,j)$ denote the continuous CAP encoding density of the $h$-th latent space, which is nonnegative and satisfies $\int_{\mathcal D}a_{\mathcal F}^h(\xi,j)\,d\mu(\xi)=1$.
CAP encodes the input field into latent token sequences in each space:
\begin{equation}
    \mathbf Z_{\mathrm{encode}}^h(j)
    =
    \int_{\mathcal D}
    a_{\mathcal F}^h(\xi,j)\mathcal F(\xi)\,d\mu(\xi),
    \qquad h=1,\ldots,H.
    \label{eq:pala_continuous_encoding}
\end{equation}

Following the PaLA construction in the main text, the attention weights for information transfer from space $k$ to space $h$ are:
\begin{equation}
    \mathbf S^{k\rightarrow h}
    =
    \operatorname{Softmax}_{\mathrm{row}}
    \left(
    \frac{
    (\mathbf Z_{\mathrm{encode}}^h\mathbf W_{\mathrm{match}})
    (\mathbf Z_{\mathrm{encode}}^k\mathbf W_{\mathrm{match}})^\top
    }{\sqrt C}
    \right),
    \qquad k\ne h,
    \label{eq:pala_continuous_weights}
\end{equation}
where $\mathbf W_{\mathrm{match}}$ is the shared matching projection matrix and $(\mathbf S^{k\rightarrow h})_{j\ell}$ is the attention weight assigned by the $j$-th token in target space $h$ to the $\ell$-th token in source space $k$.
With the residual connection omitted, PaLA averages incoming information from the other spaces and applies the output projection to obtain the sequence for decoding:
\begin{equation}
    \mathbf Z_{\mathrm{decode}}^h
    =
    \frac{1}{H-1}\sum_{k\ne h}
    \mathbf S^{k\rightarrow h}
    \mathbf Z_{\mathrm{encode}}^k
    \mathbf{W}_{\mathrm{value}}^{PaLA}\mathbf W_{\mathrm{out}}.
    \label{eq:pala_continuous_update}
\end{equation}

Applying CAP decoding and averaging across spaces, then successively substituting the PaLA update and encoding expressions, gives:
\begin{equation}
\begin{aligned}
    \mathcal T_{\mathrm{cross}}(\mathcal F)(x)
    &=
    \frac{1}{HM}\sum_{h=1}^{H}\sum_{j=1}^{M}
    a_{\mathcal F}^h(x,j)\mathbf Z_{\mathrm{decode}}^h(j)
    \\
    &=
    \frac{1}{HM(H-1)}
    \sum_{h=1}^{H}\sum_{k\ne h}\sum_{j,\ell=1}^{M}
    a_{\mathcal F}^h(x,j)
    (\mathbf S^{k\rightarrow h})_{j\ell}
    \mathbf Z_{\mathrm{encode}}^k(\ell)
    \mathbf{W}_{\mathrm{value}}^{PaLA}\mathbf W_{\mathrm{out}}
    \\
    &=
    \frac{1}{HM(H-1)}
    \sum_{h=1}^{H}\sum_{k\ne h}\sum_{j,\ell=1}^{M}
    a_{\mathcal F}^h(x,j)
    (\mathbf S^{k\rightarrow h})_{j\ell}
    \int_{\mathcal D}
    a_{\mathcal F}^k(\xi,\ell)\mathcal F(\xi)
    \mathbf{W}_{\mathrm{value}}^{PaLA}\mathbf W_{\mathrm{out}}
    \,d\mu(\xi).
\end{aligned}
\label{eq:pala_continuous_expansion}
\end{equation}

By the normalization of the CAP densities, the boundedness of the input field, and the nonnegativity and row normalization of the attention weights, we have:
\begin{equation}
\begin{aligned}
    &\frac{1}{HM(H-1)}
    \sum_{h=1}^{H}\sum_{k\ne h}\sum_{j,\ell=1}^{M}
    a_{\mathcal F}^h(x,j)
    (\mathbf S^{k\rightarrow h})_{j\ell}
    \int_{\mathcal D}
    a_{\mathcal F}^k(\xi,\ell)
    \|\mathcal F(\xi)\mathbf{W}_{\mathrm{value}}^{PaLA}\mathbf W_{\mathrm{out}}\|
    \,d\mu(\xi)
    \\
    &\qquad\leq
    \frac{1}{HM}\sum_{h=1}^{H}\sum_{j=1}^{M}
    a_{\mathcal F}^h(x,j)
    \|\mathcal F\|_{L^\infty(\mu)}
    \|\mathbf{W}_{\mathrm{value}}^{PaLA}\|
    \|\mathbf W_{\mathrm{out}}\|
    <\infty,
\end{aligned}
\label{eq:pala_cross_integrability}
\end{equation}
where $\|\mathbf{W}_{\mathrm{value}}^{PaLA}\|$ and $\|\mathbf W_{\mathrm{out}}\|$ denote the operator norms of the value and output projection matrices induced by the vector norm, respectively.
Therefore, we can exchange the finite sums with the integral to obtain:
\begin{equation}
\begin{aligned}
    \mathcal T_{\mathrm{cross}}(\mathcal F)(x)
    &=
    \int_{\mathcal D}
    \kappa_{\mathcal F}^{\mathrm{cross}}(x,\xi)
    \mathcal F(\xi)
    \mathbf{W}_{\mathrm{value}}^{PaLA}\mathbf W_{\mathrm{out}}
    \,d\mu(\xi),
    \\
    \kappa_{\mathcal F}^{\mathrm{cross}}(x,\xi)
    &=
    \frac{1}{HM(H-1)}
    \sum_{h=1}^{H}\sum_{k\ne h}\sum_{j,\ell=1}^{M}
    a_{\mathcal F}^h(x,j)
    (\mathbf S^{k\rightarrow h})_{j\ell}
    a_{\mathcal F}^k(\xi,\ell).
\end{aligned}
\label{eq:pala_cross_kernel_representation}
\end{equation}
This completes the proof.
\end{proof}

In the full PaLA interaction core (Equation~(\ref{eq:pala_cross_kernel_representation})), the cross-space update is added to the original latent representation through a residual connection. Therefore, after CAP decoding and averaging across latent spaces, the output contains both the contribution preserved within each latent space and the contribution transferred across latent spaces.
Following the kernel formulation in Theorem~\ref{thm:cap_additive_kernel}, the residual branch induced by CAP encoding and decoding within each latent space can be represented by an input-dependent kernel. Combining this residual contribution with the cross-space kernel in Equation~(\ref{eq:pala_cross_kernel_representation}), the full PaLA interaction can be written as their superposition:
\begin{equation}
\begin{aligned}
    \mathcal T_{\mathrm{PaLA}}(\mathcal F)(x)
    &=
    \int_{\mathcal D}
    \kappa_{\mathcal F}^{\mathrm{res}}(x,\xi)\mathcal F(\xi)\,d\mu(\xi)
    +
    \int_{\mathcal D}
    \kappa_{\mathcal F}^{\mathrm{cross}}(x,\xi)\mathcal F(\xi)
    \mathbf{W}_{\mathrm{value}}^{PaLA}\mathbf W_{\mathrm{out}}\,d\mu(\xi),
    \\
    \kappa_{\mathcal F}^{\mathrm{res}}(x,\xi)
    &=
    \frac{1}{HM}\sum_{h=1}^{H}\sum_{j=1}^{M}
    a_{\mathcal F}^{h}(x,j)a_{\mathcal F}^{h}(\xi,j).
\end{aligned}
\label{eq:pala_residual_kernel}
\end{equation}
$\kappa_{\mathcal F}^{\mathrm{res}}$ characterizes the same-space information preserved by the residual connection, while $\kappa_{\mathcal F}^{\mathrm{cross}}$ characterizes the cross-space information exchange introduced by PaLA.

%% file: sec/properties_theorem2.tex
We analyze the discrete structure of PaLA and the information transfer across latent spaces.

\subsection{Discrete Kernel Representation of Parallel Latent Spaces with PaLA}
Corresponding to Equation~(\ref{eq:pala_residual_kernel}), the CAP-encoded sequence is $\mathbf Z_{\mathrm{encode}}^h=(A^h)^\top F$, and the PaLA update with the residual connection is:
\begin{equation}
    \mathbf Z_{\mathrm{new}}^h
    =
    \mathbf Z_{\mathrm{encode}}^h
    +
    \frac{1}{H-1}\sum_{k\ne h}
    \mathbf S^{k\rightarrow h}
    \mathbf Z_{\mathrm{encode}}^k
    \mathbf{W}_{\mathrm{value}}^{PaLA}\mathbf W_{\mathrm{out}}.
    \label{eq:pala_discrete_residual_update}
\end{equation}
Applying CAP decoding to the updated sequences and averaging across spaces gives:
\begin{equation}
\begin{aligned}
    F_{\mathrm{PaLA}}
    &=
    \frac{1}{H}\sum_{h=1}^{H}
    \frac{N}{M}A^h\mathbf Z_{\mathrm{new}}^h
    \\
    &=
    \underbrace{
    \left[
    \frac{N}{HM}\sum_{h=1}^{H}A^h(A^h)^\top
    \right]
    }_{\displaystyle \mathbf K_F^{\mathrm{res}}}
    F
    +
    \underbrace{
    \left[
    \frac{N}{HM(H-1)}
    \sum_{h=1}^{H}\sum_{k\ne h}
    A^h\mathbf S^{k\rightarrow h}(A^k)^\top
    \right]
    }_{\displaystyle \mathbf K_F^{\mathrm{cross}}}
    F\mathbf{W}_{\mathrm{value}}^{PaLA}\mathbf W_{\mathrm{out}},
\end{aligned}
\label{eq:pala_discrete_expansion}
\end{equation}
where $\mathbf K_F^{\mathrm{res}},\mathbf K_F^{\mathrm{cross}}\in\mathbb R^{N\times N}$ denote the same-space residual kernel and the cross-space interaction kernel, respectively.

\subsection{Cross-Space Kernel Coupling}
\label{sec:pala_kernel_coupling}

With $A_{\oplus}=[A^1,\ldots,A^H]$, we construct a block matrix $\mathbf S_{\oplus}\in\mathbb R^{HM\times HM}$ whose $(h,k)$-th block is:
\begin{equation}
    (\mathbf S_{\oplus})_{hk}
    =
    \begin{cases}
        \mathbf S^{k\rightarrow h}, & h\ne k,\\
        \mathbf 0, & h=k.
    \end{cases}
    \label{eq:pala_block_attention}
\end{equation}
The two discrete kernels can therefore be written as:
\begin{equation}
    \mathbf K_F^{\mathrm{res}}
    =
    \frac{N}{HM}A_{\oplus}A_{\oplus}^{\top},
    \qquad
    \mathbf K_F^{\mathrm{cross}}
    =
    \frac{N}{HM(H-1)}
    A_{\oplus}\mathbf S_{\oplus}A_{\oplus}^{\top}.
    \label{eq:pala_block_kernel}
\end{equation}
Compared with the block-diagonal structure of independent within-space interactions, PaLA introduces off-diagonal blocks that directly connect encoding in source space $k$ with decoding in target space $h$.
Consequently, information extracted through one aggregation pattern can return to the observation space through another, enabling cross-space combinations of latent representations.

\subsection{Transfer of Information Omitted by Individual Encodings}
\label{sec:pala_information_transfer}

\begin{table}[t]
    \caption{
        Fractions of actual input features omitted by each latent space and retained by the other latent spaces of PANO-P on Airfoil.
    }
    \centering
    \small
    \setlength{\tabcolsep}{5pt}
    \renewcommand{\arraystretch}{1.05}
    \begin{tabular}{cc|cccc}
        \toprule
        Block
        & Target $h$
        & Space 1
        & Space 2
        & Space 3
        & Space 4 \\
        \midrule
        \multirow{4}{*}{Block 1}
        & 1 & --         & 2.7521\%  & 1.2822\%  & 10.8355\% \\
        & 2 & 0.0096\%  & --         & 0.0001\% & 0.0055\% \\
        & 3 & 0.1095\%   & 0.0443\% & --         & 0.1870\%  \\
        & 4 & 12.7288\%  & 3.9956\%  & 1.7093\%  & --         \\
        \midrule
        \multirow{4}{*}{Block 2}
        & 1 & --         & 0.0012\% & 0.0013\% & 0.0002\% \\
        & 2 & 0.0232\%  & --         & 0.0106\% & 0.0046\% \\
        & 3 & 0.0025\%  & 0.0010\% & --         & 0.0021\% \\
        & 4 & 0.0331\%  & 0.1125\%  & 0.1051\%  & --         \\
        \bottomrule
    \end{tabular}
    \label{tab:actual_cross_space_retention}
\end{table}

To analyze cross-space information transfer, we first examine the proportion of the input component $F_{\mathrm{ignore}}^h$ omitted by the $h$-th latent space that is retained by the other latent spaces.
Specifically, we evaluate the proportion of $F_{\mathrm{ignore}}^h$ in the $k$-th latent space with $\left\|(A^k)^\top F_{\mathrm{ignore}}^h\right\|_F^2/\left\|(A^k)^\top F\right\|_F^2$, where $k\ne h$.
As shown in Table~\ref{tab:actual_cross_space_retention}, when $F_{\mathrm{ignore}}^h$ constitutes a large omitted component in space $h$, it is effectively retained by the other latent spaces.

We further analyze whether $F_{\mathrm{ignore}}^h$ is not only retained by the other latent spaces but also transferred back into space $h$ through PaLA.
According to Equation~(\ref{eq:pala_discrete_residual_update}), the representation change introduced by $F_{\mathrm{ignore}}^h$ after PaLA is:
\begin{equation}
\begin{aligned}
    \mathbf Z_{\mathrm{new}}^h\left(F_{\mathrm{ignore}}^h\right)
    &=
    (A^h)^\top F_{\mathrm{ignore}}^h
    +
    \frac{1}{H-1}\sum_{k\ne h}
    \mathbf S^{k\rightarrow h}
    (A^k)^\top F_{\mathrm{ignore}}^h
    \mathbf W_{\mathrm{value}}^{PaLA}
    \mathbf W_{\mathrm{out}}
    \\
    &=
    \frac{1}{H-1}\sum_{k\ne h}
    \mathbf S^{k\rightarrow h}
    (A^k)^\top F_{\mathrm{ignore}}^h
    \mathbf W_{\mathrm{value}}^{PaLA}
    \mathbf W_{\mathrm{out}}.
\end{aligned}
\label{eq:actual_pala_ignored_transfer}
\end{equation}

\begin{table}[t]
    \caption{
        Fractions of cross-space information transferred by PaLA relative to the updated latent representations of PANO-P on Airfoil.
    }
    \centering
    \small
    \setlength{\tabcolsep}{5pt}
    \renewcommand{\arraystretch}{1.05}
    \begin{tabular}{lcccc}
        \toprule
        Block
        & Space 1
        & Space 2
        & Space 3
        & Space 4 \\
        \midrule
        Block 1
        & 6.1682\%
        & 3.6895\%
        & 2.9733\%
        & 5.6461\% \\
        Block 2
        & 0.9488\%
        & 0.8416\%
        & 1.0126\%
        & 1.0638\% \\
        \bottomrule
    \end{tabular}
    \label{tab:actual_pala_incoming}
\end{table}

When the total incoming cross-space contribution is nonzero, $\mathbf Z_{\mathrm{new}}^h\left(F_{\mathrm{ignore}}^h\right)\ne0$, allowing the $h$-th latent space to distinguish input changes that its own encoding did not retain.
As shown in Table~\ref{tab:actual_pala_incoming}, PaLA produces nonzero cross-space information transfer in every latent space.
This property shows that PaLA can transfer information retained by other aggregation patterns into the current latent space, enabling updates beyond the information obtained through its own encoding.

%% file: sec/detailed_results.tex
Across the benchmarks, we compare PANO with 21 baseline methods, including MLP, PointNet~\citep{pointnet}, Graph U-Net~\citep{graphunet}, MeshGraphNet~\citep{meshgraphnet}, GNO~\citep{gno}, FNO~\citep{fno}, GEO-FNO~\citep{geo-fno}, F-FNO~\citep{f-fno}, Galerkin~\citep{galerkin}, OFormer~\citep{oformer}, GNOT~\citep{gnot}, FactFormer~\citep{factformer}, ONO~\citep{ono}, LSM~\citep{lsm}, LNO~\citep{lno}, GINO~\citep{gino}, 3D-GeoCA~\citep{3d-geoca}, Transolver~\citep{transolver}, Transolver++~\citep{transolverpp}, LinearNO~\citep{linearno}, and Unisoma~\citep{Unisoma}.
The following tables provide the complete results for the methods evaluated on benchmarks.

\noindent \textbf{Standard Benchmarks.}
Table~\ref{tab:detailed_standard} reports the comparison on Airfoil, Plasticity, NS2D, and Elasticity in relative L2 error.

\begin{table}[t]
  \caption{Comparison on standard benchmarks in relative L2 error ($\downarrow$). "-" indicates an unavailable result. The best and second-best results are \textbf{bold} and \underline{underlined}, respectively.}
  \centering
  \small
  \setlength{\tabcolsep}{6pt}
  \renewcommand{\arraystretch}{1.05}
  \begin{tabular}{lcccc}
    \toprule
    Method & Airfoil & Plasticity & NS2D & Elasticity \\
    \midrule
    FNO~\citep{fno} & - & - & 0.1556 & - \\
    GEO-FNO~\citep{geo-fno} & 0.0138 & 0.0074 & 0.1556 & 0.0229 \\
    F-FNO~\citep{f-fno} & 0.0078 & 0.0047 & 0.2322 & 0.0263 \\
    Galerkin~\citep{galerkin} & 0.0118 & 0.0120 & 0.1401 & 0.0240 \\
    OFormer~\citep{oformer} & 0.0183 & 0.0017 & 0.1705 & 0.0183 \\
    GNOT~\citep{gnot} & 0.0076 & 0.0336 & 0.1380 & 0.0086 \\
    FactFormer~\citep{factformer} & 0.0071 & 0.0312 & 0.1214 & - \\
    ONO~\citep{ono} & 0.0061 & 0.0048 & 0.1195 & 0.0118 \\
    LSM~\citep{lsm} & 0.0059 & 0.0025 & 0.1535 & 0.0218 \\
    LNO~\citep{lno} & 0.0053 & 0.0028 & 0.0830 & 0.0066 \\
    Transolver~\citep{transolver} & 0.0053 & 0.0013 & 0.0882 & 0.0065 \\
    Transolver++~\citep{transolverpp} & 0.0051 & 0.0014 & 0.1010 & 0.0064 \\
    LinearNO~\citep{linearno} & \underline{0.0049} & 0.0011 & \underline{0.0699} & 0.0050 \\
    \midrule
    \textbf{PANO-P (ours)} & \textbf{0.0044} & \underline{0.0006} & 0.0720 & \underline{0.0040} \\
    \textbf{PANO-D (ours)} & \textbf{0.0044} & \textbf{0.0004} & \textbf{0.0395} & \textbf{0.0029} \\
    \bottomrule
  \end{tabular}
  \label{tab:detailed_standard}
\end{table}

\noindent \textbf{AirfRANS.}
Table~\ref{tab:detailed_airfrans_standard} reports the complete physical-field and lift-coefficient prediction results under the standard AirfRANS setting.
Table~\ref{tab:detailed_airfrans_ood} further reports all metrics under Reynolds-number and angle-of-attack extrapolation to evaluate out-of-distribution generalization.

\begin{table}[t]
  \caption{Comparison on AirfRANS under the standard setting. Vol. and Surf. report prediction errors for surrounding and surface physical fields. $C_L$ reports the lift-coefficient prediction error, and $\rho_L$ is the Spearman rank correlation between predicted and ground-truth lift coefficients. The best and second-best results are \textbf{bold} and \underline{underlined}, respectively.}
  \centering
  \small
  \setlength{\tabcolsep}{6pt}
  \renewcommand{\arraystretch}{1.05}
  \begin{tabular}{lcccc}
    \toprule
    Method & Vol.$\downarrow$ & Surf.$\downarrow$ & $C_L\downarrow$ & $\rho_L\uparrow$ \\
    \midrule
    MLP & 0.0081 & 0.0200 & 0.2108 & 0.9932 \\
    PointNet~\citep{pointnet} & 0.0253 & 0.0996 & 0.1973 & 0.9919 \\
    Graph U-Net~\citep{graphunet} & 0.0076 & 0.0144 & 0.1677 & 0.9949 \\
    MeshGraphNet~\citep{meshgraphnet} & 0.0214 & 0.0387 & 0.2252 & 0.9945 \\
    GNO~\citep{gno} & 0.0269 & 0.0405 & 0.2016 & 0.9938 \\
    GEO-FNO~\citep{geo-fno} & 0.0361 & 0.0301 & 0.6161 & 0.9257 \\
    Galerkin~\citep{galerkin} & 0.0074 & 0.0159 & 0.2336 & 0.9951 \\
    GNOT~\citep{gnot} & 0.0049 & 0.0152 & 0.1992 & 0.9942 \\
    GINO~\citep{gino} & 0.0297 & 0.0482 & 0.1821 & 0.9958 \\
    LNO~\citep{lno} & 0.0214 & 0.0268 & 0.1480 & 0.9744 \\
    Transolver~\citep{transolver} & 0.0023 & 0.0085 & 0.1230 & 0.9978 \\
    Transolver++~\citep{transolverpp} & 0.0068 & 0.0159 & 0.1880 & 0.9910 \\
    LinearNO~\citep{linearno} & 0.0011 & 0.0077 & 0.0491 & \underline{0.9992} \\
    \midrule
    \textbf{PANO-P (ours)} & \textbf{0.0005} & \textbf{0.0010} & \underline{0.0236} & \textbf{0.9997} \\
    \textbf{PANO-D (ours)} & \underline{0.0009} & \underline{0.0031} & \textbf{0.0225} & \textbf{0.9997} \\
    \bottomrule
  \end{tabular}
  \label{tab:detailed_airfrans_standard}
\end{table}

\begin{table}[t]
  \caption{Complete AirfRANS results for Reynolds-number and angle-of-attack extrapolation. Vol. and Surf. report prediction errors for surrounding and surface physical fields. $C_L$ reports the lift-coefficient prediction error, and $\rho_L$ is the Spearman rank correlation between predicted and ground-truth lift coefficients. The best and second-best results are \textbf{bold} and \underline{underlined}, respectively.}
  \centering
  \small
  \setlength{\tabcolsep}{4pt}
  \renewcommand{\arraystretch}{1.05}
  \begin{tabular}{lcccccccc}
    \toprule
    \multirow{2}{*}{Method} & \multicolumn{4}{c}{OOD Reynolds} & \multicolumn{4}{c}{OOD Angles} \\
    \cmidrule(lr){2-5}\cmidrule(lr){6-9}
    & Vol.$\downarrow$ & Surf.$\downarrow$ & $C_L\downarrow$ & $\rho_L\uparrow$ & Vol.$\downarrow$ & Surf.$\downarrow$ & $C_L\downarrow$ & $\rho_L\uparrow$ \\
    \midrule
    MLP & 0.0669 & 0.1153 & 0.6205 & 0.9578 & 0.1309 & 0.3311 & 0.4128 & 0.9572 \\
    PointNet~\citep{pointnet} & 0.0838 & 0.1403 & 0.3836 & 0.9806 & 0.2021 & 0.4649 & 0.4425 & 0.9784 \\
    Graph U-Net~\citep{graphunet} & 0.0538 & 0.1168 & 0.4664 & 0.9645 & 0.0979 & 0.2391 & 0.3756 & 0.9816 \\
    MeshGraphNet~\citep{meshgraphnet} & 0.2789 & 0.2382 & 1.7718 & 0.7631 & 0.4902 & 1.1071 & 0.6525 & 0.8927 \\
    GNO~\citep{gno} & 0.0833 & 0.1562 & 0.4408 & 0.9878 & 0.1626 & 0.2359 & 0.3038 & 0.9884 \\
    Galerkin~\citep{galerkin} & 0.0330 & 0.0972 & 0.4615 & 0.9826 & 0.0577 & 0.2773 & 0.3814 & 0.9821 \\
    GNOT~\citep{gnot} & 0.0305 & 0.0959 & 0.3268 & 0.9865 & 0.0471 & 0.3466 & 0.3497 & 0.9868 \\
    GINO~\citep{gino} & 0.0839 & 0.1825 & 0.4180 & 0.9645 & 0.1589 & 0.2469 & 0.2583 & 0.9923 \\
    LNO~\citep{lno} & 0.0825 & 0.1762 & 0.7440 & 0.9399 & \underline{0.0346} & \textbf{0.0790} & 0.3732 & 0.9905 \\
    Transolver~\citep{transolver} & 0.0122 & 0.0550 & 0.1622 & 0.9904 & 0.0480 & 0.2335 & 0.2438 & 0.9948 \\
    LinearNO~\citep{linearno} & \underline{0.0112} & \underline{0.0372} & 0.2400 & 0.9951 & 0.0464 & 0.2500 & 0.0987 & 0.9963 \\
    \midrule
    \textbf{PANO-P (ours)} & 0.0113 & \textbf{0.0296} & \underline{0.1314} & \underline{0.9970} & \textbf{0.0261} & \underline{0.1709} & \underline{0.0671} & \underline{0.9977} \\
    \textbf{PANO-D (ours)} & \textbf{0.0106} & 0.0462 & \textbf{0.0834} & \textbf{0.9985} & 0.0362 & 0.1823 & \textbf{0.0658} & \textbf{0.9980} \\
    \bottomrule
  \end{tabular}
  \label{tab:detailed_airfrans_ood}
\end{table}

\noindent \textbf{Unilateral Stamping.}
Table~\ref{tab:detailed_unilateral_stamping} reports predictions of geometry, stress, and equivalent plastic strain for metal and rubber in the multi-solid stamping system.

\begin{table}[t]
  \caption{Comparison on Unilateral Stamping in relative L2 error ($\downarrow$). Geometry, Stress, and PEEQ evaluate the final deformed geometry, stress, and equivalent plastic strain of metal and rubber, respectively. The best and second-best results are \textbf{bold} and \underline{underlined}, respectively.}
  \centering
  \small
  \setlength{\tabcolsep}{6pt}
  \renewcommand{\arraystretch}{1.05}
  \begin{tabular}{lcccccc}
    \toprule
    \multirow{2}{*}{Method} & \multicolumn{2}{c}{Geometry} & \multicolumn{2}{c}{Stress} & \multicolumn{2}{c}{PEEQ} \\
    \cmidrule(lr){2-3}\cmidrule(lr){4-5}\cmidrule(lr){6-7}
    & Metal & Rubber & Metal & Rubber & Metal & Rubber \\
    \midrule
    GEO-FNO~\citep{geo-fno} & 0.0109 & 0.0154 & 0.1113 & 0.1124 & 0.3771 & 0.3574 \\
    GINO~\citep{gino} & 0.0188 & 0.0219 & 0.1262 & 0.1438 & 0.4651 & 0.4885 \\
    GNO~\citep{gno} & 0.0426 & 0.0424 & 0.1351 & 0.1814 & 0.6413 & 0.6859 \\
    LSM~\citep{lsm} & 0.0367 & 0.0357 & 0.1327 & 0.1335 & 0.5889 & 0.6026 \\
    Galerkin~\citep{galerkin} & 0.1007 & 0.1010 & 0.1725 & 0.2794 & 0.7986 & 0.8555 \\
    FactFormer~\citep{factformer} & 0.0178 & 0.0166 & 0.1095 & 0.1035 & 0.3886 & 0.3022 \\
    OFormer~\citep{oformer} & 0.0119 & 0.0098 & 0.1103 & 0.0981 & 0.3269 & 0.2675 \\
    ONO~\citep{ono} & 0.0433 & 0.0437 & 0.1377 & 0.1311 & 0.5140 & 0.4748 \\
    MeshGraphNet~\citep{meshgraphnet} & 0.0391 & 0.0383 & 0.1252 & 0.1417 & 0.5805 & 0.5694 \\
    Transolver~\citep{transolver} & 0.0079 & 0.0075 & 0.0839 & 0.0862 & 0.2938 & 0.2693 \\
    Unisoma~\citep{Unisoma} & 0.0071 & 0.0063 & 0.0807 & 0.0773 & 0.2591 & 0.2114 \\
    \midrule
    \textbf{PANO-P (ours)} & \textbf{0.0052} & \textbf{0.0044} & \textbf{0.0743} & \textbf{0.0667} & \textbf{0.1827} & \textbf{0.1515} \\
    \textbf{PANO-D (ours)} & \underline{0.0067} & \underline{0.0058} & \underline{0.0756} & \underline{0.0710} & \underline{0.2248} & \underline{0.1822} \\
    \bottomrule
  \end{tabular}
  \label{tab:detailed_unilateral_stamping}
\end{table}

\noindent \textbf{ShapeNetCar.}
Table~\ref{tab:detailed_shapenetcar} reports the complete comparison of surrounding velocity, surface pressure, and drag-coefficient predictions on ShapeNetCar.

\begin{table}[t]
  \caption{Comparison on ShapeNetCar. Vol. and Surf. report prediction errors for the surrounding velocity field and surface pressure field. $C_D$ reports the drag-coefficient prediction error, and $\rho_D$ is the Spearman rank correlation between predicted and ground-truth drag coefficients. The best and second-best results are \textbf{bold} and \underline{underlined}, respectively.}
  \centering
  \small
  \setlength{\tabcolsep}{6pt}
  \renewcommand{\arraystretch}{1.05}
  \begin{tabular}{lcccc}
    \toprule
    Method & Vol.$\downarrow$ & Surf.$\downarrow$ & $C_D\downarrow$ & $\rho_D\uparrow$ \\
    \midrule
    MLP & 0.0512 & 0.1304 & 0.0307 & 0.9496 \\
    PointNet~\citep{pointnet} & 0.0494 & 0.1104 & 0.0298 & 0.9583 \\
    Graph U-Net~\citep{graphunet} & 0.0471 & 0.1102 & 0.0226 & 0.9725 \\
    GNO~\citep{gno} & 0.0383 & 0.0815 & 0.0172 & 0.9834 \\
    GEO-FNO~\citep{geo-fno} & 0.1670 & 0.2378 & 0.0664 & 0.8280 \\
    Galerkin~\citep{galerkin} & 0.0339 & 0.0878 & 0.0179 & 0.9764 \\
    GNOT~\citep{gnot} & 0.0329 & 0.0798 & 0.0178 & 0.9833 \\
    GINO~\citep{gino} & 0.0386 & 0.0810 & 0.0184 & 0.9826 \\
    LNO~\citep{lno} & 0.0269 & 0.0870 & 0.0174 & 0.9781 \\
    Transolver~\citep{transolver} & 0.0221 & 0.0785 & \underline{0.0117} & 0.9914 \\
    Transolver++~\citep{transolverpp} & 0.0226 & 0.0800 & 0.0132 & 0.9914 \\
    LinearNO~\citep{linearno} & \underline{0.0194} & \textbf{0.0754} & \textbf{0.0106} & \textbf{0.9925} \\
    \midrule
    \textbf{PANO-P (ours)} & 0.0214 & 0.0817 & 0.0126 & 0.9874 \\
    \textbf{PANO-D (ours)} & \textbf{0.0187} & \underline{0.0784} & \underline{0.0117} & \underline{0.9917} \\
    \bottomrule
  \end{tabular}
  \label{tab:detailed_shapenetcar}
\end{table}

\noindent \textbf{AirCraft.}
Table~\ref{tab:detailed_aircraft} reports surface physical-field and lift-coefficient prediction results on the large-scale three-dimensional AirCraft benchmark.

\begin{table}[t]
  \caption{Comparison on AirCraft. Surf. reports the relative L2 error of surface physical fields, $C_L$ reports the lift-coefficient prediction error, and $R_L^2$ is the coefficient of determination for lift predictions. Parentheses retain the additional precision reported for PANO. The best and second-best results are \textbf{bold} and \underline{underlined}, respectively.}
  \centering
  \small
  \setlength{\tabcolsep}{6pt}
  \renewcommand{\arraystretch}{1.05}
  \begin{tabular}{lccc}
    \toprule
    Method & Surf.$\downarrow$ & $C_L\downarrow$ & $R_L^2\uparrow$ \\
    \midrule
    PointNet~\citep{pointnet} & 0.169 & 0.095 & 0.982 \\
    Graph U-Net~\citep{graphunet} & 0.161 & 0.063 & 0.953 \\
    MeshGraphNet~\citep{meshgraphnet} & 0.113 & 0.038 & 0.993 \\
    GNO~\citep{gno} & 0.129 & 0.031 & 0.991 \\
    Galerkin~\citep{galerkin} & 0.118 & 0.069 & 0.879 \\
    GEO-FNO~\citep{geo-fno} & 0.395 & 0.243 & 0.903 \\
    GINO~\citep{gino} & 0.133 & 0.047 & 0.983 \\
    GNOT~\citep{gnot} & 0.093 & 0.033 & 0.991 \\
    LNO~\citep{lno} & 0.137 & 0.091 & 0.992 \\
    3D-GeoCA~\citep{3d-geoca} & 0.097 & 0.022 & 0.993 \\
    Transolver~\citep{transolver} & 0.092 & 0.037 & 0.994 \\
    Transolver++~\citep{transolverpp} & 0.064 & \underline{0.014} & \underline{0.999} \\
    LinearNO~\citep{linearno} & 0.065 & 0.052 & 0.995 \\
    \midrule
    \textbf{PANO-P (ours)} & \underline{0.063} & 0.016 & \textbf{1.000 (0.9995)} \\
    \textbf{PANO-D (ours)} & \textbf{0.056} & \textbf{0.011} & \textbf{1.000 (0.9997)} \\
    \bottomrule
  \end{tabular}
  \label{tab:detailed_aircraft}
\end{table}

%% file: sec/dataset_intro.tex
\subsection{Standard Benchmarks}

\begin{table*}[!t]
    \caption{Data split of benchmarks.}
    \centering
    \small
    \renewcommand{\tabcolsep}{6pt}
    \renewcommand{\arraystretch}{1.1}
    \begin{tabular}{lcc}
        \toprule
        Benchmark & Training Set & Test Set \\
        \midrule
        Airfoil & 1000 & 200\\
        Plasticity & 900 & 80\\
        NS2D & 1000 & 200\\
        Elasticity & 1000 & 200\\
        Darcy & 1000 & 200\\
        AirfRANS (Standard Setting) & 720 & 200 \\
        Unilateral Stamping & 1000 & 100\\
        ShapeNetCar & 789 & 100\\
        Aircraft & 140 & 10\\
        \bottomrule
    \end{tabular}
    \label{tb:split}
\end{table*}

\begin{table*}[!t]
    \caption{Training and test ranges for the AirfRANS OOD Reynolds and OOD Angles evaluations.}
    \centering
    \small
    \renewcommand{\tabcolsep}{6pt}
    \renewcommand{\arraystretch}{1.1}
    \begin{tabular}{lcc}
        \toprule
        Split & OOD Reynolds: Reynolds Number Range & OOD Angles: AoA Range \\
        \midrule
        Training Set
        & $[3\times10^{6},\,5\times10^{6}]$
        & $[-2.5^{\circ},\,12.5^{\circ}]$ \\
        Test Set
        & $[2\times10^{6},\,3\times10^{6}]
          \cup[5\times10^{6},\,6\times10^{6}]$
        & $[-5^{\circ},\,-2.5^{\circ}]
          \cup[12.5^{\circ},\,15^{\circ}]$ \\
        \bottomrule
    \end{tabular}
    \label{tab:ood_generalization_configuration}
\end{table*}

\begin{table}[t]
    \caption{
        Training hyperparameters of PANO-P and PANO-D.
        Peak LR denotes the maximum learning rate of the OneCycle schedule.
        Paired values correspond to PANO-P / PANO-D.
    }
    \centering
    \small
    \setlength{\tabcolsep}{4pt}
    \renewcommand{\arraystretch}{1.1}
    \begin{tabular}{lccccc}
        \toprule
        Benchmark & Epochs & Peak LR & Batch size & Optimizer & Scheduler \\
        \midrule
        Airfoil
        & 500 & $8\times10^{-4}$ & 4 & AdamW & OneCycle \\
        Plasticity
        & 500 & $10^{-3}$ & 8 & AdamW & OneCycle \\
        NS2D
        & 500 & $10^{-3}$ & 2 & AdamW & OneCycle \\
        Elasticity
        & 500 & $10^{-3}$ & 4 & AdamW & OneCycle \\
        Darcy85
        & 500 & $10^{-3}$ & 4 & AdamW & OneCycle \\
        Darcy141
        & 500 & $10^{-3}$ & 4 & AdamW & OneCycle \\
        Darcy211
        & 500 & $10^{-3}$ & 4 / 2 & AdamW & OneCycle \\
        \midrule
        AirfRANS (Standard)
        & 1600 & $2\times10^{-4}$ & 4 & AdamW & OneCycle \\
        AirfRANS (OOD Reynolds)
        & 1600 & $2\times10^{-4}$ & 4 & AdamW & OneCycle \\
        AirfRANS (OOD Angles)
        & 1600 & $2\times10^{-4}$ & 4 & AdamW & OneCycle \\
        Unilateral Stamping
        & 100 & $10^{-3}$ & 1 & AdamW & OneCycle \\
        ShapeNetCar
        & 400 & $8\times10^{-4}$ & 2 & AdamW & OneCycle \\
        Aircraft
        & 200 & $10^{-3}$ & 1 & AdamW & OneCycle \\
        \bottomrule
    \end{tabular}
    \label{tab:pano_training_settings}
\end{table}

\noindent \textbf{Airfoil.}
Airfoil~\citep{geo-fno} considers Mach number prediction around varying airfoil geometries.
Each sample is represented on a $221\times51$ structured mesh, and the model predicts the Mach number field from the two-dimensional mesh coordinates.
The dataset contains 1,000 training samples and 200 testing samples.

\noindent \textbf{Plasticity.}
Plasticity~\citep{geo-fno} considers time-dependent deformation of plastic materials under different dies.
The physical domain is represented on a $101\times31$ structured mesh, and the model uses the input conditions and target time to predict deformation states independently at 20 time steps.

\noindent \textbf{NS2D.}
NS2D~\citep{fno} considers the temporal evolution of two-dimensional flow fields under different initial conditions.
On a $64\times64$ regular grid, the model takes observations from 10 preceding time steps and autoregressively predicts the next 10 steps.

\noindent \textbf{Elasticity.}
Elasticity~\citep{geo-fno} considers stress distributions in varying material structures.
Each sample contains 972 observation points, and the model maps their two-dimensional coordinates to the corresponding stress values.

\noindent \textbf{Darcy.}
Darcy~\citep{fno} considers steady flow through porous media with varying permeability distributions, requiring the model to predict the pressure field from spatial coordinates and permeability coefficients.
We subsample the original $421\times421$ grid to $85\times85$, $141\times141$, and $211\times211$, forming Darcy85, Darcy141, and Darcy211 to evaluate predictive performance at different observation densities.

\subsection{Advanced Benchmarks}

\noindent \textbf{AirfRANS.}
AirfRANS~\citep{airfrans} consists of two-dimensional incompressible steady flows around varying airfoil geometries at different Reynolds numbers and angles of attack, generated using the Reynolds-averaged Navier--Stokes equations.
Each case contains a variable number of observation points, with spatial coordinates, inlet velocity, distance to the airfoil, and surface normals as inputs, and velocity, pressure, and turbulent kinematic viscosity as prediction targets.
We consider three settings: Standard Setting, OOD Reynolds, and OOD Angles, evaluating standard predictive performance and generalization to Reynolds numbers and angles of attack outside the training ranges.
In addition to prediction errors in the surrounding flow field and on the airfoil surface, we assess aerodynamic prediction using lift-coefficient errors and Spearman rank correlations.

\noindent \textbf{Unilateral Stamping.}
Unilateral Stamping~\citep{Unisoma} models the deformation of a hollow metal workpiece and an internal rubber insert during multi-point unilateral stamping, supporting long-time state prediction in complex multi-body systems.
The system contains 16 moving rigid dies and one stationary rigid die that shape the metal workpiece, while the rubber insert helps preserve its cross-sectional integrity during deformation.
Geometric parameters are sampled uniformly within prescribed ranges to generate diverse stamping configurations.
Each sample provides the initial and final system states, with prediction targets comprising the final geometry, internal stress, and equivalent plastic strain (PEEQ) of both metal and rubber.

\noindent \textbf{ShapeNetCar.}
ShapeNetCar~\citep{shapenetcar} comprises aerodynamic simulations of 889 distinct car geometries traveling at 72 km/h, covering physical fields on the vehicle surface and in the surrounding domain.
Each case is represented by 32,186 unstructured mesh points, with three-dimensional coordinates, a signed distance function, and normal vectors as inputs, and three-dimensional velocity and pressure as prediction targets.
Following the data split used by Transolver~\citep{transolver}, we use 789 cases for training and 100 for testing.

\noindent \textbf{AirCraft.}
AirCraft~\citep{transolverpp} is a large-scale three-dimensional aerodynamic benchmark with approximately 330K surface observation points per sample, where the inputs consist of point coordinates, surface normals, and sample-level flow conditions including Mach number, angle of attack, and sideslip angle. The model predicts six pointwise physical quantities, including the pressure coefficient $C_p$, fluid density $\rho$, three velocity components $(u,v,w)$, and pressure $p$.

\subsection{Data Split}

We follow existing methods~\cite{transolver,linearno,Unisoma} for data splitting. In particular, the data split of each benchmark is as shown in Table~\ref{tb:split}.
In addition, the setting of OOD Reynolds and OOD Angles on AirfRANS is shown in Table~\ref{tab:ood_generalization_configuration}.

%% file: sec/implementation_details.tex
\noindent \textbf{Baselines.}
We primarily reuse baseline results reported in prior work.
Specifically, baseline results on the standard benchmarks, AirfRANS, and ShapeNetCar are taken from LinearNO~\citep{linearno}, while those on AirCraft and Unilateral Stamping are taken from Transolver++~\citep{transolverpp} and Unisoma~\citep{Unisoma}, respectively.

\noindent \textbf{Implementation of PaLA.}
In practice, we implement and adopt multi-head PaLA instead of the standard single-head PaLA to further enhance cross-space interaction.
Specifically, we split the matching and value representations along the feature dimension into multiple attention heads, each independently computing cross-space attention and aggregating information.
The results from all heads are concatenated and passed through the output projection, allowing PaLA to integrate cross-space associations from multiple feature subspaces.

\noindent \textbf{Training of PANO.}
All PANO experiments use a random seed of 0, the AdamW optimizer with a weight decay of $5\times10^{-5}$, and the OneCycle learning-rate scheduler.
Table~\ref{tab:pano_training_settings} summarizes the training configurations of both variants across all benchmarks.
Both variants use the same settings unless separate values are provided in the order PANO-P / PANO-D.

For all benchmarks except AirfRANS, we perform physical field prediction at all observation points of each sample. For AirfRANS, we follow previous works~\cite{transolver,linearno} by sampling 32,000 points from each AirfRANS case during training, and repeatedly performing sampling and prediction during inference until all points are covered.
We further follow the AirfRANS sampling strategy of MoNo~\cite{mono}, where each sampled case contains all surface points together with randomly sampled volume points, yielding 32,000 points in total.

\noindent \textbf{Hardware and Software Environment.}
Except for the experiments on AirCraft, all other experiments are conducted on a single NVIDIA GeForce RTX 4090 GPU.
The experiments on AirCraft are conducted on a single NVIDIA H20 GPU with mixed precision to accelerate training.
The core software environment consists of PyTorch 2.8.0 and CUDA 12.8.

\noindent \textbf{Evaluation.}
For all experiments, we evaluate performance using the model saved at the last epoch, without any model selection strategy.
We use the relative L2 error to evaluate prediction accuracy across all standard benchmarks.
Specifically, for each test sample, the relative L2 error is defined as the L2 error between the predicted and ground-truth physical fields normalized by the L2 norm of the ground truth.
The reported result is averaged over all testing samples.

For AirfRANS, ShapeNetCar, and AirCraft, in addition to the relative L2 error of the surrounding velocity field and surface pressure field, we further evaluate aerodynamic performance~\cite{transolver,linearno}.
In particular, we introduce the lift coefficient for evaluation on AirfRANS and AirCraft. We introduce the drag coefficient for evaluation on ShapeNetCar.
The detailed computation follows Transolver~\cite{transolver} and Transolver++~\cite{transolverpp}.
To evaluate the performance on Unilateral Stamping, we mainly follow UNISOMA~\cite{Unisoma} and compute the relative L2 error for the deformed geometry, stress, and equivalent plastic strain of the metal and rubber components.

%% file: sec/showcase.tex
\begin{figure}[!t]
    \centering
    \includegraphics[width=0.99\linewidth]{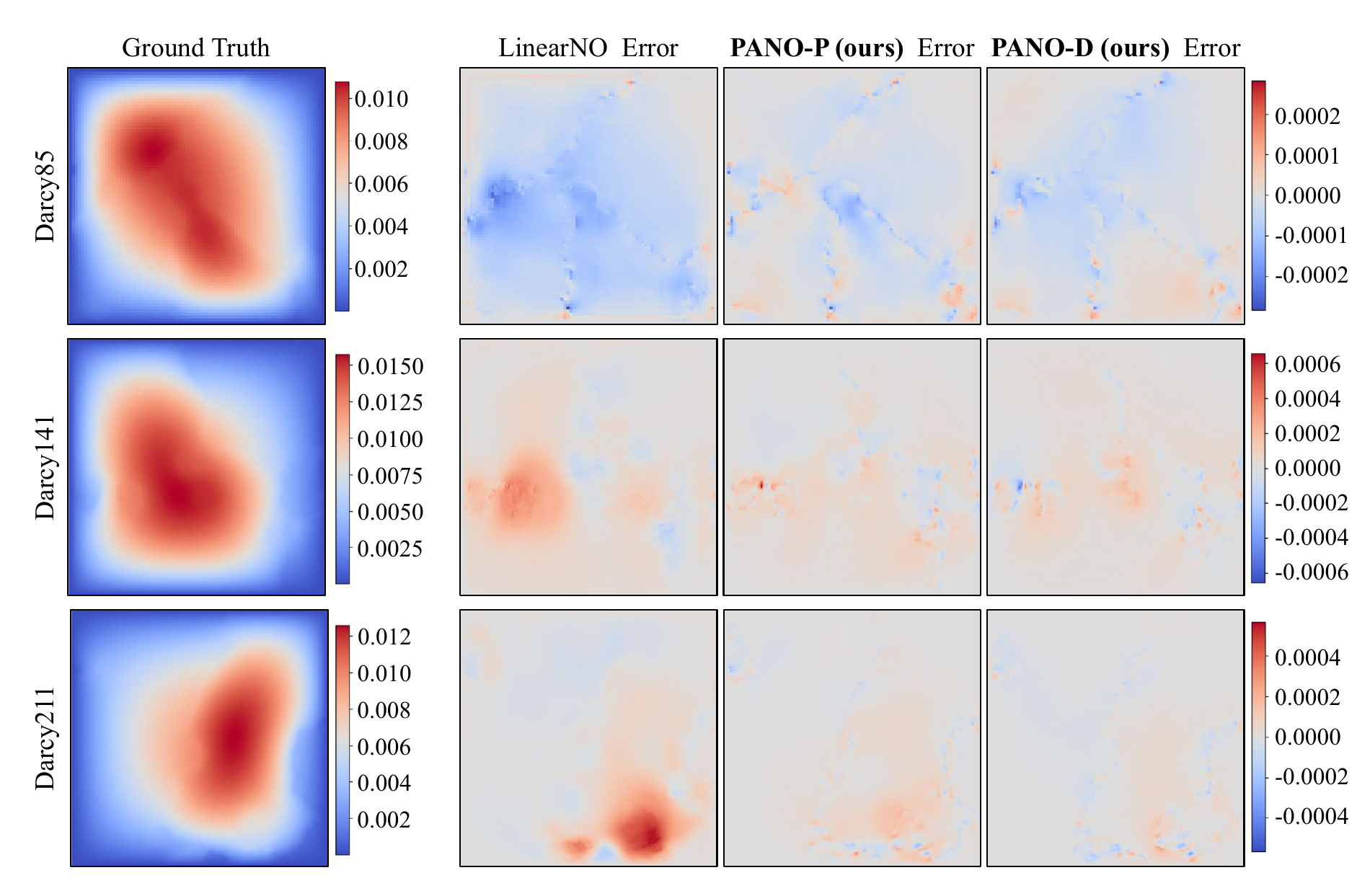}
    \caption{Visualization of prediction error on Darcy.}
    \label{fig:darcy_error}
\end{figure}
\begin{figure}[!t]
    \centering
    \includegraphics[width=0.99\linewidth]{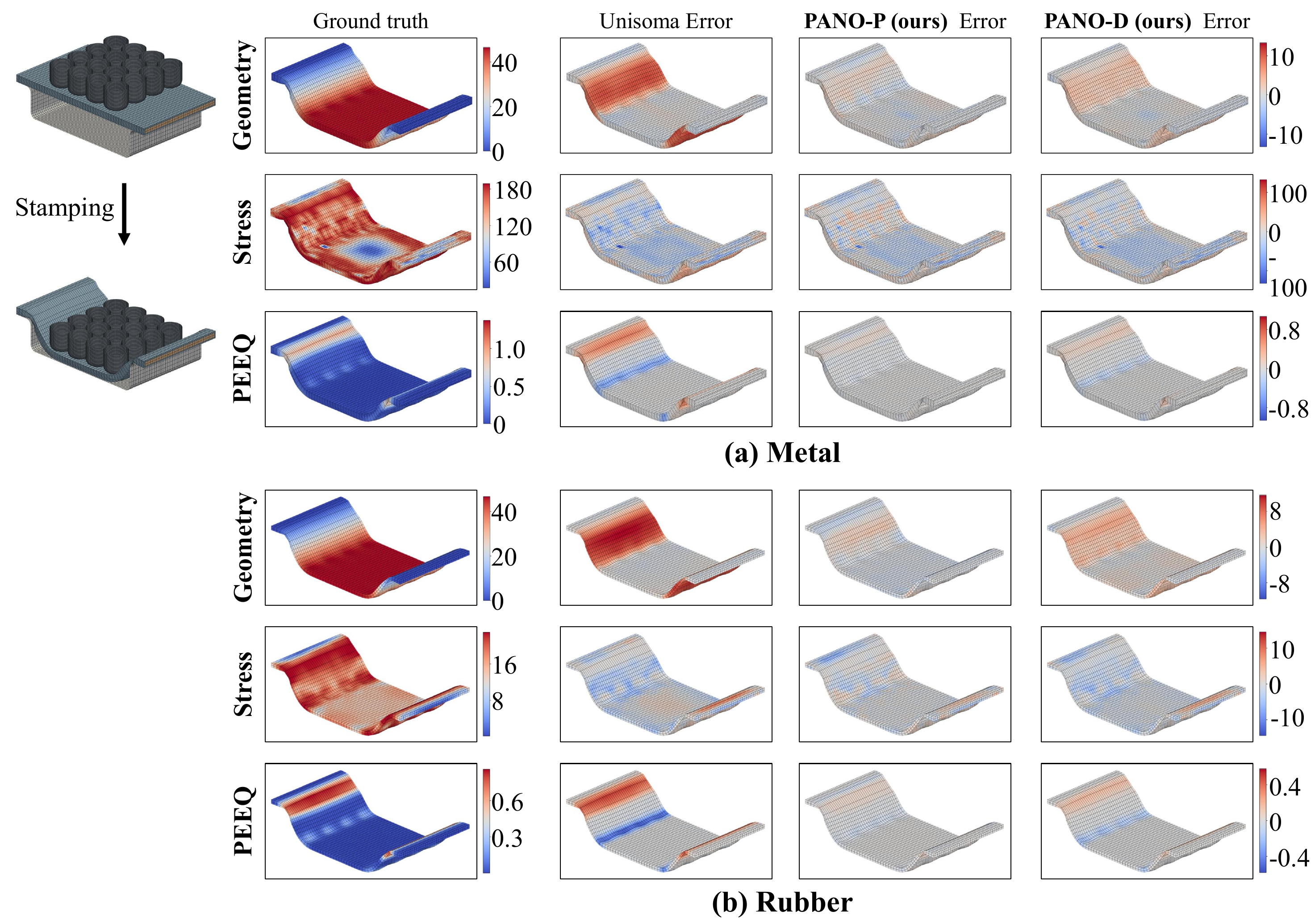}
    \caption{Visualization of prediction error on Unilateral Stamping.}
    \label{fig:stamping_error}
\end{figure}
\begin{figure}[!t]
    \centering
    \includegraphics[width=0.99\linewidth]{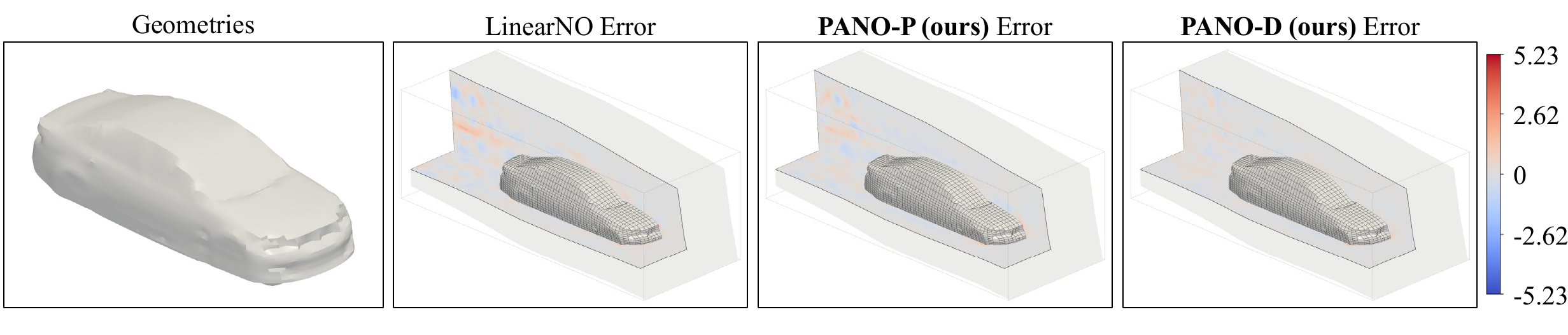}
    \caption{Visualization of prediction error on ShapeNetCar.}
    \label{fig:shapenetcar_error}
\end{figure}

\noindent \textbf{Prediction Errors.}
Figures~\ref{fig:darcy_error}, \ref{fig:stamping_error}, and \ref{fig:shapenetcar_error} visualize prediction errors on Darcy, Unilateral Stamping, and ShapeNetCar, respectively.

\begin{figure}[!t]
    \centering
    \includegraphics[width=0.85\linewidth]{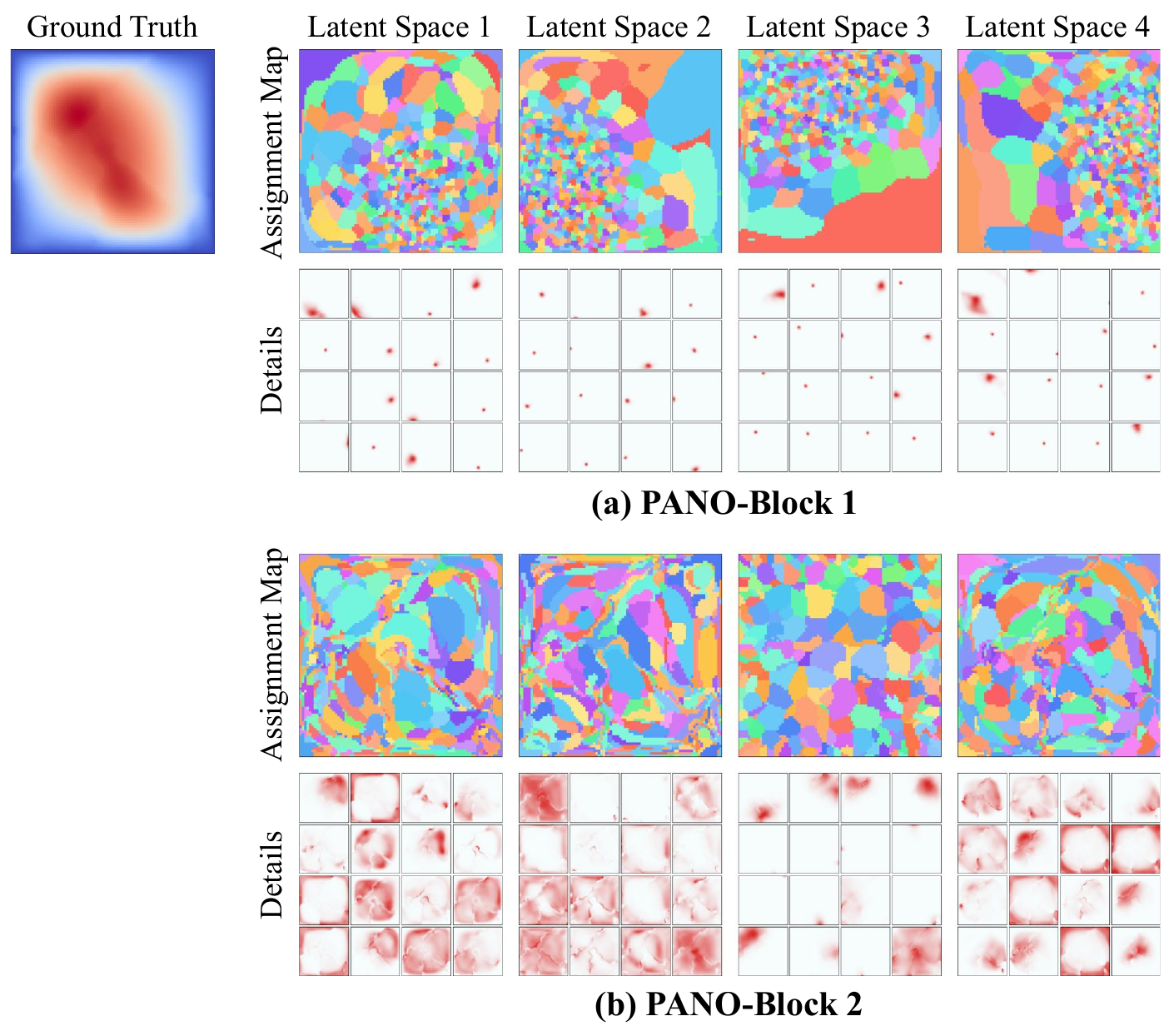}
    \caption{Visualization of PANO-P assignment maps on Darcy85.}
    \label{fig:darcy85_router_pano-p}
\end{figure}

\begin{figure}[!t]
    \centering
    \includegraphics[width=0.85\linewidth]{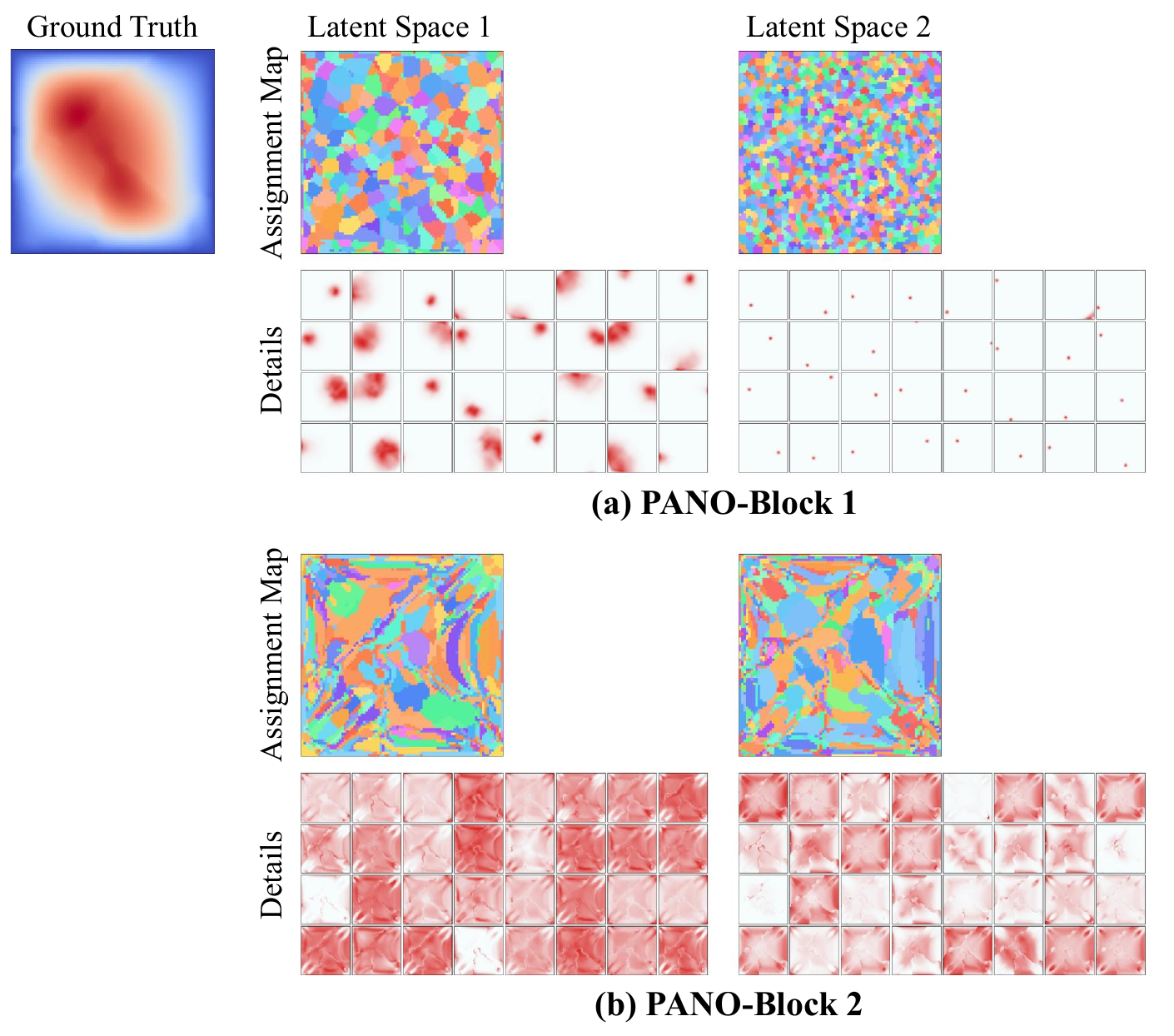}
    \caption{Visualization of PANO-D assignment maps on Darcy85.}
    \label{fig:darcy85_router_pano-d}
\end{figure}

\begin{figure}[!t]
    \centering
    \includegraphics[width=0.85\linewidth]{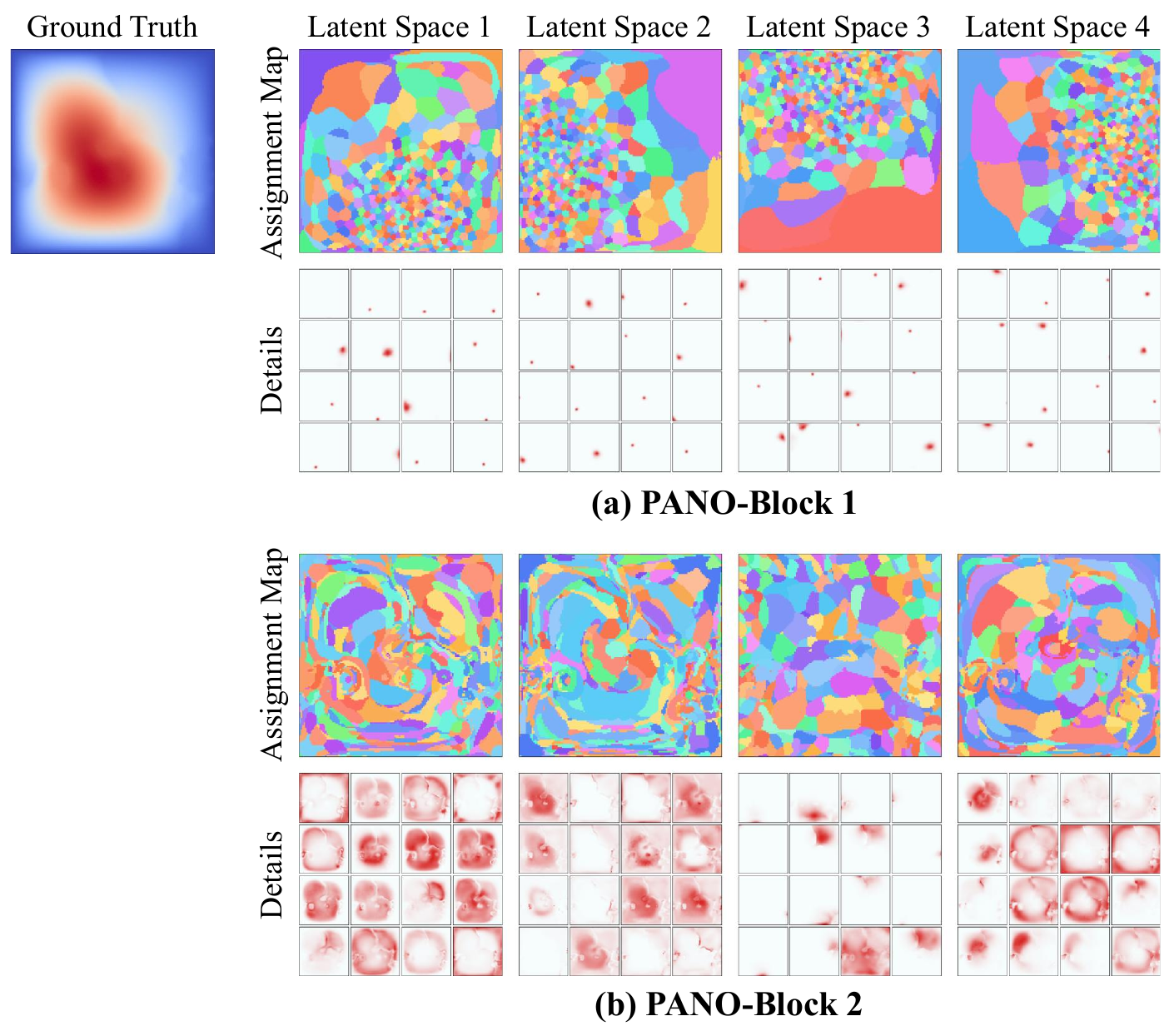}
    \caption{Visualization of PANO-P assignment maps on Darcy141.}
    \label{fig:darcy141_router_pano-p}
\end{figure}

\begin{figure}[!t]
    \centering
    \includegraphics[width=0.85\linewidth]{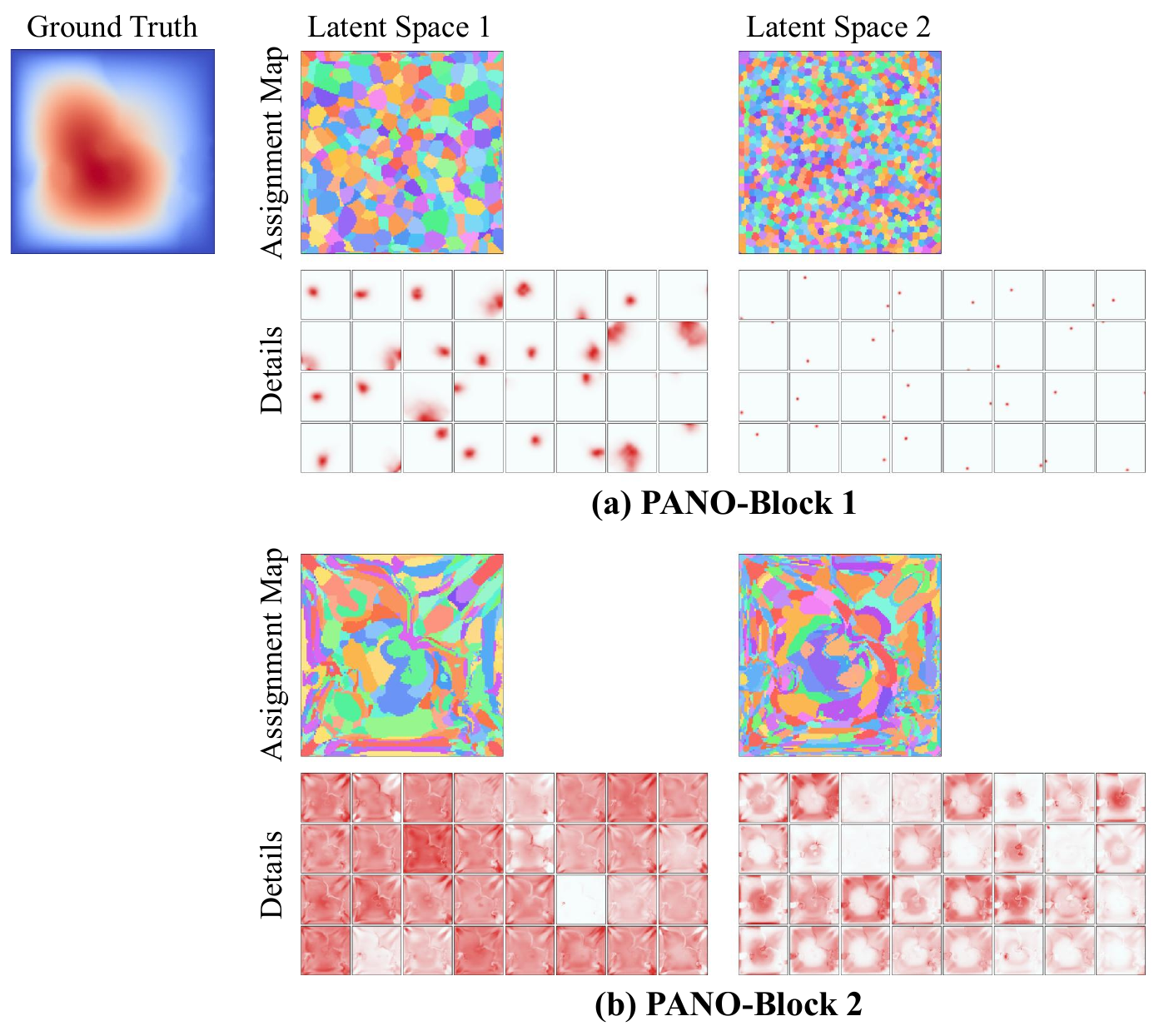}
    \caption{Visualization of PANO-D assignment maps on Darcy141.}
    \label{fig:darcy141_router_pano-d}
\end{figure}

\begin{figure}[!t]
    \centering
    \includegraphics[width=0.85\linewidth]{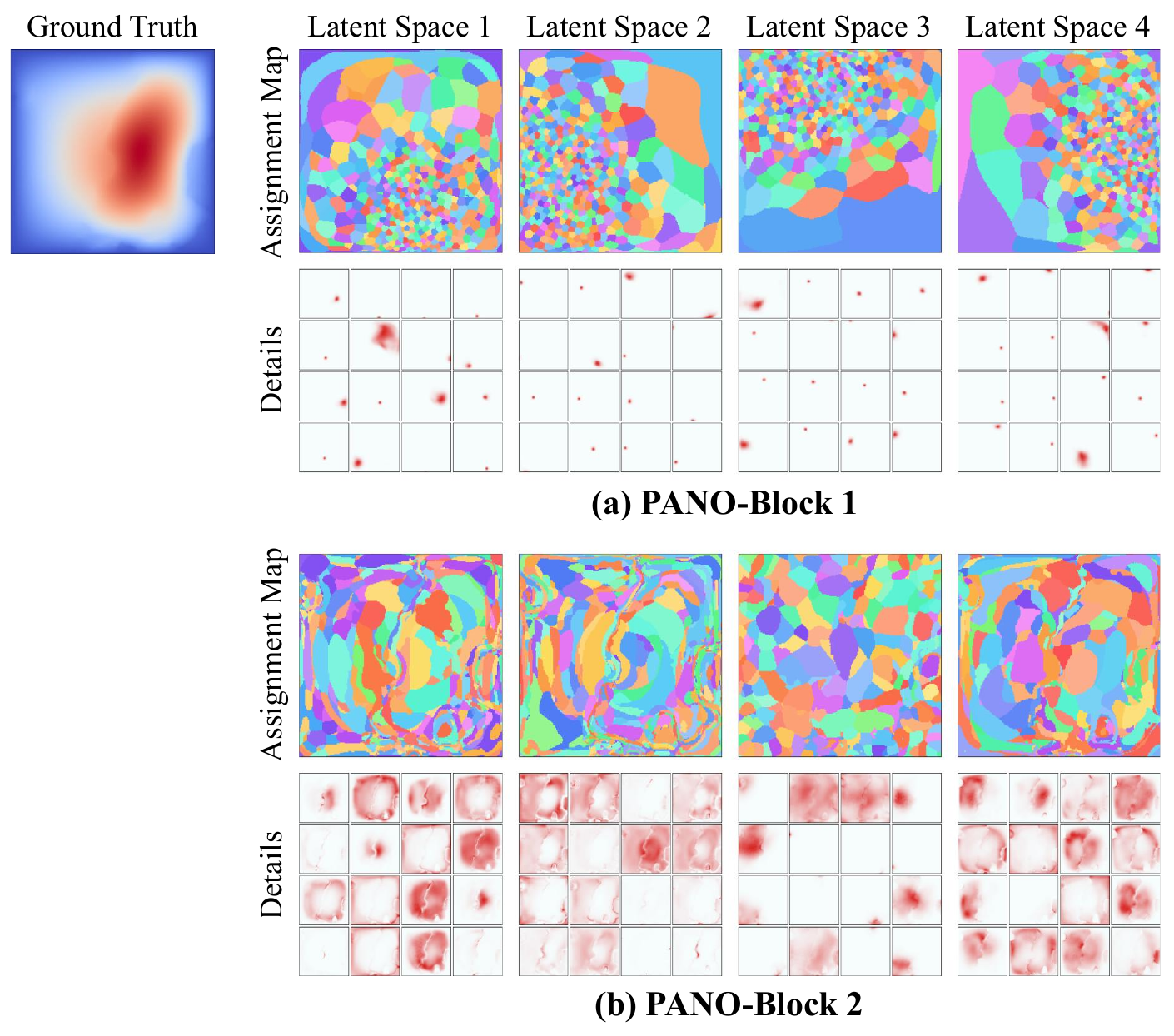}
    \caption{Visualization of PANO-P assignment maps on Darcy211.}
    \label{fig:darcy211_router_pano-p}
\end{figure}

\begin{figure}[!t]
    \centering
    \includegraphics[width=0.85\linewidth]{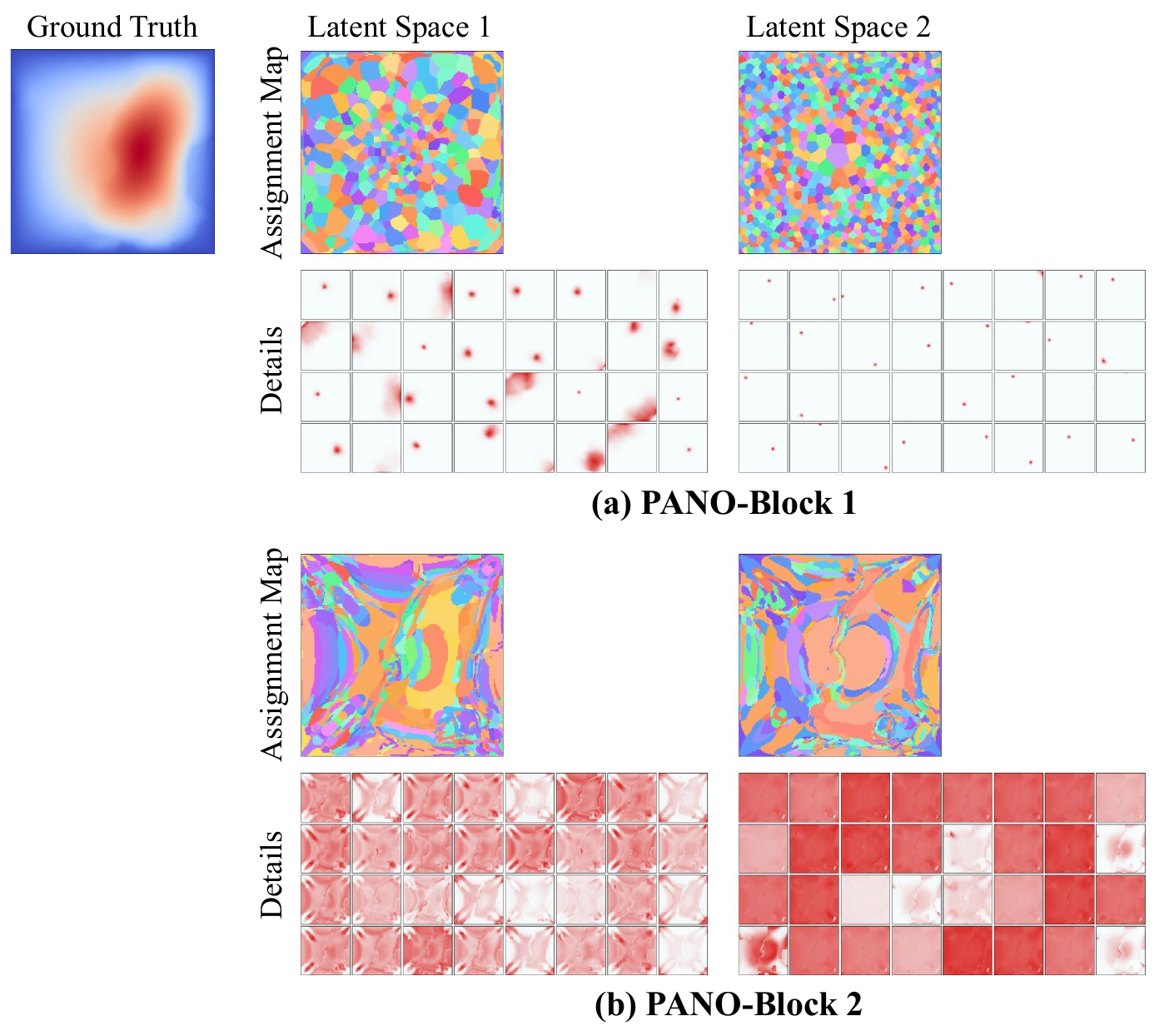}
    \caption{Visualization of PANO-D assignment maps on Darcy211.}
    \label{fig:darcy211_router_pano-d}
\end{figure}

\begin{figure}[!t]
    \centering
    \includegraphics[width=0.85\linewidth]{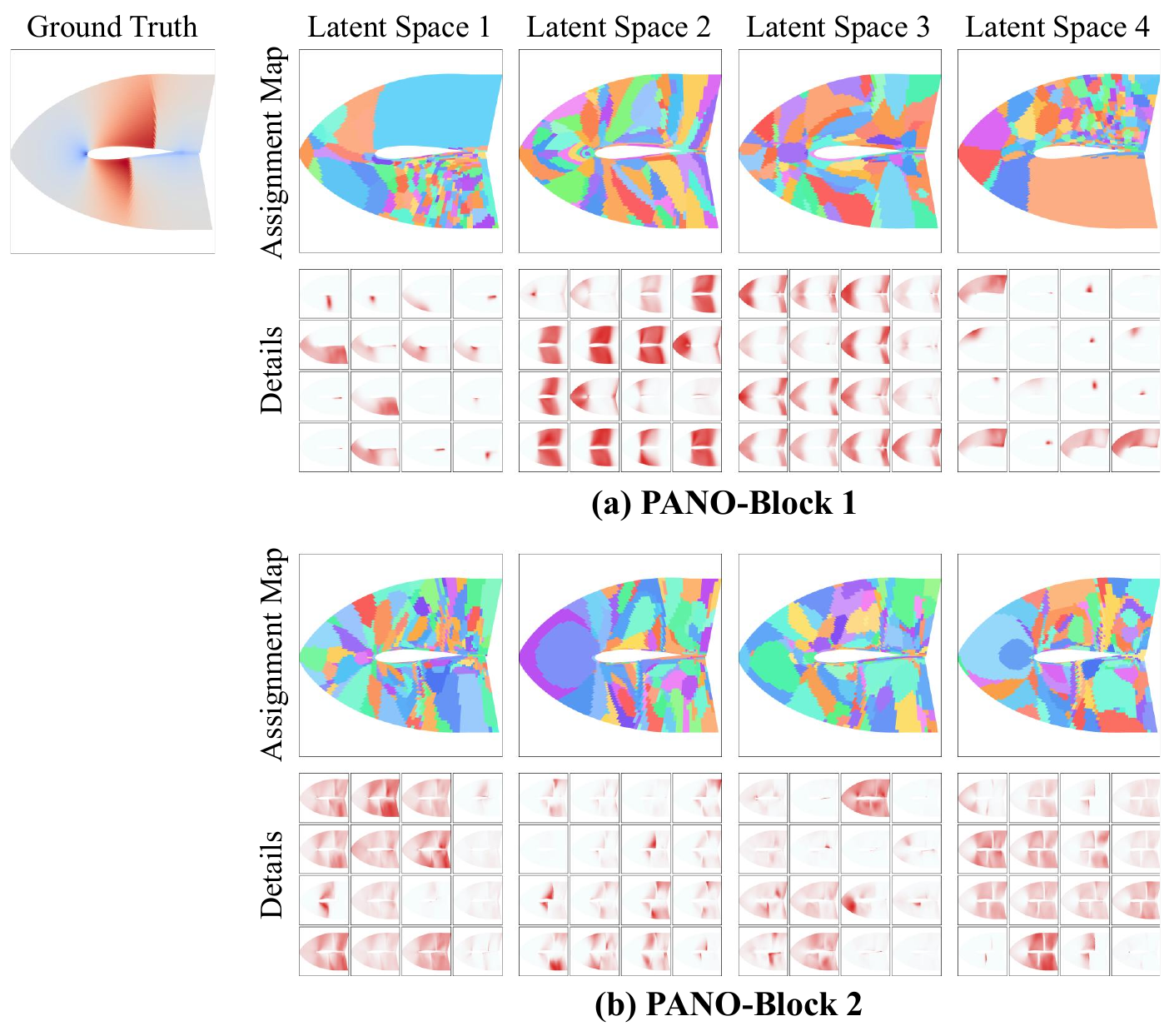}
    \caption{Visualization of PANO-P assignment maps on Airfoil.}
    \label{fig:airfoil_router_pano-p}
\end{figure}

\begin{figure}[!t]
    \centering
    \includegraphics[width=0.85\linewidth]{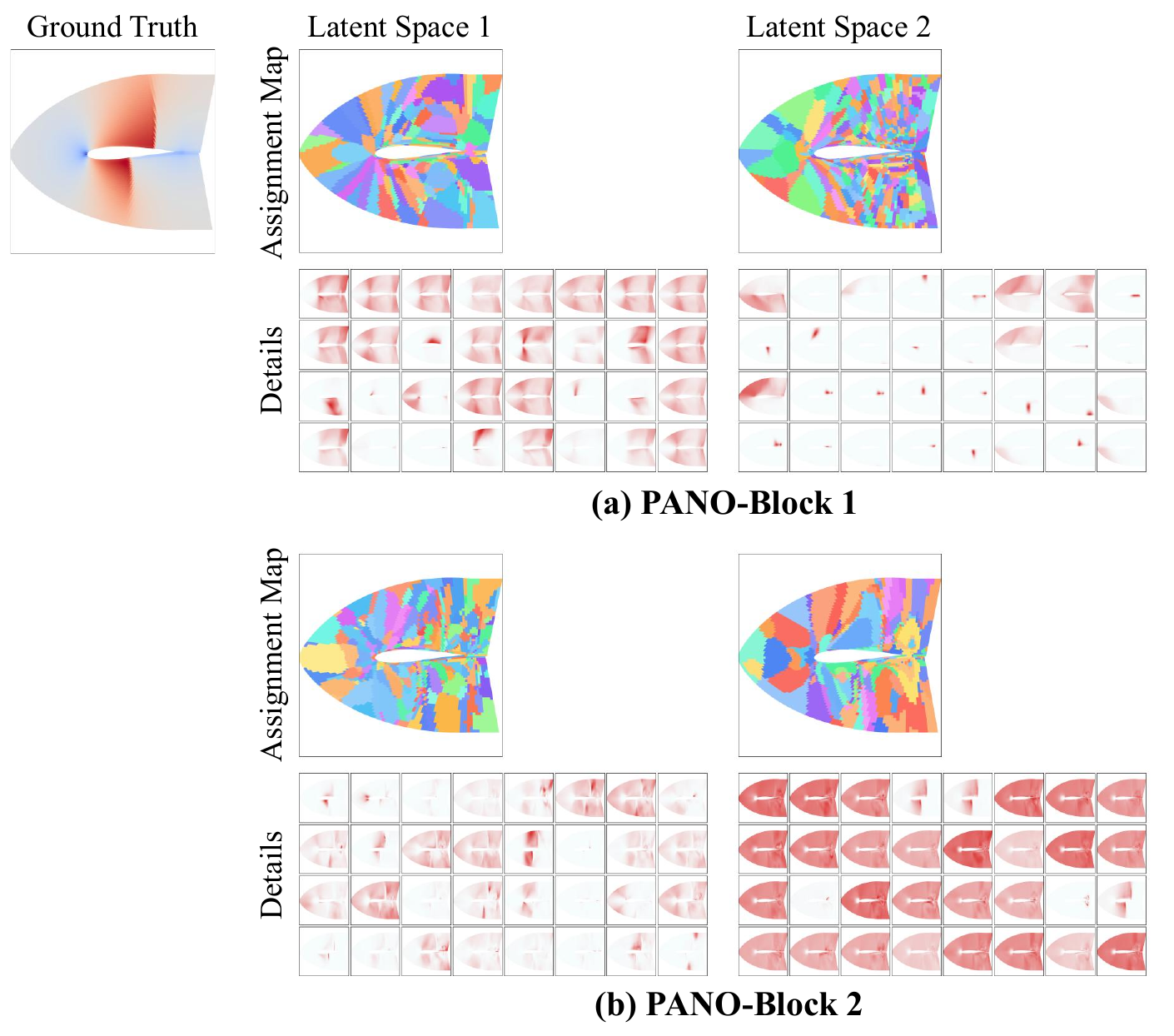}
    \caption{Visualization of PANO-D assignment maps on Airfoil.}
    \label{fig:airfoil_router_pano-d}
\end{figure}

\begin{figure}[!t]
    \centering
    \includegraphics[width=0.85\linewidth]{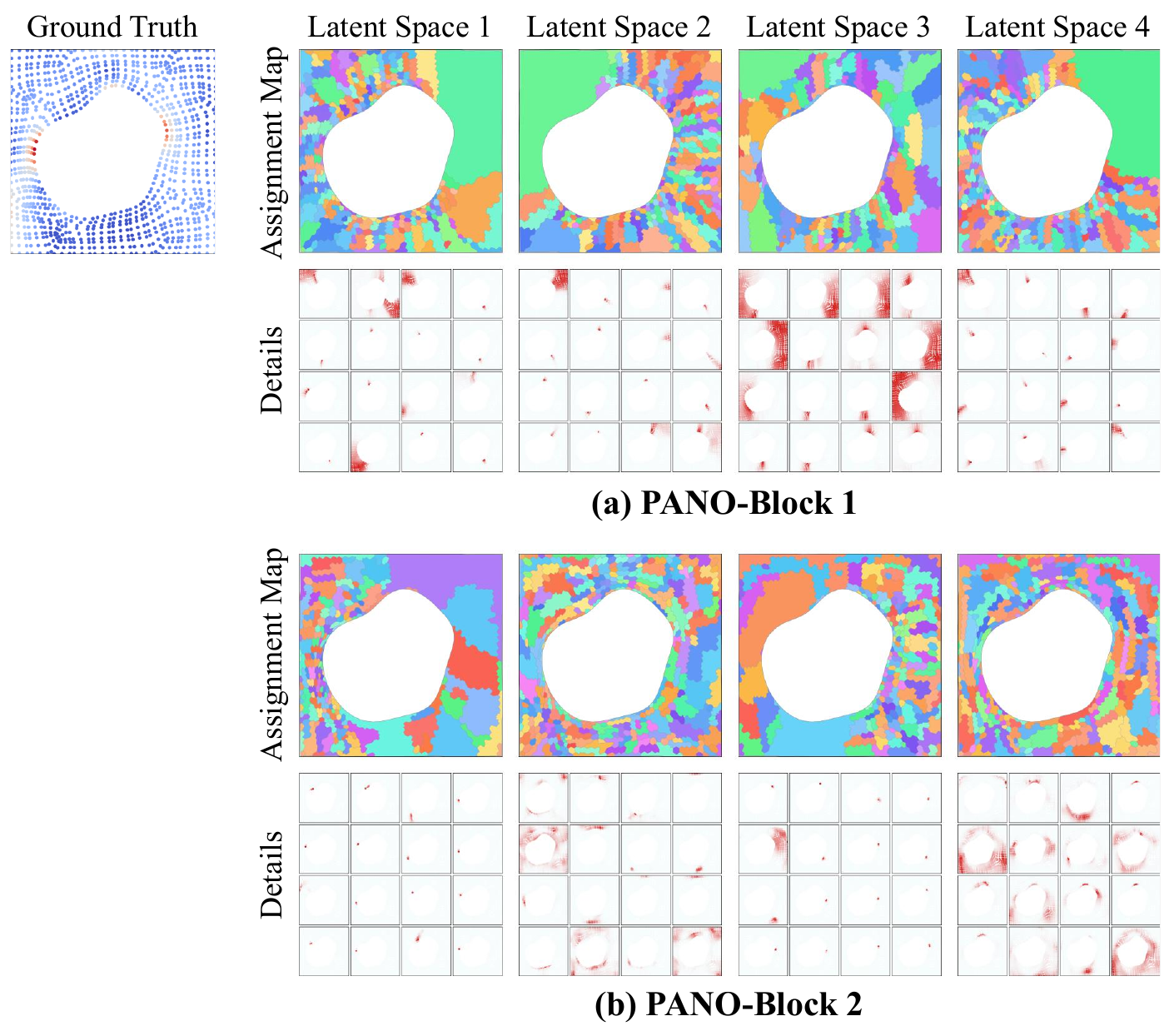}
    \caption{Visualization of PANO-P assignment maps on Elasticity.}
    \label{fig:elasticity_router_pano-p}
\end{figure}

\begin{figure}[!t]
    \centering
    \includegraphics[width=0.85\linewidth]{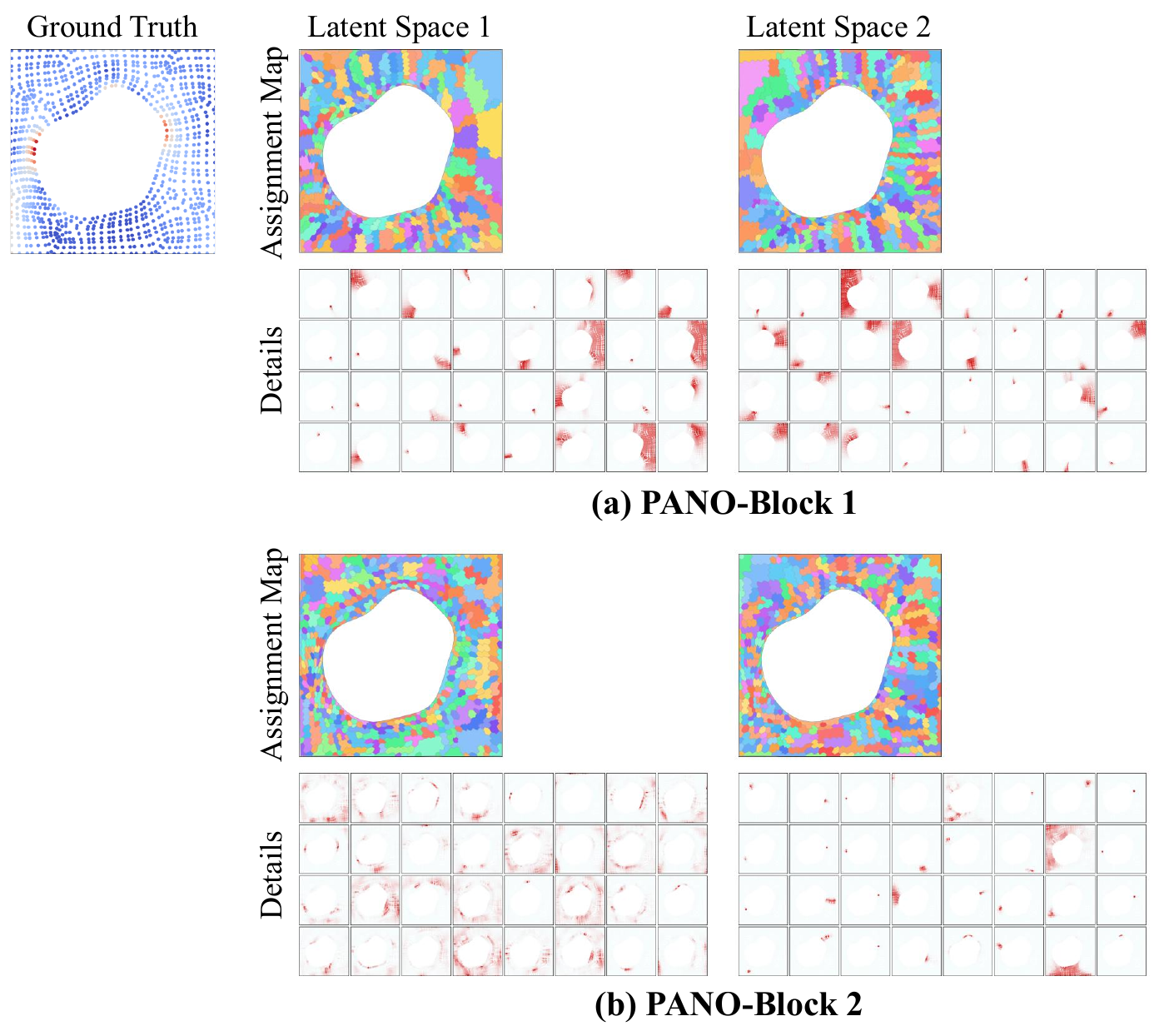}
    \caption{Visualization of PANO-D assignment maps on Elasticity.}
    \label{fig:Elasticity_router_pano-d}
\end{figure}

\begin{figure}[!t]
    \centering
    \includegraphics[width=0.85\linewidth]{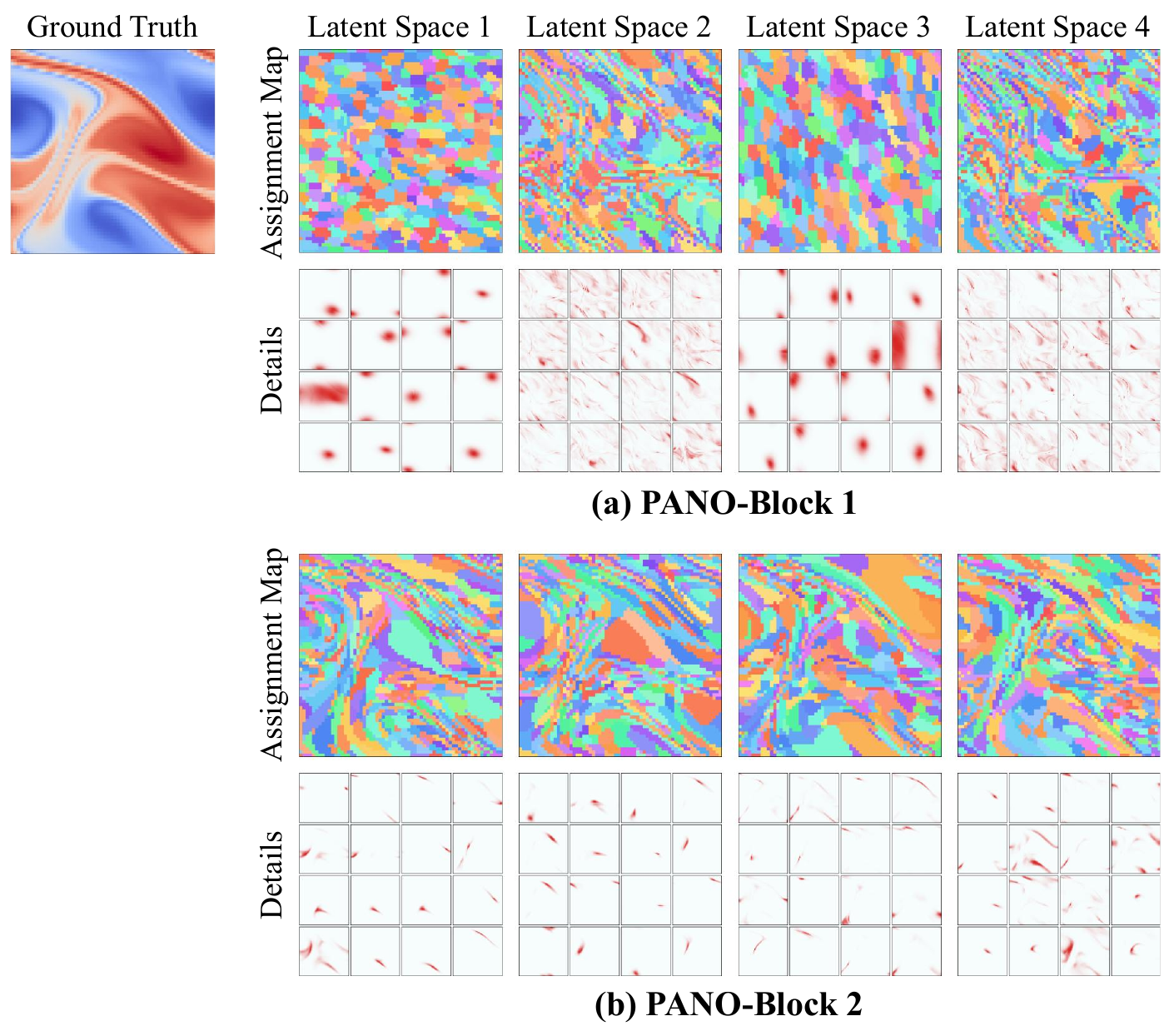}
    \caption{Visualization of PANO-P assignment maps on NS2D.}
    \label{fig:ns2d_router_pano-p}
\end{figure}

\begin{figure}[!t]
    \centering
    \includegraphics[width=0.85\linewidth]{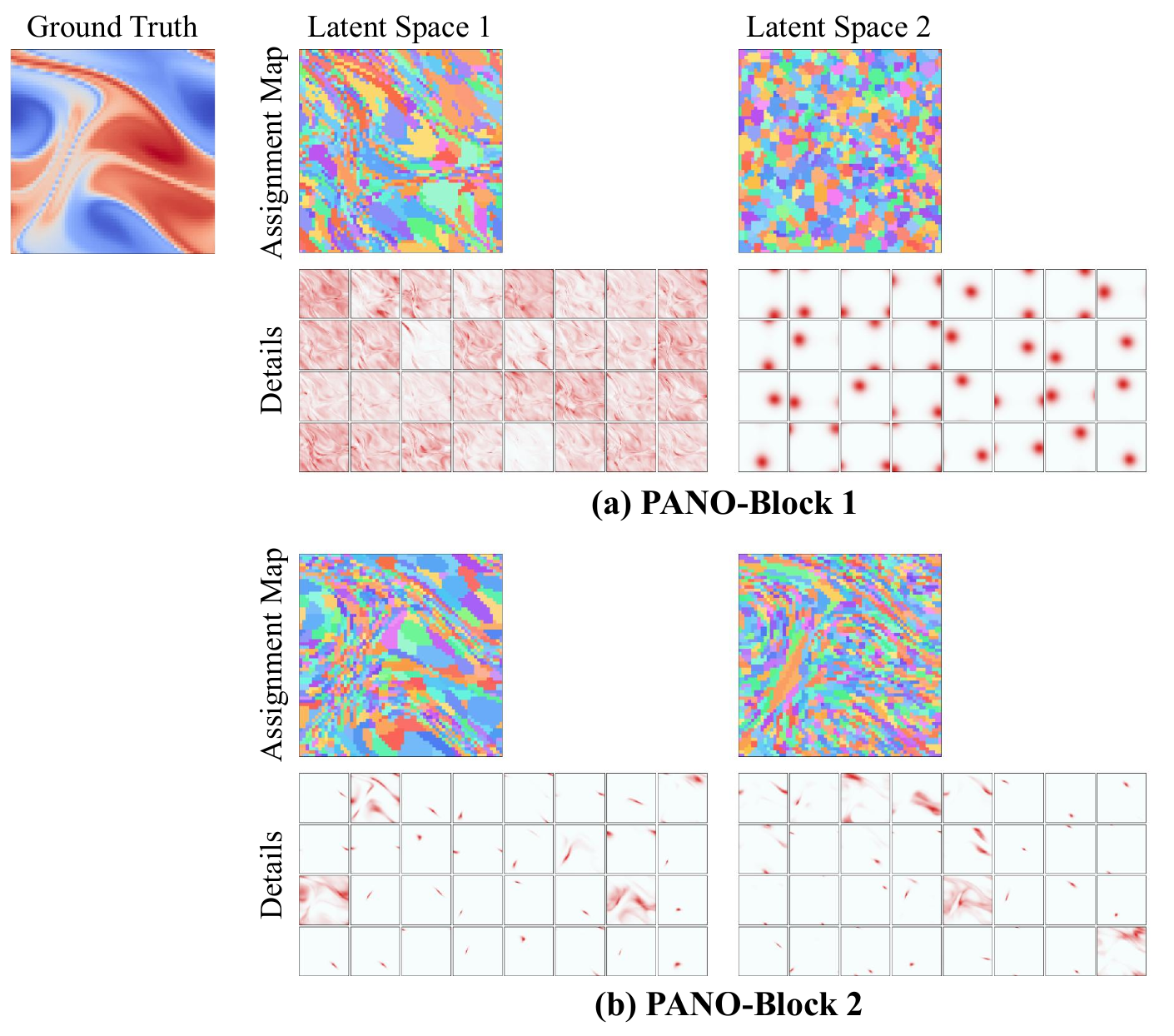}
    \caption{Visualization of PANO-D assignment maps on NS2D.}
    \label{fig:ns2d_router_pano-d}
\end{figure}

\begin{figure}[!t]
    \centering
    \includegraphics[width=0.85\linewidth]{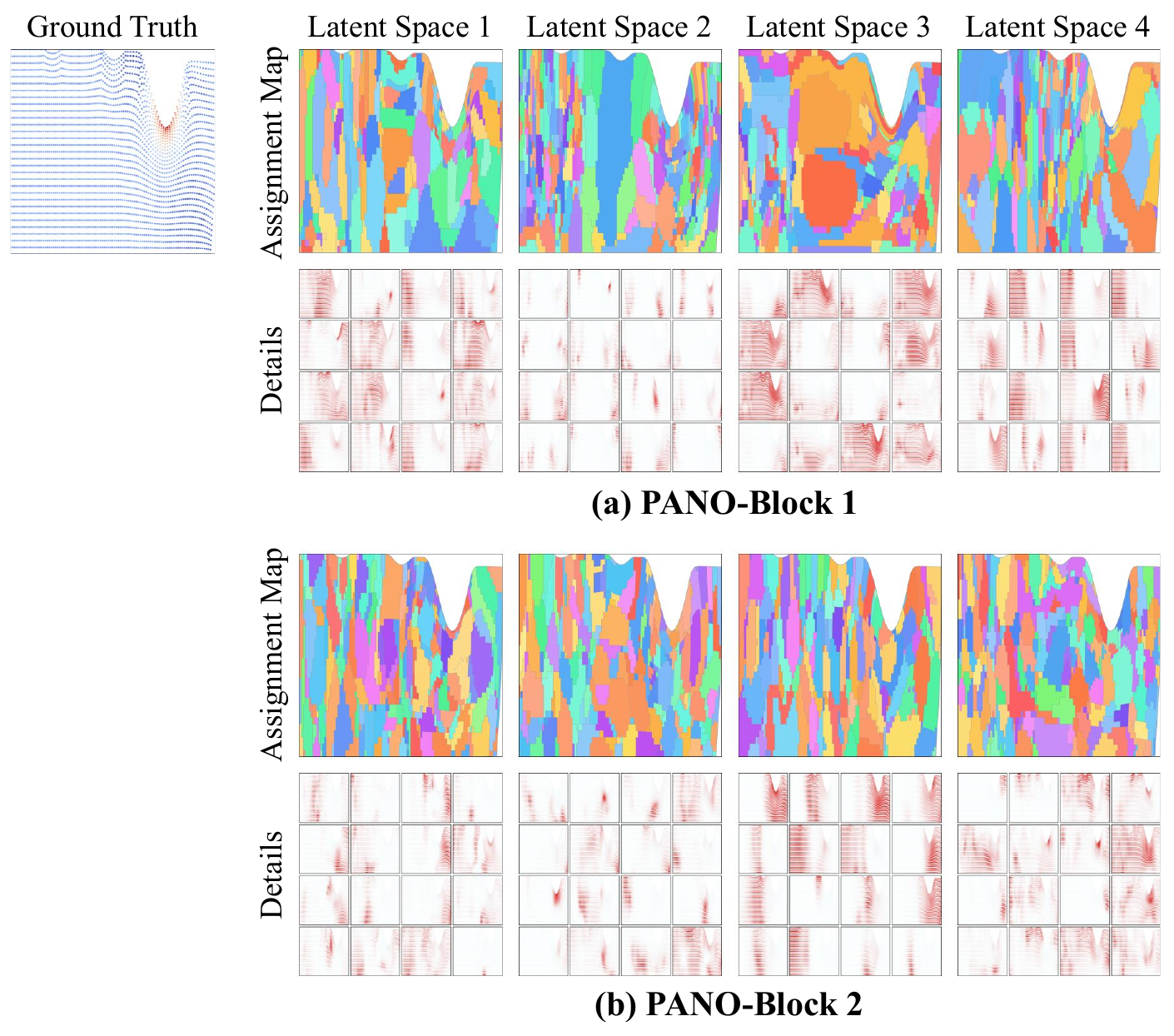}
    \caption{Visualization of PANO-P assignment maps on Plasticity.}
    \label{fig:plasticity_router_pano-p}
\end{figure}

\begin{figure}[!t]
    \centering
    \includegraphics[width=0.85\linewidth]{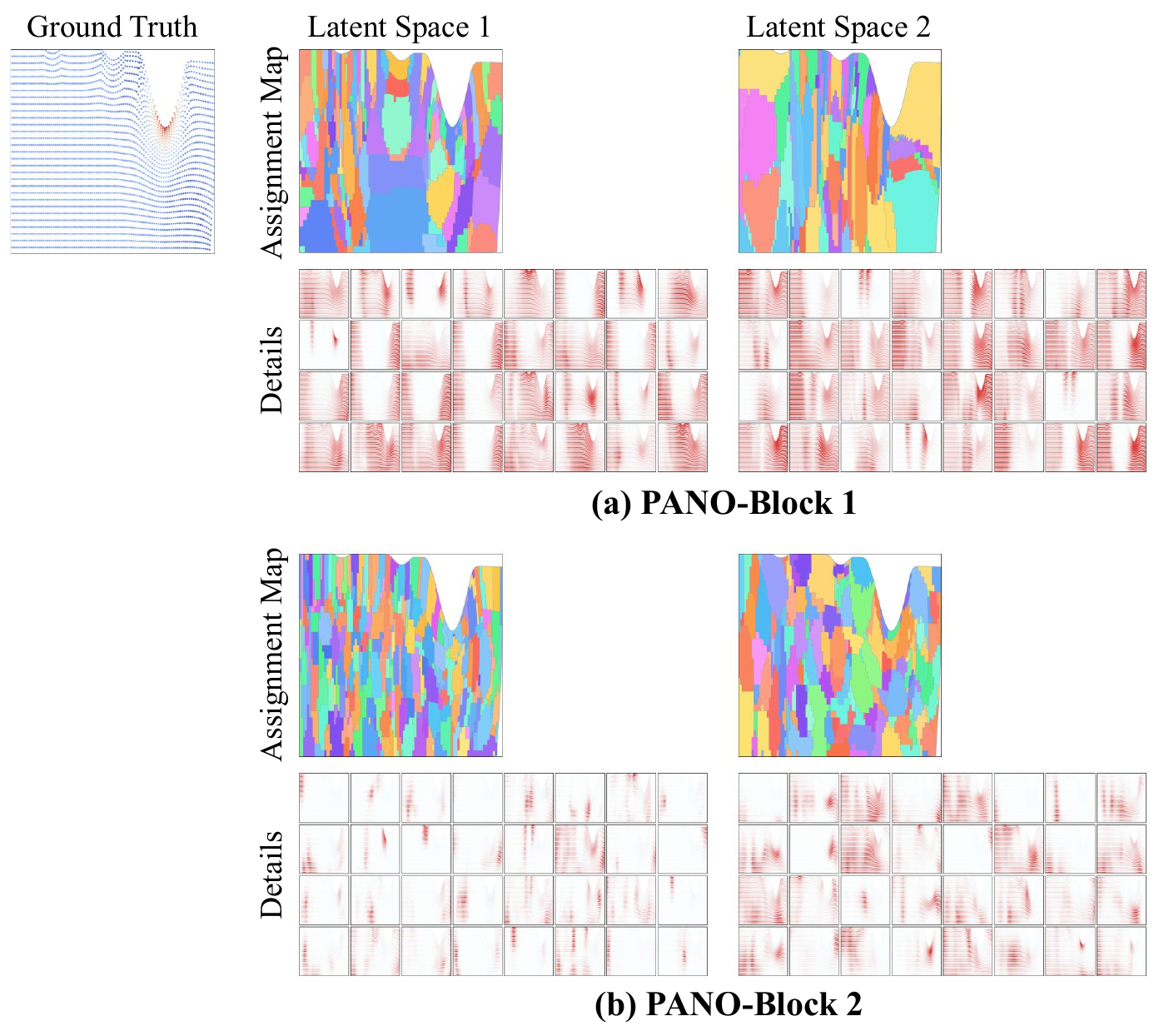}
    \caption{Visualization of PANO-D assignment maps on Plasticity.}
    \label{fig:plasticity_router_pano-d}
\end{figure}

\begin{figure}[!t]
    \centering
    \includegraphics[width=0.85\linewidth]{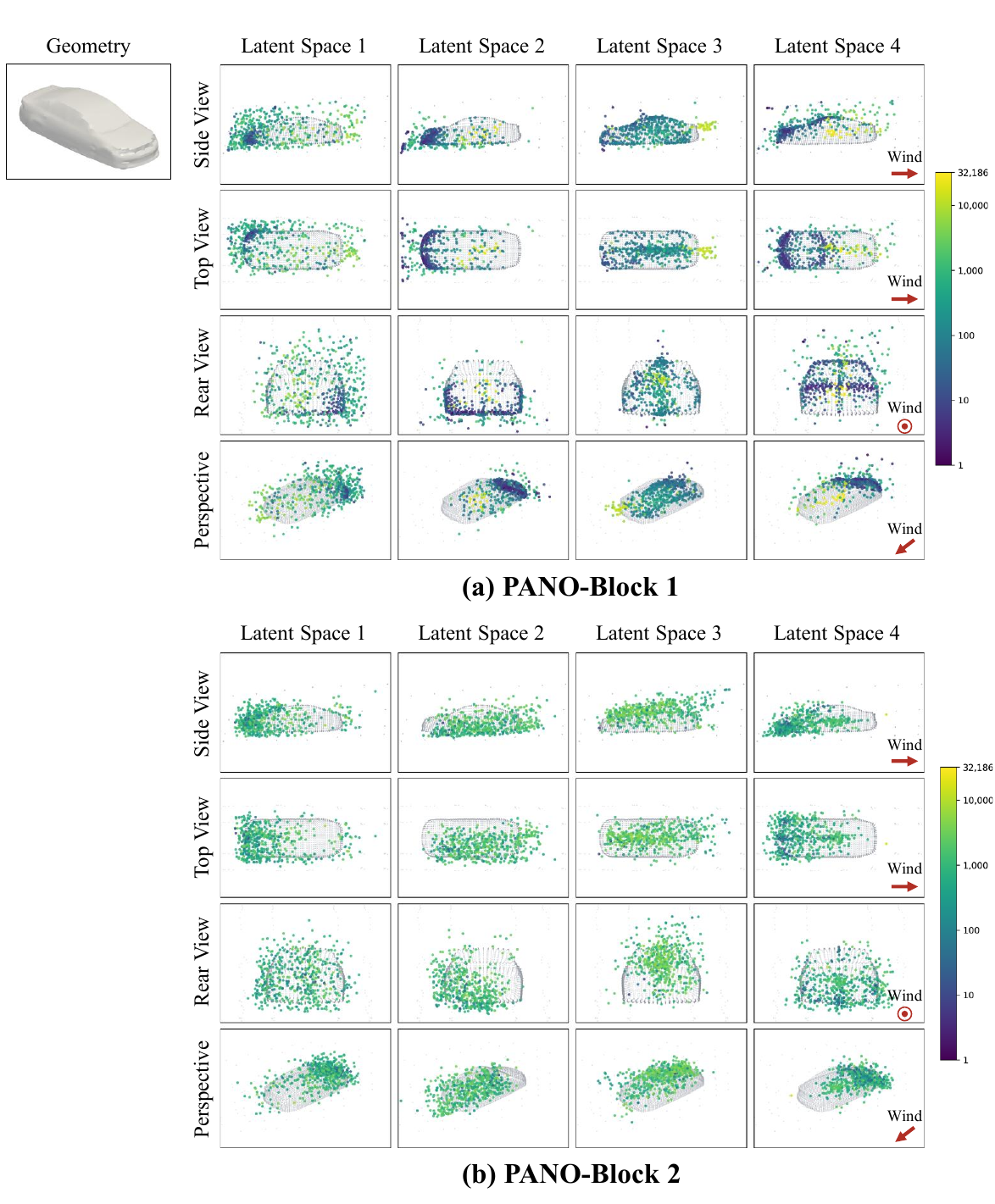}
    \caption{Visualization of latent token centers learned by PANO-P on ShapeNetCar.}
    \label{fig:shapenetcar_router_pano-p}
\end{figure}

\begin{figure}[!t]
    \centering
    \includegraphics[width=0.85\linewidth]{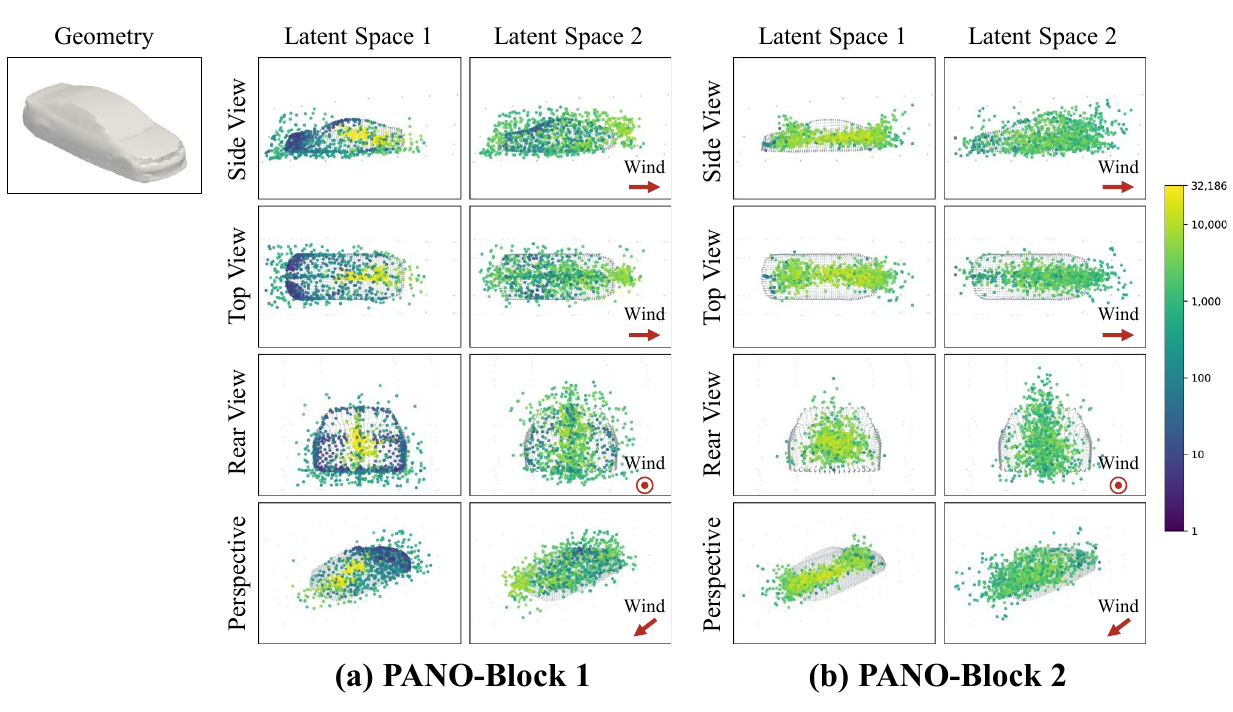}
    \caption{Visualization of latent token centers learned by PANO-D on ShapeNetCar.}
    \label{fig:shapenetcar_router_pano-d}
\end{figure}

\noindent \textbf{Latent-Space Representations.}
We further visualize the latent-space assignments learned by PANO-P and PANO-D.
Assignment maps are shown for 2D benchmarks, while latent-token spatial centers are shown for ShapeNetCar.
Results for Darcy85, Darcy141, and Darcy211 are shown in Figures~\ref{fig:darcy85_router_pano-p}-\ref{fig:darcy85_router_pano-d}, Figures~\ref{fig:darcy141_router_pano-p}-\ref{fig:darcy141_router_pano-d}, and Figures~\ref{fig:darcy211_router_pano-p}-\ref{fig:darcy211_router_pano-d}, respectively.
Additional visualizations are provided for Airfoil in Figures~\ref{fig:airfoil_router_pano-p}-\ref{fig:airfoil_router_pano-d}, Elasticity in Figures~\ref{fig:elasticity_router_pano-p}-\ref{fig:Elasticity_router_pano-d}, NS2D in Figures~\ref{fig:ns2d_router_pano-p}--\ref{fig:ns2d_router_pano-d}, Plasticity in Figures~\ref{fig:plasticity_router_pano-p}--\ref{fig:plasticity_router_pano-d}, and ShapeNetCar in Figures~\ref{fig:shapenetcar_router_pano-p}-\ref{fig:shapenetcar_router_pano-d}.

